\documentclass[journal,transmag]{IEEEtran}
\ifCLASSINFOpdf
\else
\fi

\usepackage{xcolor,soul,framed} 
\usepackage{booktabs, multirow} 
\usepackage[colorinlistoftodos]{todonotes}
\colorlet{shadecolor}{yellow}
\usepackage{changepage,threeparttable}
\usepackage[cmex10]{amsmath}
\usepackage{array}
\usepackage{mdwmath}
\usepackage{mdwtab}
\usepackage{eqparbox}
\usepackage{url}
\usepackage{tabularx}
\usepackage{amsfonts}
\usepackage{amssymb}
\usepackage{acronym}
\usepackage{verbatim}
\usepackage{xfrac,scalerel}
\usepackage{adjustbox}
\usepackage{graphicx}
\usepackage{gensymb}
\usepackage[justification=centering]{caption}
\newcommand{\mycomment}[1]{}

\usepackage{scalerel}
\usepackage{tikz}
\usetikzlibrary{svg.path}

\definecolor{orcidlogocol}{HTML}{A6CE39}
\tikzset{
  orcidlogo/.pic={
    \fill[orcidlogocol] svg{M256,128c0,70.7-57.3,128-128,128C57.3,256,0,198.7,0,128C0,57.3,57.3,0,128,0C198.7,0,256,57.3,256,128z};
    \fill[white] svg{M86.3,186.2H70.9V79.1h15.4v48.4V186.2z}
                 svg{M108.9,79.1h41.6c39.6,0,57,28.3,57,53.6c0,27.5-21.5,53.6-56.8,53.6h-41.8V79.1z M124.3,172.4h24.5c34.9,0,42.9-26.5,42.9-39.7c0-21.5-13.7-39.7-43.7-39.7h-23.7V172.4z}
                 svg{M88.7,56.8c0,5.5-4.5,10.1-10.1,10.1c-5.6,0-10.1-4.6-10.1-10.1c0-5.6,4.5-10.1,10.1-10.1C84.2,46.7,88.7,51.3,88.7,56.8z};
  }
}

\newcommand\orcidicon[1]{\href{https://orcid.org/#1}{\mbox{\scalerel*{
\begin{tikzpicture}[yscale=-1,transform shape]
\pic{orcidlogo};
\end{tikzpicture}
}{|}}}}

\usepackage{hyperref}

\begin{document}
%
\title{Revisiting Generative Adversarial Networks for Binary Semantic Segmentation on Imbalanced Pavement Datasets}


\author{Lei~Xu\IEEEauthorrefmark{1} \orcidicon{0000-0003-0115-2770},~\IEEEmembership{Student Member,~IEEE,}

      and~Moncef Gabbouj\IEEEauthorrefmark{1}
      \orcidicon{0000-0002-9788-2323},~\IEEEmembership{Fellow,~IEEE}
      \\
      \IEEEauthorblockA{\IEEEauthorrefmark{1}Department of Computing Sciences, Tampere University, Finland\\
Emails: \IEEEauthorrefmark{1}lei.xu@tuni.fi,
\IEEEauthorrefmark{1}moncef.gabbouj@tuni.fi}

\thanks{Corresponding author: L. Xu (Email: lei.xu@tuni.fi).}}

\markboth{arXiv}%
{Shell \MakeLowercase{\textit{et al.}}: Bare Demo of IEEEtran.cls for IEEE Transactions on Magnetics Journals}
%



\IEEEtitleabstractindextext{%
}

\maketitle

\begin{abstract}
Anomalous crack region detection is a typical binary semantic segmentation task, which aims to detect pixels representing cracks on pavement surface images automatically by algorithms. Although existing deep learning-based methods have achieved outcoming results on specific public pavement datasets, the performance would deteriorate dramatically on imbalanced datasets. The input datasets used in such tasks suffer from severely between-class imbalanced problems, hence, it is a core challenge to obtain a robust performance on diverse pavement datasets with generic deep learning models. To address this problem, in this work, we propose a deep learning framework based on conditional Generative Adversarial Networks (cGANs) for the anomalous crack region detection tasks at the pixel level. In particular, the proposed framework containing a cGANs and a novel auxiliary network is developed to enhance and stabilize the generator's performance under two alternative training stages, when estimating a multiscale probability feature map from heterogeneous and imbalanced inputs iteratively. Moreover, several attention mechanisms and entropy strategies are incorporated into the cGANs architecture and the auxiliary network separately to mitigate further the performance deterioration of model training on severely imbalanced datasets. We implement extensive experiments on six accessible pavement datasets. The experimental results from both visual and quantitative evaluation show that the proposed framework can achieve state-of-the-art results on these datasets efficiently and robustly without acceleration of computation complexity. The code is available at \url{https://github.com/LeiXuAI/Imbalanced_crack_detection_with_GANs.git}.

\end{abstract}


\IEEEdisplaynontitleabstractindextext

%
\IEEEpeerreviewmaketitle

\section{Introduction}

\IEEEPARstart{C}{racks} 
can be distinguished from road surfaces clearly by human vision, however, such manual effort consumes time and human labor \cite{Yang2019}. Therefore, alternative approaches based on deep learning have been fully explored with the prosperous development of deep learning techniques for real applications. Usually, the pavement images captured under complex situations suffer from a severely between-class imbalanced distribution and noisy backgrounds. The different capturing conditions cause the intensity inhomogeneity of cracks and complexity of the background \cite{Yang2019}, aggravating the convergence of generic models. Moreover, the between-class imbalanced datasets can lead to a deteriorated result as the network may converge to the state labeling all positive pixels for cracks as negative ones for background \cite{Zhang2021CrackGAN:Learning}, when employing generic algorithms directly \cite{Sampath2021}. Furthermore, the anomalous crack region detection on the pavement images is a typical binary semantic segmentation task \cite{Tabernik2020, Liu2019, Xia2020}, because it targets to delicately categorize each pixel in an image either into anomalous class (cracks) labeling with $1$ or not the anomalous class (backgrounds) labeling with $0$.

Strategies and approaches to solving the between-class imbalance problems can be mainly categorized into data-level, algorithm-level, and hybrid-level \cite{Rezaei2019a, Sampath2021}. Usually, the data-level approaches are used to manipulate the number of data samples to restore the balance on imbalance datasets, such as the strategy of re-sampling data samples using Synthetic Minority Over-sampling Technique (SMOTE) \cite{Chawla2002, Sampath2021} or expanding data samples with synthesizing transformation technique \cite{Simard2003, Sampath2021}. The algorithm-level approaches aim to enhance the contribution of minor classes \cite{Rezaei2019a} using specific objective functions, such as dice loss \cite{Rezaei2017a}, focal loss \cite{Lin2020}. Hybrid-level approaches take advantage of data-level and algorithm-level simultaneously \cite{Sampath2021}. Many traditional computer vision algorithms have been intensively employed to tackle the class imbalance semantic segmentation tasks efficiently based on the exploration of poster-possibility \cite{Lee2005, Krahenbuhl2011}, region-based information \cite{Caesar2015}, or spatial context information \cite{Yang2014}, etc.

Researchers have devoted abundant efforts to addressing such tasks either with traditional computer vision methods \cite{Shi2016, Lim2014, Nakazawa2019}, or deep learning models \cite{Liu2019, Yang2019, Tabernik2020, Son2017, Rezaei2019a, Zou2019DeepCrack:Detection}. 
The existing state-of-the-art 
Although current works have achieved state-of-the-art performance, these works face some drawbacks. (1) Existing traditional methods usually require either a complicated and expensive hardware system \cite{Mohan2018, Nithya2015a, Meksen2010} or specific image processing algorithms as a pre-processing step \cite{Cubero-Fernandez2017a} or feature descriptor \cite{Shi2016} for more accurate outcomes. (2) Existing deep learning models seldom achieve robust performance generality on diverse datasets \cite{Yang2019}. (3) To gain improvement on complex datasets (captured under low light, polymorphic cracks structure, etc), it is necessary to increase the complexity of the network either from computation or parameters. 

This work aims to tackle anomalous crack region detection tasks using the algorithm-level strategy on pavement datasets captured under diverse conditions. The key challenge of developing an efficient and effective deep-learning model with robust performance on diverse datasets lies in restoring the severely between-class imbalanced distribution of the original inputs and distinguishing the heterogenous crack pixels from the noisy backgrounds. To conquer the limits existing in current works, in this work, we explore the conditional Generative Adversarial Networks (cGANs) working with a novel auxiliary network \cite{Mirza2014} as a framework to learn a more accurate probability feature map on diverse pavement datasets in two stages. The proposed method is to build a cGANs architecture network to restore the balance of extremely imbalanced data at the first stage considering the potential of generative models on restoring balance for imbalance tasks \cite{Sampath2021}. After that, a novel auxiliary network si assists in training the generator for a refined generation and learns to distinguish the ground truth and its corresponding generation precisely at the second stage. The extensive experiments on six diverse pavement datasets

The contributions of our work are as follows: (1) we proposed an end-to-end cGANs-based network to segment the anomalous crack regions on road surface images in two stages. The first stage utilizes the architecture of cGANs as the backbone to initially estimate the distribution of cracks with a UNet-based generator and a discriminator. The second stage utilizes the training of the generator and a novel auxiliary network for a refined probability feature map of cracks. (2) The entropy originating from information theory has been exploited as the perceptual and reconstruction loss from the output of the auxiliary network and the corresponding ground truth to refine the outcomes of the generator. (3) We exploited a side network originally used for small objects or line detection tasks working with our proposed backbone network for this severely imbalanced task. (4) We adopted three kinds of attention mechanisms with the generator to highlight the crack regions. (6) We employed extensive experiments on six pavement imbalanced datasets to demonstrate that the proposed cGANs-based framework can achieve more robust and precise quantitative results on diverse inputs and a satisfactory balance between effectiveness and efficiency. 

The remainder of this paper is organized as follows. Section II presents the relevant literature on binary semantic segmentation. Section III describes the details of our proposed framework. The experimental setting and results are given in Section IV. Section V is the conclusion of this work.

\section{Related Works}
This section briefly introduces the related works with deep learning models on binary semantic segmentation tasks related to our proposed framework. 

\subsection{Deep Learning Techniques for binary semantic Segmentation}
Intense efforts have been devoted to exploring deep learning models for binary semantic segmentation tasks, such as edge detection, crack detection, anomalous regions detection on medical images, etc. Xie et al. \cite{Xie2015} proposed a multi-scale network architecture based on VGGNet \cite{Simonyan2015VeryRecognition} and deeply supervised nets called holistically-nested edge detection (HED). The HED can be used to learn the natural image edge and object boundary in an image-to-image manner directly with the nested multi-scale feature learning architecture. Moreover, Xie et al. \cite{Xie2015} introduced a class-balanced cross-entropy loss function to offset the imbalance between edge and non-edge. Due to the similarity between edge detection and crack detection, the HED structure can be used as the backbone architecture involved in crack detection \cite{Yang2019, Liu2021CrackFormer:Detection, Liu2019, Zou2019DeepCrack:Detection}. Liu et al. \cite{Liu2019} introduced a HED-like automatic network named DeepCrack with a modified cross-entropy loss function for pavement crack segmentation. The DeepCrack consists of the extended Fully Convolutional Networks(FCN) for feature map learning and the deeply supervised nets (DSN) for feature map integration of each convolutional stage. Yang et al. proposed a feature pyramid and hierarchical boosting network (FPHBN) in \cite{Yang2019} to detect cracks on road surfaces based on the HED architecture. To enhance the performance of hard samples, Yang et al. introduced the hierarchical boosting module into the FPHBN \cite{Yang2019} from top to bottom, by which a sample reweighting operation is conducted between adjacent side networks. In \cite{Liu2021CrackFormer:Detection, Liu2023CrackFormerSegmentation}, a transformer-based encoder-decoder architecture named CrackFormer was proposed by incorporating self-attention and scaling-attention mechanisms for pavement crack detection. The CrackFormer \cite{Liu2021CrackFormer:Detection} uses the transformer-based self-attention blocks to establish long-range interaction between the low-level features, and then the scaling-attention block is used for sharpening crack boundaries. Zou et al. \cite{Zou2019DeepCrack:Detection} proposed a multi-scale deep convolutional neural network learning hierarchical features for automatic surface-crack detection. The proposed network \cite{Zou2019DeepCrack:Detection} is designed with a SegNet-based \cite{Badrinarayanan2017SegNet:Segmentation} architecture and pixel-wise prediction loss calculated by a skip-layer fusion procedure for binary semantic segmentation on pavement images.  

Furthermore, generative adversarial networks (GANs) can be extended to tackle various binary semantic segmentation tasks, due to the proven functionality of alleviating the performance deterioration caused by imbalanced problems \cite{Sampath2021}. A V-GANs architecture \cite{Son2017} was proposed to generate the probability maps of retinal vessels from given fundus images. An UNet-based generator with skip-connection is used to retain low-level features such as edges in the V-GANs architecture, and the discriminator has several models with different output sizes, such as pixel-level, image-level, and patch-level. The proposed architecture combined with an objective function as the summation of the GANs objective and the segmentation loss has achieved successful outcomes \cite{Son2017}. Kamran et al. proposed a multi-scale generative adversarial network called RVGAN for accurate retinal vessel segmentation in \cite{Kamran2021}. RVGAN \cite{Kamran2021} consists of two generators and two multi-scale autoencoder-based discriminators, combining reconstruction and weighted feature matching loss, for high accuracy on pixel-wise segmentation of retinal vessels. To alleviate data imbalance problems, Zhang et al. \cite{Zhang2021CrackGAN:Learning} proposed a CrackGAN for pavement crack detection. The CrackGAN utilized the crack-patch-only supervised generative adversarial learning strategy and an asymmetric U-Net architecture generator to reserve cracks even when the cracks are thin or ground truth images are partially accurate \cite{Zhang2021CrackGAN:Learning}. A refined cGANs proposed in \cite{Rezaei2019a} is trained in two stages aiming to solve binary semantic segmentation with a lower miss-classification rate caused by imbalanced data in medical images. The proposed method in Rezaei's work \cite{Rezaei2019a} consists of a generator network, a discriminator network, and a refinement network. The UNet-based generator produces a corresponding segmentation label and the discriminator determines whether its input is real from ground truth or fake from the generator. The refinement network with the UNet architecture can learn the false prediction of the cGAN.

\subsection{Attention Mechanism}
Attention mechanism in deep learning \cite{Bahdanau2015NeuralTranslate} was originally proposed to improve the performance of a recurrent neural network (RNN) encoder-decoder architecture with a soft searching strategy in the Natural Language Processing (NLP) field. Furthermore, an attention mechanism can be employed in visual tasks to emphasize on required salient parts using a long short-term memory (LSTM) decoder with two different attention strategies \cite{Xu2015ShowAttention}. In \cite{Oktay2018b}, the attention mechanism was first introduced into a feed-forward convolutional neural network (CNN) model U-net for medical image analysis. The proposed attention mechanisms integrating with UNet can simplify previous segmentation frameworks in an end-to-end manner, instead of with separate localization and segmentation steps \cite{Oktay2018b}. The attention UNet architecture has achieved highly beneficial results in the identification and localization of small-size organs \cite{Oktay2018b, Schlemper2019}. Since then, attention mechanism strategies have been intensely explored incorporating deep learning models for various specific topics \cite{Hu2018Squeeze-and-ExcitationNetworks, Laakom2021LearningCNNs, Wang2022UCTransNet:Transformer, Brauwers2023ALearning}. 

Hu et al. \cite{Hu2018Squeeze-and-ExcitationNetworks} designed SENet architectures based on the squeeze-and-excitation (SE) block, which is used to highlight the relationship between the channels. The proposed SE block utilizes feature maps across spatial dimensions with global average-pooling for channel-wise attention, and then feature recalibration is completed by reweighting the feature maps with the attention descriptor. The stacked SE blocks work with various deep learning models as SENets, which provide an effective compromise between complexity and performance \cite{Hu2018Squeeze-and-ExcitationNetworks}. Woo et al. proposed a simple and effective attention module named the Convolutional Block Attention Module (CBAM). To further emphasize informative features, the CBAM consists of two principal attention models to extract cross-channel and spatial information together. The cross-channel attention model utilizes the concept of SE block \cite{Hu2018Squeeze-and-ExcitationNetworks} but with an extra max-pooling branch for optimal feature extraction. The spatial attention model learns the inter-spatial dependencies of features and decides the location to focus on as a complementarity of the channel attention. Sufficient experiments have verified that the CBAM can enhance the representation power of deep learning models from 'what' and 'where' aspects. Laakom et al. \cite{Laakom2021LearningCNNs} further validated the SE block and CBAM model incorporating a novel learning to ignore schema. The difference from general attention mechanisms \cite{Woo2018CBAMModule, Oktay2018b}, the proposed learning to ignore schema aims to highlight irrelevant or confusing information in the feature maps. To achieve this goal, an ignoring mask is introduced into the CBAM model. The ignoring mask suppresses the high values for irrelevant attributes and regions in the feature map and the final attention mask is obtained by flipping the ignoring mask. 

Recently, self-attention-based architecture: Transformer \cite{Dosovitskiy2021ANSCALE} has been widely explored to combine CNN-like architectures for image recognition due to its effectiveness in retaining long-distance features \cite{Liu2023CrackFormerSegmentation}. To fit the original natural language processing model, the input image is split into patches as tokens. Then each token is mapped with linear embedding operation for a sequence as an input to a Transformer. In \cite{Wang2022UCTransNet:Transformer}, Wang et al. proposed a new segmentation framework UC-TransNet, with a Channel Transformer module (CTrans). To replace the UNet skip connections with multi-scale channel-wise fusion, the CTrans module is designed with two sub-modules: multi-scale Channel Cross fusion with Transformer (CCT) module and Channel-wise Cross-Attention module. Such architecture can mitigate the ambiguity caused by skip connection in the original UNet. To retain the ability of long-range correlation of Transformer and eliminate its high computational complexity, Wang et al. proposed a Mixed Transformer Module (MTM) in \cite{Wang2022MixedSegmentation}. The MTM is a U-shaped model with self-attention modules for fine-grained local features and coarse-grained global features in deeper layers at a lower complexity. 

\subsection{Loss Functions for Imbalance Problems}
To mitigate the bias of commonly used loss functions for severely imbalanced datasets, various efforts \cite{Lin2017FeatureDetection, Abraham2019ASegmentation, Zhang2020ARecognition, Cui2019Class-balancedSamples, Salehi2017TverskyNetworks} have been devoted to investigating proper loss functions for highlighting the importance of minor class(es). In \cite{Lin2017FeatureDetection}, Lin et al. proposed a focal loss to address class imbalance problems. The focal loss is defined based on the binary cross entropy (CE) loss incorporating the focusing parameter $\gamma$. In practice, the focal loss can work with an $\alpha-$balanced hyperparameter to improve learning further. Salehi et al. introduced the Tversky similarity index into a generalized loss function to address imbalance problems in \cite{Salehi2017TverskyNetworks}. The proposed Tversky loss function manipulates the magnitude of penalties between false positive samples and false negative samples through two hyperparameters $\alpha$ and $\beta$. Abraham et al. further extended the Tversky similarity index as Focal Tversky Loss (FTL) motivated by the exponential combination of Dice score and cross-entropy \cite{Abraham2019ASegmentation}. The FTL is used for severely imbalanced datasets and small regions of interest segmentation using a trade-off between precision and recall. Yeung et al. proposed another variant of focal loss: Unified Focal Loss in \cite{Yeung2022UnifiedSegmentation}. The Unified Focal Loss is a general framework with Dice-based and Cross-entropy loss functions, which aims to group functionally equivalent hyperparameters and emphasize the suppressive and enhancing effects of the focal parameters after grouping. 

\section{Proposed methods}
In this section, we first present the proposed network framework in detail. Then the attention mechanisms incorporated in the network are illustrated. After that, the loss functions are depicted, and the discussions among the proposed method and other related methods are presented.

\subsection{Overview of Proposed Methods}
\begin{figure}[ht]
\centering
  \includegraphics[width=0.7\linewidth]{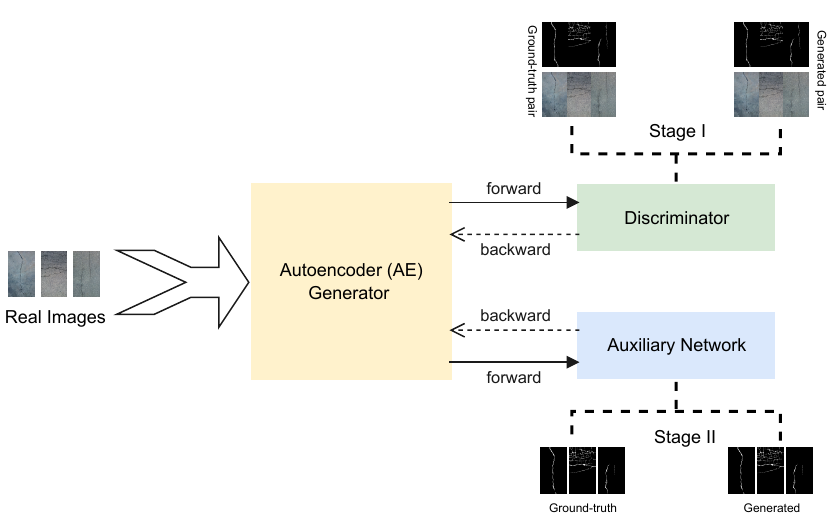}
  \caption{The general schema of our proposed method.}
  \label{Fig:general_schema}
\end{figure}

As shown in Fig. \ref{Fig:general_schema}, the general training schema of our proposed methods consists of two stages. The pipeline for Stage I is based on the cGANs architecture which contains an encoder-decoder generator and a discriminator working together in a min-max zero game manner for a probability feature map generation. Moreover, we explore two network architectures for the discriminator to figure out the most efficient structure to solve such binary semantic segmentation tasks with severely imbalanced inputs. The pipeline for Stage II contains the same encoder-decoder generator at Stage I and a simple auxiliary network to regularize and improve the authenticity of the probability feature map from the generator at Stage I. Stage I and Stage II are trained iteratively to achieve optimal results. We adopt the attention UNet \cite{Oktay2018b} and feature pyramid networks \cite{Lin2017FeatureDetection} for the generator architecture. Furthermore, we explore three strategies of attention mechanism \cite{Bahdanau2015NeuralTranslate}, entropy-based loss functions \cite{Abraham2019ASegmentation}, \cite{Cover1991ElementsTheory} aiming to figure out an optimal and robust framework for the severe class imbalance problems.

\mycomment{
\begin{figure*}[ht]
\centering
  \includegraphics[width=1.0\textwidth]{Figures/Scheme_for_cracks.jpg}
  \caption{The network architecture of our proposed method}
  \label{Fig:network_architecture}
\end{figure*}}

\subsection{Conditional Generative Adversarial Networks}

Generative Adversarial Nets (GANs) architecture was originally proposed in \cite{Goodfellow2013}. The original architecture of GANs constitutes a generator (G) and a discriminator (D) with a noise vector $\mathbf{z} \sim p_{\mathbf{z}}$ as its input. The objective of the G with the noisy input is to generate fake data which can deceive the D to evaluate it as the real data. The D aims to distinguish the real data and the generated fake data. The goals of the generator and discriminator are opposite as the name of GANs shows. Conditional Generative Adversarial Nets (cGANs) \cite{Mirza2014} is an extension work of GANs, which can introduce extra information conditioned on both G and D. The vanilla objective function of the cGANs is formulated as
\begin{multline} \label{eq:vanilla_cgan_loss}
\underset{G}{min}\:\underset{D}{max}\:V(D,G) = \mathbb{E}_{\mathbf{x}\sim p_{data}(\mathbf{x})}[log(D(\mathbf{x} | \mathbf{y}))] + \\
\mathbb{E}_{\mathbf{z}\sim p_{\mathbf{z}}(\mathbf{z})} [log(1 - D(G(\mathbf{z} | \mathbf{y})))].
\end{multline}
Here $D(\mathbf{x|y})$ indicates the output of the discriminator with a probability whether $\mathbf{x}$ fits the distribution of real data or the generated fake data conditioned on some extra information $\mathbf{y}$. $G(\mathbf{z|y}$) presents the fake output of the generator based on the joint hidden representation of the prior input noise $p_{\mathbf{z}}(\mathbf{z})$ and the extra information $\mathbf{y}$. 

\begin{figure}[ht]
\centering
  \includegraphics[width=1.0\linewidth]{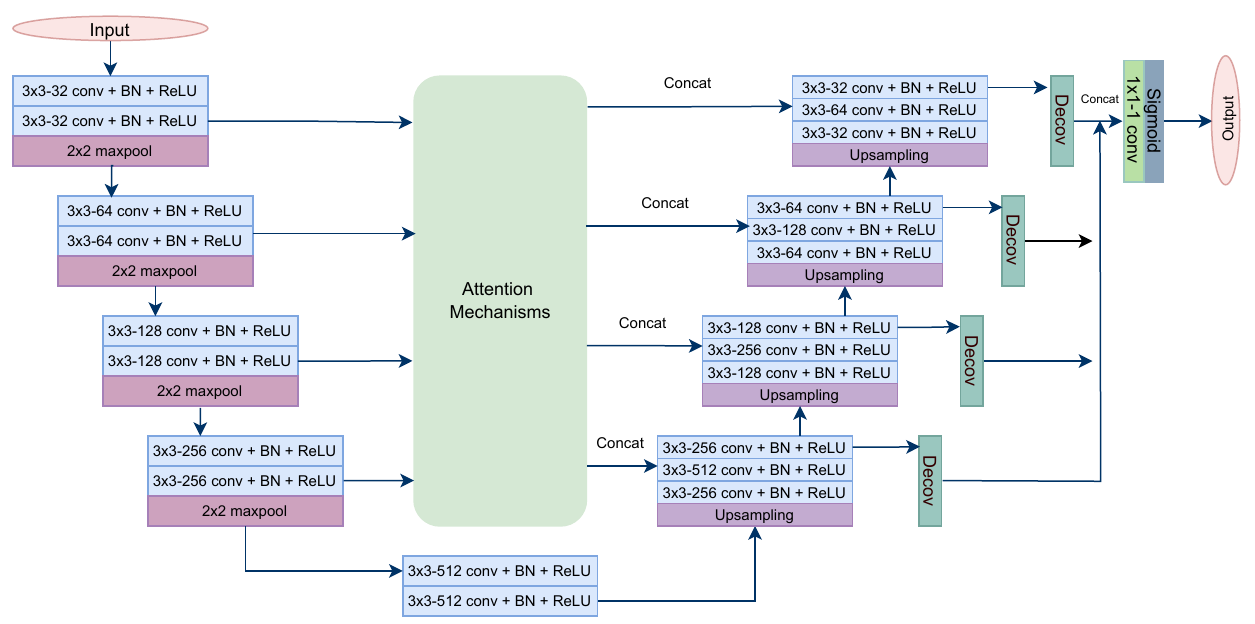}
  \caption{The generator architecture of our proposed method.}
  \label{Fig:generator}
\end{figure}
\subsubsection{Generator}
Generator G produces a continuous probability feature map close to a binary image conditioned on the corresponding input. We integrate the network structure strategies of attention UNet \cite{Oktay2018b} and conditional generative adversarial nets (cGANs) \cite{Mirza2014} as shown in Fig. \ref{Fig:generator}. The encoder-decoder generator with skipped connections aims to achieve pixel-wise binary segmentation with a probability feature map. The proposed generator contains 23 convolutional layers and each convolutional layer is followed by a batch normalization (BN) and a rectified linear unit (ReLU) sequentially besides the last one directly followed by a sigmoid activation function for the final output. Along with the contracting path of the generator, four $2\times2$ max-pooling layers are carried out to implement down-sampling as in \cite{Ronneberger2015}. 
 
Considering the efficiency and effectiveness of the attention mechanism on explicit detection and emphasis of relevant parts \cite{Laakom2021LearningCNNs, Oktay2018b}, we adapt the attention mechanism into the skipped connection to preserve the minor but important information of crack pixels and suppress the irrelevant but major information of background pixels in feature maps. The input of attention gates (AGs) is the output of the convolutional layers on the contracting path. The output of the AGs is concatenated into the convolutional layers on the expansive path. As mentioned above, the significant challenge in such crack detection tasks using convolutional neural networks (CNNs) is all pixels could be labeled as background pixels when converging \cite{Zhang2021CrackGAN:Learning}, due to the imbalance problems. To preserve features of cracks as much as achievable, we use four multi-stage side layers \cite{Xie2015, Yang2019} on the expansive path for hierarchical features fusion. The sigmoid activation is then used for each multi-stage side layer and the outputs from the sigmoid activation are concatenated after up-sampling as the same size of the input to form a fused output. More sparse features such as location and boundaries can be preserved using up-sampling. After that, The final output of the generator is the fused multi-stage feature map refined by a convolutional layer and the sigmoid activation.

\subsubsection{Discriminator}

\begin{figure}[ht]
\centering
  \includegraphics[width=0.8\linewidth]{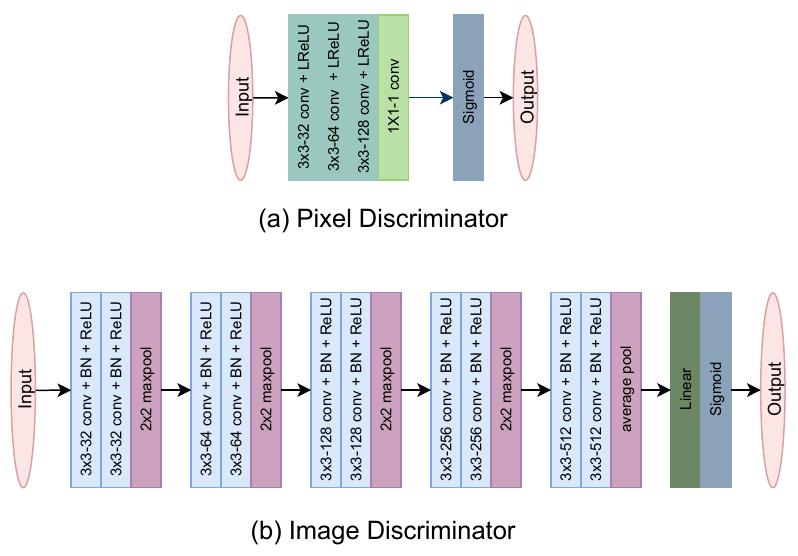}
  \caption{Two discriminator architectures.}
  \label{Fig:discriminators}
\end{figure} 

\begin{figure}[ht]
\centering
  \includegraphics[width=0.8\linewidth]{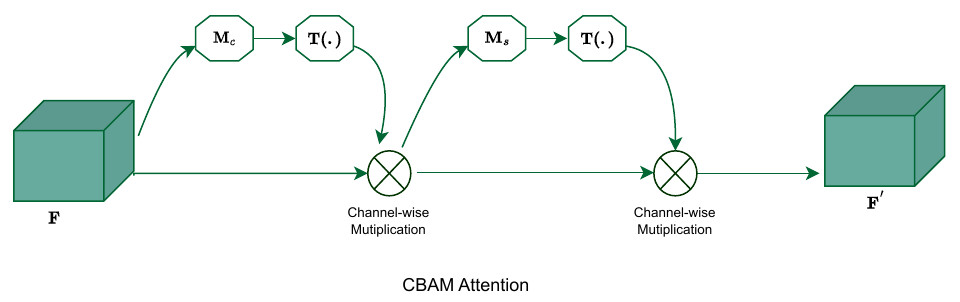}
  \caption{CBAM attention architectures.}
  \label{Fig:atten_cbam}
\end{figure}

\begin{figure}[ht]
\centering
  \includegraphics[width=0.8\linewidth]{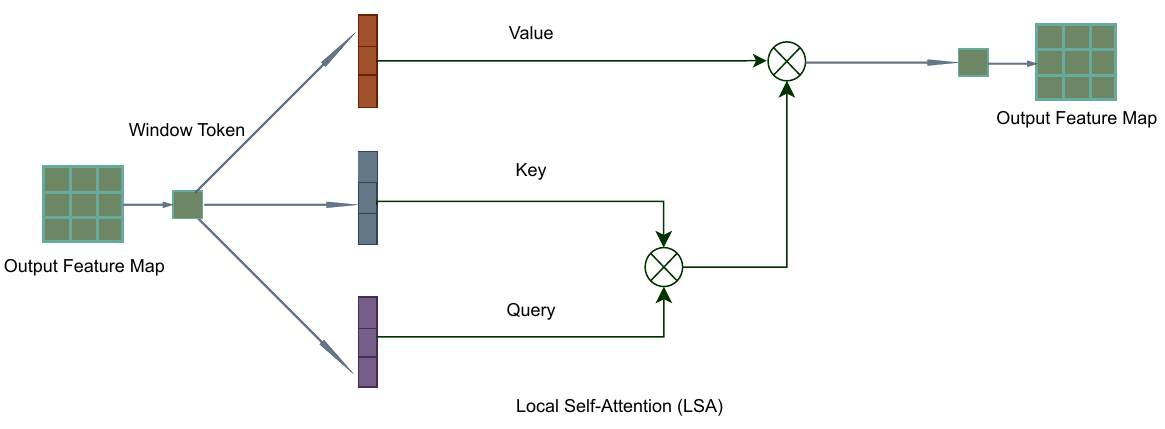}
  \caption{Local self-attention (LSA) layer architectures.}
  \label{Fig:atten_vit}
\end{figure}

\begin{figure}[ht]
\centering
  \includegraphics[width=0.8\linewidth]{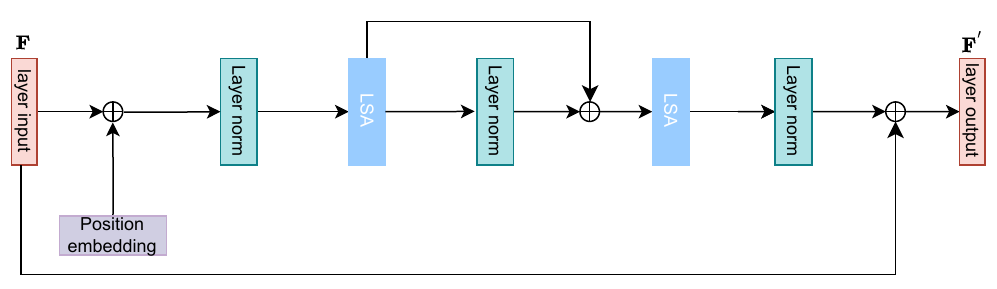}
  \caption{LSA module structure.}
  \label{Fig:atten_lsa}
\end{figure}

The discriminator D needs to distinguish whether its input pair belongs to the real pair or the generated pair with learned probability. Hence the D should maximize the output of the image pair \{$\mathbf{X}$, $\mathbf{Y}$\} and minimize the output of the image pair \{$\mathbf{X}$, $\mathbf{\hat{Y}}$\}. To investigate how a discriminator architecture can affect its judgment results, we exploited two discriminator architectures \cite{Zhu2017UnpairedNetworks, Son2017}: pixel-level and image-level discriminators. The pixel-level discriminator is a lightweight but computationally complex convolutional neural network with three $3\times3$ convolution layers and each one is followed by a leaky ReLU activation sequentially. The last convolutional layer is a $1\times1$ convolution layer followed by a sigmoid activation function. The final output is a probability matrix the same size as the raw input image to categorize each pixel belonging to a crack or background. The image-level discriminator constitutes ten $3\times3$ convolutional layers and each is followed by a BN and a ReLU sequentially as a convolutional block. Besides, we adopt four max-pooling layers and one average-pooling layer after every two convolutional blocks sequentially. After that, a linear mapping function is used to produce a final probability value to indicate whether the input pair is real or fake. The pixel-discriminator and the image-discriminator architectures are demonstrated in Fig. \ref{Fig:discriminators}. 

\subsubsection{Auxiliary Network}
Although the performance of cGANs on binary semantic segmentation is promising enough \cite{Son2017, Rezaei2019a}, it is still difficult to force the generator for a real binary output but a close to binary output directly. Hence, we adopt an auxiliary network working with the encoder-decoder generator to refine its outputs for a closer binary output. The auxiliary network works with the generator in parallel with the discriminator as shown in Fig. \ref{Fig:general_schema} at Stage II. The architecture of the auxiliary network is the same as the pixel discriminator shown in Fig. \ref{Fig:discriminators}(a). During the training Stage II, the auxiliary network $\Phi$ is trained with the inputs of the ground truth $\mathbf{Y}$ and the generated $\mathbf{\hat{Y}}$. The outputs of the auxiliary network consist of two parts. The first one is used to update the auxiliary network based on the Kullback-Leibler (KL) divergence \cite{Cover1991ElementsTheory} to enhance the learning ability of the auxiliary network. The latter one as a feature reconstruction loss \cite{Johnson2016PerceptualSuper-resolution} is for refining the output of the generator based on the binary cross-entropy loss. 

\subsection{Attention Mechanism}
The challenge of losing required feature fields while training is more severe in this work not only due to the down-sampling in the network but also the extremely imbalanced input images. To tackle this problem, we adopt three kinds of attention mechanisms: a Convolutional Block Attention Module (CBAM) \cite{Woo2018CBAMModule}, a transformed CBAM for ignoring \cite{Laakom2021LearningCNNs}, and a Local Self-Attention \cite{Wang2022MixedSegmentation} module to capture more semantic features from the relevant areas further. As demonstrated in Fig. \ref{Fig:generator}, the multi-scale attention mechanisms are integrated into the generator through four skip connections.

\begin{itemize}
    
\item CBAM \cite{Woo2018CBAMModule, Laakom2021LearningCNNs} is a lightweight and general attention model for any CNN architecture training end-to-end for refined feature maps. The CBAM module constitutes two sequential sub-modules as an attention block: a channel-wise module for a $1D$ channel attention map and a spatial-wise attention module for a $2D$ spatial attention map. The channel attention module has a Global Average Pooling and a Global Max Pooling to generate a channel attention map $\mathbf{M}_c$. The spatial attention module exploits the inter-spatial relationship of input feature maps from the channel-wise module with a spatial attention map $\mathbf{M}_s$. Laakom et al. proposed a novel attention mechanism called learning to ignore in \cite{Laakom2021LearningCNNs} as learning to identify and suppress the irrelevant areas based on the CBAM architecture. Contrary to the CBAM, the learning-to-ignore approach learns to identify irrelevant areas by a channel-ignoring mask and a spatial ignoring mask separately using a channel-wise module and a spatial-wise module as in CBAM. The high pixel values in the ignoring mask present the background. To obtain the final attention mask, an inverse function $\mathbf{T(.)}$ is introduced into the CBAM framework to transpose the importance of background and crack pixels in an ignoring mask. Therefore, when the CBAM architecture in Fig. (\ref{Fig:atten_cbam}) is used to learn a refined feature map, the $\mathbf{T(.)}$ equals $1$, otherwise a function with an inversely proportional to its input ignoring mask. In this work, we adopt the function for flipping the ignore mask from \cite{Laakom2021LearningCNNs}
\begin{equation} \label{eq:inverse}
\mathbf{T(M_c(F))} = 1 - \mathbf{M_c(F)},
\end{equation}
where $\mathbf{M_c(F)}$ presents the channel-wise ignoring mask matrix with values bonded between $[0, 1]$. 

\item Local self-attention \cite{Wang2022MixedSegmentation} (LSA) layer is derived from self-attention \cite{Dosovitskiy2021ANSCALE}. The only difference is that the LSA layer is calculated based on neighboring pixels to investigate the correlation of nearby areas inside each local window. We adopted the LSA to highlight the importance of adjacent pixels. The key (K), query (Q), and value (V) matrices are calculated based on each window for attention map computation as shown in Fig. \ref{Fig:atten_vit}. Then the LSA module for each skip connection consists of two LSA layers demonstrated in Fig. \ref{Fig:atten_lsa}.
\end{itemize}

\subsection{Objective Functions}
In this work, we first adopt the general cGAN loss \cite{Mirza2014} during the training stage I. Then, we proposed a novel perceptual loss and reconstruction loss for the auxiliary network to refine the output of the G during stage II. Moreover, we investigated the effectiveness of Tversky loss \cite{Abraham2019ASegmentation, Salehi2017TverskyNetworks} and side loss \cite{Xie2015, Yang2019} to our proposed framework. 

Give a training dataset with $N$ image pairs as $\{(\mathbf{X}_i, \mathbf{Y}_i), i = 1, ..., N\}$, where $\mathbf{X}_i$ presents a raw road RGB image and $\mathbf{Y}_i$ is the corresponding binary ground-truth image. $\{(\mathbf{X}_i, \hat{\mathbf{Y}}_i), i = 1, ..., N\}$ present the generated image pairs, where $\hat{\mathbf{Y}}$ is the generated probability feature map closer to a binary feature map. Let G map a raw RGB road image $\mathbf{X}$ into a probability feature map $\hat{\mathbf{Y}}$. 

\subsubsection{cGANs Loss}
without the random variable $\mathbf{z}$ in the generator, the $\mathbf{z}$ from the original definition can be eliminated by a deterministic generator G \cite{Nguyen2017}. Hence, the objective function of the cGANs is formulated as

\begin{multline} \label{eq:cgans_loss}
\mathcal{L}_{cGANs}(G, D) = \mathbb{E}_{\mathbf{X}\sim p_{data}(\mathbf{X, Y})}[log(D(\mathbf{X, Y}))] + \\
\mathbb{E}_{\mathbf{X}\sim p_{data}(\mathbf{X})} [log(1 - D(\mathbf{X}, G(\mathbf{X})))].
\end{multline}

Here the G is trained to learn the distribution of input data $p_{data}(\mathbf{X, Y})$ as $G: \mathbf{X} \rightarrow \mathbf{\hat{Y}}$. The output $\mathbf{\hat{Y}}$ contains values between [0, 1] to indicate the probability of each pixel belonging to cracks or background. The D maps the ground truth image pair $\{\mathbf{X}_i, \mathbf{Y}_i \}$ or the generated image pair $\{\mathbf{X}_i, \hat{\mathbf{Y}}_i\}$ into a probability decision map with the pixel-level discriminator or a probability decision value with the image-level discriminator as shown in Fig. \ref{Fig:discriminators}. The probability decision map is $\mathbf{F}$ with the same size of $\hat{\mathbf{Y}}$ to decide whether each pixel is real or fake. The probability decision value $\mathbf{f}$ is between [0, 1] to decide whether the input pair is real or fake. 

\subsubsection{Perceptual Loss and Reconstruction Loss}
The auxiliary network $\phi$ takes the ground-truth $\mathbf{Y}$ and its corresponding generated feature map $\hat{\mathbf{Y}}$ as its inputs. With a similar strategy in \cite{Johnson2016PerceptualSuper-resolution}, we use the auxiliary network to learn an approximate pixel-level feature representation between $\phi (\hat{\mathbf{Y}}) $ and $\phi (\mathbf{Y}) $ computed by the KL-divergence loss rather than direct pixel matching between $\mathbf{Y}$ and $\hat{\mathbf{Y}}$. The loss based on the auxiliary network consists of a perceptual loss and a reconstruction loss. The perceptual loss trains the auxiliary network for a precise approximation function in stage II while the generator training is fixed. The definition is

\begin{equation} \label{eq:perceptual_loss}
\mathcal{L}_{KL} = \sum \phi (\mathbf{Y}) log \frac{\phi(\mathbf{Y})}{\phi (\hat{\mathbf{Y}})},
\end{equation}

Furthermore, we also wish the auxiliary network to audit the generator. To achieve this goal, we adopt the reconstruction loss to penalize the output difference of the auxiliary network. The reconstruction loss is calculated based on the binary cross-entropy between the absolute difference values of the auxiliary network outputs and a zero matrix.

\begin{equation} \label{eq:reconstruction_loss}
\mathcal{L}_{BCE} = H(\lvert \phi(\mathbf{Y})-\phi (\hat{\mathbf{Y}})\rvert; [0]),
\end{equation}

\subsubsection{Side Network Loss}
Side network loss was initially introduced in Xie's work \cite{Xie2015} and widely used in related crack or line detection \cite{Liu2021CrackFormer:Detection, Yang2019, Liu2019a} to automatically balance the loss between positive and negative classes with a class-balancing weight $\beta$ on a per-pixel term basis. With the target to trade off precision and recall, the $\beta$ incorporates the binary cross-entropy loss with an output from
a sigmoid function $\sigma(.)$ for each side layer loss and the final fuse loss. Then the side layer losses and fuse loss are calculated by the binary cross-entropy \cite{Sun2022DMA-Net:Segmentation} 

\begin{multline} \label{eq:bce_loss}
\mathcal{L}_{bce} = -\frac{1}{N} \sum_{n=1}^N \beta_p y_n log(p_n) + (1 - y_n)log(1 - p_n), 
\end{multline}
where $p_n$ is the predicted pixel $n$ in an output probability map and $y_n$ is the corresponding ground truth pixel $n$. The total loss from the side network is denoted as

\begin{equation} \label{eq:sn_loss}
\mathcal{L}_{Side} = \sum_{i=1}^4 (\mathcal{L}_{bce})_{side}^i + (\mathcal{L}_{bce})_{fuse}, 
\end{equation}
 
\subsubsection{Tversky Loss}
Tversky Loss is defined based on the Tversky index (TI) \cite{Salehi2017TverskyNetworks} aiming to detect minor classes for severely imbalanced datasets. To achieve this goal, two parameters $\alpha$ and $\beta$ are introduced for precision and recall trade-off. The Tversky index for binary semantic segmentation is defined \cite{Salehi2017TverskyNetworks}
\begin{equation} \label{eq:tversky_ind}
TI = \frac{\sum_{i=1}^N p_{1i} g_{1i} + \epsilon}{\sum_{i=1}^N p_{1i} g_{1i} + \alpha  \sum_{i=1}^N p_{1i} g_{0i} + \beta \sum_{i=1}^N p_{0i} g_{1i} + \epsilon}.
\end{equation}
Here $p_{1i}$ presents the pixel $i$ as a crack pixel with the probability and $p_{0i}$ is for the background. $g_{1i}=1$ for a crack pixel $i$ and $g_{1i} = 0$ for a background pixel $i$, vice versa for $g_{0i}$.  

The Tversky loss is then defined 
\begin{equation} \label{eq:tversky_index}
\mathcal{L}_{TL} = 1 - TI.
\end{equation}

The whole loss function is then defined as
\begin{equation} \label{eq:total_loss}
\mathcal{L} = \gamma \mathcal{L}_{cGANs} + \mathcal{L}_{KL} + \mathcal{L}_{CE} + \mathcal{L}_{Side} + \mathcal{L}_{TL}.
\end{equation}

\subsection{Relation to Other Methods}
\textit{Relation to medical image segmentation}: The imbalance problems have widely existed in medical image segmentation tasks as binary semantic segmentation topics\cite{Son2017, Wang2022MixedSegmentation}. The backbone structure of our proposed work is referred from the V-GAN \cite{Son2017}. However, there are three main differences in our framework: 1) In general, our proposed framework works with two training stages for learning the refined representation. 2) We add attention gates and multiscale side layers in the generator architecture to solve the server imbalance situation. 3) Besides the loss of cGAN, we adopt novel entropy-based losses working with the auxiliary network. Moreover, the LSA module is mainly referred from the Mixed Transformer Module (MTM) in \cite{Wang2022MixedSegmentation}. Some inherent parts of the MTM are not suitable for this work, hence we only adopt the local self-attention layer to rebuild the LSA module.

\section{Experiments and Results}
In this section, the experimental settings as well as datasets for this work are briefly depicted first. Then we present the evaluation metrics, complexity analysis, and ablation study. Finally, we compare our proposed network with the state-of-the-art works on the same datasets with the evaluation metrics and visual results. 

\subsection{Experimental Settings}
We train our model on NVIDIA GPUs Tesla V100 with Tensorflow 1.15. We adopt the Adam optimizer \cite{Kingma2015Adam:Optimization} for stage I and stage II training with a learning rate of $0.0001$ and the momentum term as $0.2$. The initial training iteration number is $50000$ with batch size $8$, and the trained model is evaluated every $2000$ iterations with the validation datasets. During the validation, the best model is selected according to the average value of dice, accuracy, sensitivity, and specificity calculated between a predicted feature map after binarizing with the otsu filter \cite{Otsu1979a} and its ground-truth image. 

The other hyperparameters are set as follows. An LSA module contains two LSA layers with a window size of $8$ as Fig. \ref{Fig:atten_lsa} to calculate LSA attention maps. Then, we have $\beta_p = 1 $ in Eq. \eqref{eq:bce_loss} to avoid the trade-off between precision and recall from the side network. Moreover, we set $\alpha = 0.3$ and $\beta = 0.7$ to emphasize recall (false negative pixels) more than precision in the Tversky loss in Eq.\eqref{eq:tversky_index}, aiming to improve the performance for severely imbalanced datasets. Furthermore, we wish the proposed network could pay more attention to the stage II training, hence we set $\gamma = 0.25$ in Eq. \eqref{eq:total_loss}.

\subsection{Datasets}

We conduct extensive experiments to verify the effectiveness of our proposed framework on six datasets: CRACK500, CrackTree26, CrackLS315, CFD, CRKWH100, and DeepCrack\_DB. Each dataset contains pavement images captured under different situations and places, the corresponding binary ground truth images are annotated in a pixel-wise manner. After image cropping, we retained cropped images with more than 1,000 crack pixels first. Then we carried out the data augmentation by rotation with $90\degree$, $180\degree$, and $270\degree$ and randomly selected $10\%$ for testing and $90\%$ for training and validation on each dataset.    

\textbf{CRACK500 \cite{Yang2019}}: is collected by Yang et al. using cell phones. The original CRACK500 dataset contains 500 pavement images with a size of $2,000 \times 1,500$ pixels and the corresponding binary ground truth images with pixel-wise annotation. In this work, we cropped each original image into a size of $512 \times 512$ pixels with 10\% overlap. The total number of this dataset is $12072$.   

\textbf{CrackTree260\cite{Zou2019DeepCrack:Detection}}: originally contains 260 pavement images with different size $960 \times 720$ and $800 \times 600$ captured by an area-array camera under low light. The pavement state of this dataset is complex and diverse since the images contain various occlusions, shadows, etc. We cropped each original image into a size of $512 \times 512$ pixels with 30\% overlap in width and 10\% overlap in height. The total number of this dataset is $4,112$. 

\textbf{CrackLS315\cite{Zou2019DeepCrack:Detection}}: contains $315$ pavement images captured under laser illumination with a line-array camera. The original image size is $512 \times 512$. We did not crop images on this dataset. The total number of this dataset is 1,260. Notably, the state of pavement images in this dataset is very complex, due to the subtlety and invisibility of crack pixels. 

\textbf{CFD\cite{Shi2016}}: is proposed by Shi et al. \cite{Shi2016}. The original CFD dataset contains 118 images captured by an iPhone5 in Beijing urban areas with a size of $480 \times 320$ pixels. Images from the CFD dataset contain various noises such as shadows, oil spots, and water stains 
\cite{Shi2016}. To augment this dataset, we cropped each original image into a size of $320 \times 320$ pixels with $20\%$ overlap in width. The total number of this dataset is $862$.    

\textbf{DeepCrack-DB \cite{Liu2019}}: contains $537$ pavement images with an original size of $544 \times 384$ pixels. We cropped each original image into a size of $384 \times 384$ pixels with $20\%$ overlap in width. The total number of this dataset is $4,211$.     

\textbf{CRKWH100 \cite{Liu2019}}: originally contains 100 pavement images captured by a line-array camera with a size of $512 \times 512$ pixels. We did not crop images on this dataset. The total number of this dataset is $400$. 

\subsection{Evaluation Metrics}
To evaluate the effectiveness of our proposed framework for the crack detection of severely imbalanced datasets, as well as conducting comparisons between different competing methods, we use five metrics to evaluate the performance of the six pavement datasets: Optimal Dataset Scale (ODS), Optimal Image Scale (OIS), Average Precision (AP), Global Accuracy, and Mean Intersection over Union (IoU). OIS, ODS, and AP as $\mathit{F-}$measure-based metrics are commonly used for line-related detection \cite{Xiao2023PavementTransformers, Zou2019DeepCrack:Detection}. ODS is the best $\mathit{F-}$measure from each dataset for a fixed threshold and OIS is the best $\mathit{F-}$measure from each dataset for the best threshold. Besides, we use Global Accuracy and Mean IoU to evaluate the performance. 

\begin{table}[!htp]\centering
\caption{Complexity Comparison with a 512x512 Input }\label{tab: complexity_comparison }
\begin{tabular}{cccc}
\toprule
Model                & FLOPs   & Params  & Time/image (s) \\\midrule
UNet\cite{Ronneberger2015}                 & 60.91G & 5.76M   &  0.0090            \\
HED  \cite{Xie2015}                & 0.27G  & 2.8K    &  0.0114             \\
FPHB   \cite{Yang2019}               & 273.91G & 44.70M  & 0.0457             \\
V-GAN (pixel)\cite{Son2017}        & 160.98G & 8.04M   & 0.0178         \\
V-GAN (image) \cite{Son2017}       & 115.02G & 12.67M  & 0.0180        \\
DeepCrack \cite{Liu2019}           & 160.65G & 14.72M &  0.0468             \\
Crackformer II \cite{Liu2023CrackFormerSegmentation}        & 176.60G & 4.96M &  0.1060                \\\midrule
cGAN\_LSA (pixel)                & 207.15G &25.26M& 0.0630             \\
cGAN\_CBAM (pixel)           & 179.95G &8.85M  & 0.0194            \\
cGAN\_CBAM\_Ig (pixel)       & 179.95G &8.85M & 0.0203            \\
cGAN\_LSA (image)            & 163.08G & 29.88M & 0.0626             \\
cGAN\_CBAM (image)           & 131.46G &13.47M  & 0.0188           \\
cGAN\_CBAM\_Ig (image)      & 131.46G &13.47M& 0.0188             
\\\midrule
\bottomrule
\end{tabular}
\end{table}
\subsection{Complexity Analysis}
We further evaluate the computational complexity and efficiency of our proposed framework and competing models with floating-point operations (FLOPs), the number of network parameters, and the inference time for each image. As shown in Table. \ref{tab: complexity_comparison }, the complexity of our proposed framework with the CBAM attention architecture does not have a significant growth in the computational efficiency compared to the competing methods FPHB \cite{Yang2019}, DeepCrack \cite{Zou2019DeepCrack:Detection}, and CrackformerII \cite{Liu2023CrackFormerSegmentation}. The inference time of our proposed framework is minimal compared to the inference time working with the competing methods FPHB \cite{Yang2019}, DeepCrack \cite{Zou2019DeepCrack:Detection}, or CrackformerII \cite{Liu2023CrackFormerSegmentation}. The number of parameters is less when our proposed framework has a pixel-level discriminator than an image-level discriminator, but the pixel-level discriminator increases the computational complexity. 

\begin{table}[!htp]\centering
\caption{ Result on CRACK500 }\label{tab: results_crack500}
\resizebox{\columnwidth}{!}{%
\begin{tabular}{cccccc}
\toprule
\text{Model}  &\text{ ODS}   & \text{ OIS} &\text{ AP} & \text{ Global Accuracy} & \text{ Mean IOU} 
\\\midrule
UNet \cite{Ronneberger2015}         &0.6072&0.6298&0.5906&0.9627 &0.6889\\
HED \cite{Xie2015}                 & 0.5864&0.5748&0.5766&0.9633&0.6841\\
FPHB  \cite{Yang2019}             & 0.6294 &0.6214&0.5821&0.9619& 0.7091\\
V-GAN (pixel) \cite{Son2017}       &0.7420&0.6872&0.5835&0.9701&0.7793\\
V-GAN (image) \cite{Son2017}       &0.5870&0.5544&0.3528&0.9491&0.6805  \\
DeepCrack \cite{Liu2019}          &0.7571&\textbf{0.7511}&\textbf{0.8219} &0.9753&0.7912 \\
Crackformer II \cite{Liu2023CrackFormerSegmentation}      &0.6563&0.6874&0.6711 &0.9670&0.7258       \\\midrule
cGAN\_LSA (pixel)              &0.7356&0.7167&0.7324 &0.9725&0.7758\\
cGAN\_CBAM (pixel)              &\textbf{0.7675}&0.7435&0.6891 &0.9762&\textbf{0.7988}\\
cGAN\_CBAM\_Ig (pixel)           &0.7638&0.7463& 0.6855&\textbf{0.9765}&0.7963 \\
cGAN\_LSA (image)            &0.7314&0.6935&0.5923&0.9731& 0.7738 \\
cGAN\_CBAM (image)            &0.7578&0.7230&0.6196&0.9759& 0.7922 \\
cGAN\_CBAM\_Ig (image)          &0.7514&0.7162&0.6216& 0.9752&0.7877

\\\midrule
\bottomrule
\end{tabular}
}
\end{table}

\begin{table}[!htp]\centering
\caption{Result on DeepCrack-DB }\label{tab: results_deepcrack}
\resizebox{\columnwidth}{!}{%
\begin{tabular}{cccccc}
\toprule
\text{Model}  &\text{ ODS}   & \text{ OIS} &\text{ AP} & \text{ Global Accuracy} & \text{ Mean IOU} 
\\\midrule
UNet \cite{Ronneberger2015}                &0.7645&0.7706& 0.8011&0.9851&0.8017 \\
HED \cite{Xie2015}                 &0.7907&0.7770&0.8429&0.9860 &0.8194 \\
FPHB \cite{Yang2019}                &0.8089 &0.7595&0.8882& 0.9867&0.8326 \\
V-GAN (pixel)  \cite{Son2017}         &0.7007  &0.7301  &0.6063&0.9801&0.7592\\
V-GAN (image)  \cite{Son2017}        &0.7053  &0.7063  &0.5237&0.9818&0.7630\\
DeepCrack \cite{Liu2019}          &0.8212  &0.8110  &0.8928 &0.9873  &0.8416\\
Crackformer II \cite{Liu2023CrackFormerSegmentation}       &0.8751  & 0.8537 & \textbf{0.9195} &0.9911&0.8844   \\\midrule
cGAN\_LSA (pixel)            &0.8662 &0.8481 & 0.8360  &0.9905 &0.8771\\
cGAN\_CBAM (pixel)            &\textbf{0.8926} &0.8759 & 0.8662  &\textbf{0.9924} &\textbf{0.8991}\\
cGAN\_CBAM\_Ig (pixel)       &0.8921&\textbf{0.8765}&0.8833&\textbf{0.9924}&0.8986 \\
cGAN\_LSA (image)            &0.8505&0.8251&0.8000&0.9894&0.8644 \\
cGAN\_CBAM (image)          &0.8760&0.8556&0.8316&0.9912&0.8851  \\
cGAN\_CBAM\_Ig (image)        &0.8014&0.7917&0.7457&0.9864&0.8273
\\\midrule
\bottomrule
\end{tabular}
}
\end{table}

\begin{table}[!htp]\centering
\caption{ Result on CFD }\label{tab: results_cfd}
\resizebox{\columnwidth}{!}{%
\begin{tabular}{cccccc}
\toprule
\text{Model}  &\text{ ODS}   & \text{ OIS} &\text{ AP} & \text{Global Accuracy} & \text{ Mean IOU} 
\\\midrule
UNet \cite{Ronneberger2015}         & 0.7131 &0.7164&0.6085&0.9910 & 0.7724         \\
HED \cite{Xie2015}                  & 0.6212&0.6359&0.5791&0.9889&0.7175               \\
FPHB   \cite{Yang2019}              & 0.6283&0.6197&0.5852&0.9875&0.7227             \\
V-GAN (pixel) \cite{Son2017}        & 0.3896 & 0.3713& 0.1718 &0.9841&0.6129 \\
V-GAN (image) \cite{Son2017}        & 0.5309& 0.5388&0.4194&0.9872&0.6737\\
DeepCrack  \cite{Liu2019}           & 0.6520& 0.6738 &0.6618 & 0.9888& 0.7358           \\
Crackformer II \cite{Liu2023CrackFormerSegmentation}       & 0.6457 & 0.6504 &0.6289 &0.9887&0.7321    \\\midrule
cGAN\_LSA (pixel)             &0.7146  & 0.7152 & 0.5598 & 0.9909&  0.7733 \\
cGAN\_CBAM (pixel)           &\textbf{0.7968} &\textbf{0.7986} & 0.6721 & \textbf{0.9935}  & \textbf{0.8278} \\
cGAN\_CBAM\_Ig (pixel)   &0.7905&0.7942& 0.6875  &0.9933 & 0.8233   \\
cGAN\_LSA (image)        &0.6703&0.6775&0.5819&0.9897&0.7465 \\
cGAN\_CBAM (image)       &0.7836&0.7842&0.6748& 0.9932&0.8186 \\
cGAN\_CBAM\_Ig (image)   &0.7669&0.7673&\textbf{0.7007}&0.9926&0.8071
\\\midrule
\bottomrule
\end{tabular}
}
\end{table}

\begin{table}[!htp]\centering
\caption{ Result on CrackLS315 }\label{tab: results_crackls315}
\resizebox{\columnwidth}{!}{%
\begin{tabular}{cccccc}
\toprule
\text{Model}  &\text{ ODS}   & \text{ OIS} &\text{ AP} & \text{ Global Accuracy} & \text{ Mean IOU} 
\\\midrule
UNet  \cite{Ronneberger2015}         &0.2452 & 0.2324 & 0.0908&0.9939 &0.5650        \\
HED  \cite{Xie2015}                 &0.0009& 0.0008 &0.0025& 0.9977&0.4991\\
FPHB   \cite{Yang2019}                &0.0119&0.0089&0.0025&0.9977&0.4989 \\
V-GAN (pixel) \cite{Son2017}        &0.1849&0.1449&0.0404&0.9977&0.5495\\
V-GAN (image) \cite{Son2017}        &0.2411 & 0.2232& 0.0982&0.9973&0.5669  \\
DeepCrack  \cite{Liu2019}          &0.3668 & 0.3534 & 0.2707&0.9977&0.6104\\
Crackformer II \cite{Liu2023CrackFormerSegmentation}         &0.3156 &0.2853& 0.1971&0.9959&0.5916 \\\midrule
cGAN\_LSA (pixel)              &0.4553  &0.4225 & 0.2300& 0.9976& 0.6461          \\
cGAN\_CBAM (pixel)             &\textbf{0.5418}&\textbf{0.5006}&\textbf{0.3520}&\textbf{0.9980}& \textbf{0.6846} \\
cGAN\_CBAM\_Ig (pixel)        &0.5322&0.4889&0.3196&\textbf{0.9980}&0.6802  \\
cGAN\_LSA (image)           &0.4024&0.3815&0.1934&0.9974&0.6245\\
cGAN\_CBAM (image) &0.4366&0.4085&0.2240 &0.9976&0.6382          \\
cGAN\_CBAM\_Ig (image)      &0.5044&0.4658&0.2944&0.9979&0.6675
\\\midrule
\bottomrule
\end{tabular}
}
\end{table}

\begin{table}[!htp]\centering
\caption{ Result on CRKWH100 }\label{tab: results_crkwh100}
\resizebox{\columnwidth}{!}{%
\begin{tabular}{cccccc}
\toprule
\text{Model}  &\text{ ODS}   & \text{ OIS} &\text{ AP} & \text{ Global Accuracy} & \text{ Mean IOU} 
\\\midrule
UNet  \cite{Ronneberger2015}         & 0.2328 &0.2306&0.0832 &0.9901&0.5578   \\
HED  \cite{Xie2015}                 & 0.0827 &0.0807&0.0365&0.9972&0.5202   \\
FPHB    \cite{Yang2019}           & 0   &0   &0.0028 &0.9972 &0.4986       \\
V-GAN (pixel) \cite{Son2017}        &0.0250 &0.0282& 0.0036 &0.9971&0.5023 \\
V-GAN (image) \cite{Son2017}        &0.3045 &0.2764&0.1563& 0.9972& 0.5881       \\
DeepCrack \cite{Liu2019}          &0.4759&0.5212&0.4501&0.9975& 0.6544\\
Crackformer II \cite{Liu2023CrackFormerSegmentation}       &0.2973&0.3056&0.1922&0.9955&0.5842 \\\midrule
cGAN\_LSA (pixel)              &0.7465&0.7736&0.6452 &0.9986&0.7971\\
cGAN\_CBAM (pixel)             &0.7981&\textbf{0.8265}& 0.7161&\textbf{0.9989}&0.8314\\
cGAN\_CBAM\_Ig (pixel)         &0.7854&0.8134&0.6969&0.9988&0.8227\\  
cGAN\_LSA (image)             &0.7768&0.8030&0.7087&0.9988&0.8169 \\
cGAN\_CBAM (image)            &0.7972&0.8246&0.7454&\textbf{0.9989}&0.8308 \\
cGAN\_CBAM\_Ig (image)         &\textbf{0.7982}&0.8215&\textbf{0.7493}&\textbf{0.9989}& \textbf{0.8315}

\\\midrule
\bottomrule
\end{tabular}
}
\end{table}

\begin{table}[!htp]\centering
\caption{ Result on CrackTree260}\label{tab: results_cracktree260}
\resizebox{\columnwidth}{!}{%
\begin{tabular}{cccccc}
\toprule
\text{Model}  &\text{ ODS}   & \text{ OIS} &\text{ AP} & \text{ Global Accuracy} & \text{ Mean IOU} 
\\\midrule
UNet  \cite{Ronneberger2015}         &0.3608&0.3668 & 0.1330&0.9917 &0.6035\\
HED  \cite{Xie2015}                 &0.1465&0.1331&0.0635& 0.9954&0.5372 \\
FPHB    \cite{Yang2019}               & 0.1388&0.1465&0.0418& 0.9951& 0.5292 \\
V-GAN (pixel) \cite{Son2017}        &0.3440   &0.3420  &0.1330  &0.9951 &0.6009          \\
V-GAN (image) \cite{Son2017}        &0.5278  & 0.5418 &0.3321 &0.9956& 0.6771 \\
DeepCrack  \cite{Liu2019}         &0.4905  &0.4991 &0.4627 &0.9957&0.6600             \\
Crackformer II \cite{Liu2023CrackFormerSegmentation}        &0.4936  & 0.4922& 0.4055& 0.9950&0.6607      \\\midrule
cGAN\_LSA (pixel)              &0.8341 &0.8426 & 0.7677 &0.9984 &0.8569          \\
cGAN\_CBAM (pixel)             &0.8888 &0.9085 & \textbf{0.8671}&0.9989 &0.8994        \\
cGAN\_CBAM\_Ig (pixel)        &\textbf{0.8912}&\textbf{0.9123}&0.8435& \textbf{0.9990}&\textbf{0.9014}              \\
cGAN\_LSA (image)              &0.7876&0.7951&0.7041&0.9980&0.8238 \\
cGAN\_CBAM (image)             &0.6187&0.6316&0.4511&0.9966&0.7222 \\
cGAN\_CBAM\_Ig (image)         &0.7142&0.7371&0.6865& 0.9974&0.7764         
\\\midrule
\bottomrule
\end{tabular}
}
\end{table}

\begin{table}[!htp]\centering
\caption{ Alation study of loss function on CrackLS315}\label{tab: results_tversky_abla}
\resizebox{\columnwidth}{!}{%
\begin{tabular}{cccccc}
\toprule
\text{Model}  &\text{ ODS}   & \text{ OIS} &\text{ AP} & \text{ Global Accuracy} & \text{ Mean IOU} 
\\\midrule

cGAN\_CBAM\_Ig (pixel)          &0.5322&0.4889&0.3196&0.9980&0.6802  \\
cGAN\_CBAM\_Ig (w/o Side loss)  &0.5251&0.4823&0.3172&0.9980&0.6769 \\
cGAN\_CBAM\_Ig (w/o Side loss + Tversky loss) &0.5116&0.4734&0.2941& 0.9977&0.6703

\\\midrule
\bottomrule
\end{tabular}
}
\end{table}

\begin{table}[!htp]\centering
\caption{ Alation study of loss function on DeepCrack-DB }\label{tab: results_deepcrack_abla}
\resizebox{\columnwidth}{!}{%
\begin{tabular}{cccccc}
\toprule
\text{Model}  &\text{ ODS}   & \text{ OIS} &\text{ AP} & \text{ Global Accuracy} & \text{ Mean IOU} 
\\\midrule

cGAN\_CBAM\_Ig (pixel)          &0.8921&0.8765&0.8833&0.9924&0.8986 \\
cGAN\_CBAM\_Ig (w/o Side loss)  &0.8420&0.8066&0.7110& 0.9881&0.8575  \\
cGAN\_CBAM\_Ig (w/o Side loss + Tversky loss) &0.8161 &0.7772&0.7361 &0.9865 &0.8377

\\\midrule
\bottomrule
\end{tabular}
}
\end{table}

\begin{figure*}
\centering
\footnotesize
\renewcommand{\tabcolsep}{1pt} 
\renewcommand{\arraystretch}{0.2} 
\begin{tabular}{ccc}  
    \includegraphics[width=0.3\linewidth]{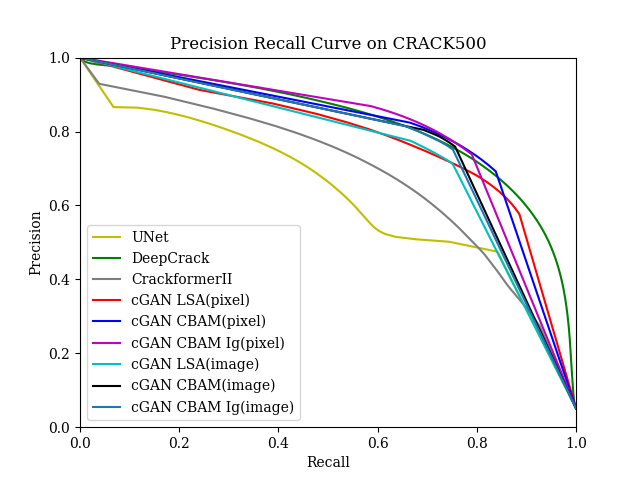} & 
    \includegraphics[width=0.3\linewidth]{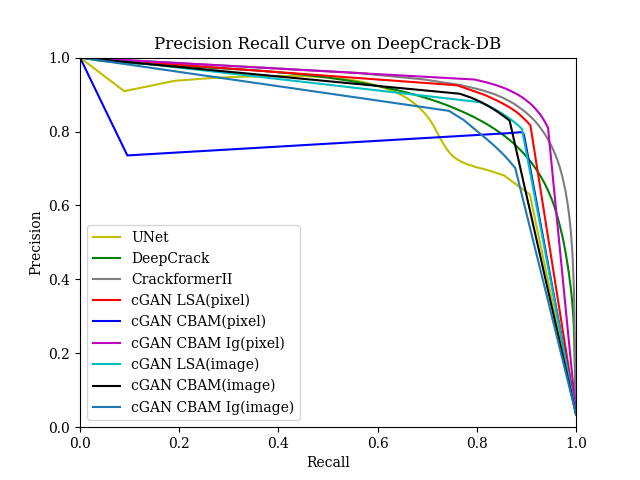} &
    \includegraphics[width=0.3\linewidth] 
    {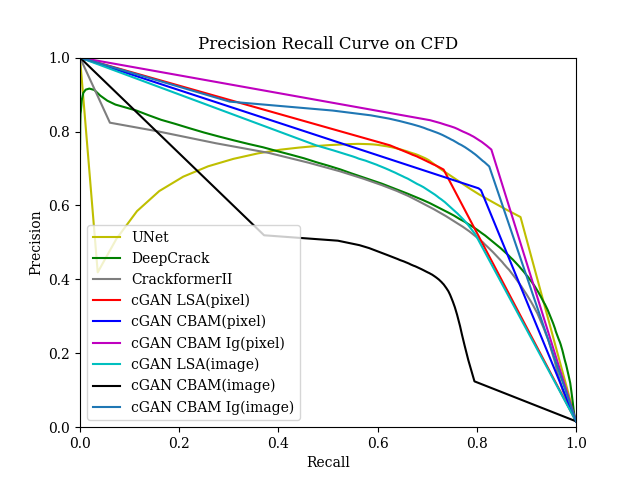}
    \\
    \includegraphics[width=0.3\linewidth]
    {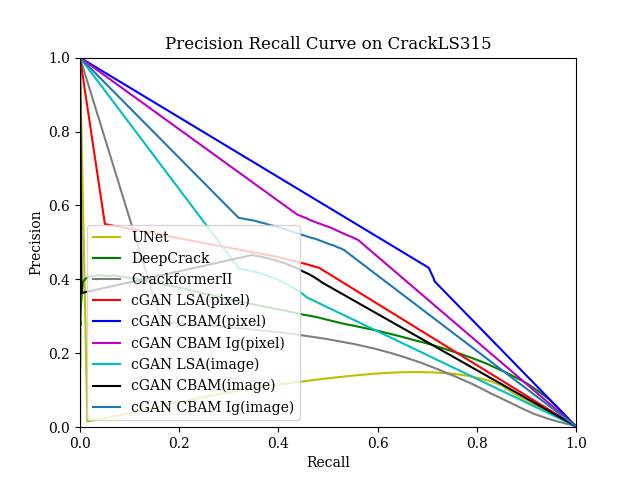}& 
    \includegraphics[width=0.3\linewidth] 
    {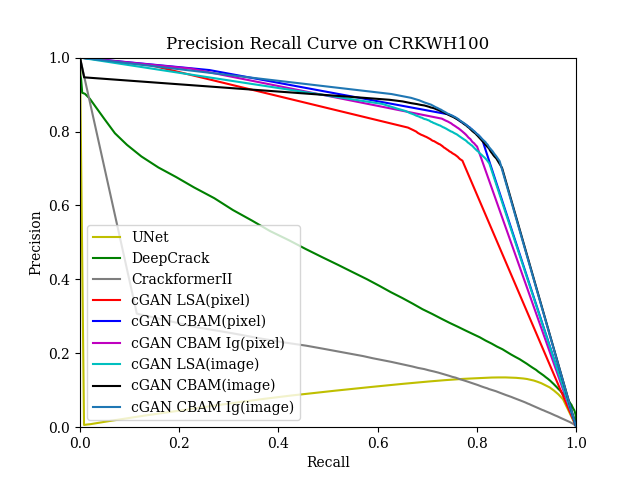} &
    \includegraphics[width=0.3\linewidth]
    {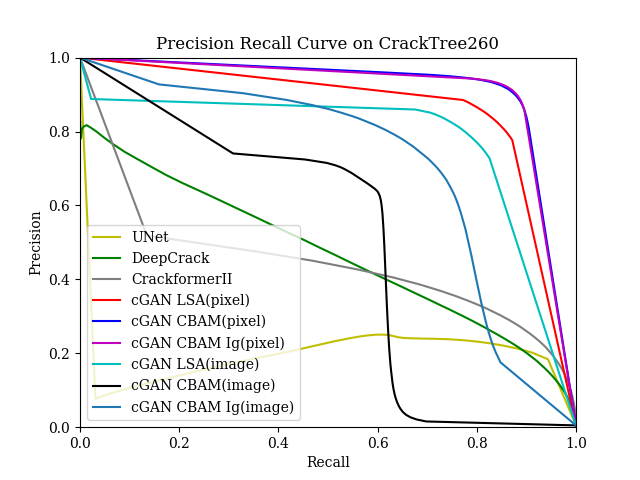}   
    \\  
\end{tabular}
\caption{ \footnotesize {\textbf{Precision and Recall Curves.} }
}
\label{fig:pre_rec}
\end{figure*}

\begin{figure*}
\centering
\footnotesize
\renewcommand{\tabcolsep}{1pt} 
\renewcommand{\arraystretch}{0.2} 
\begin{tabular}{ccccccccccccc}
    \raisebox{2.5\normalbaselineskip}[0pt][0pt]{\rotatebox[origin=c]{0}{(a)}} &  
    \includegraphics[width=0.08\linewidth]{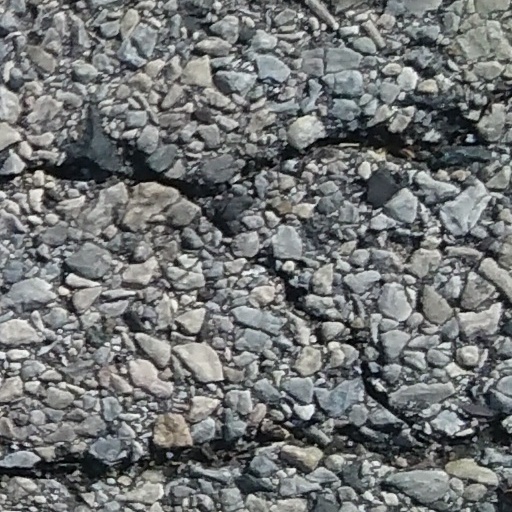} & \includegraphics[width=0.08\linewidth]{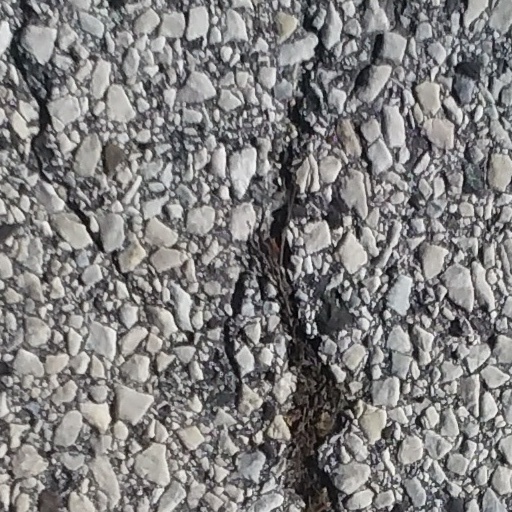} & 
    \includegraphics[width=0.08\linewidth]{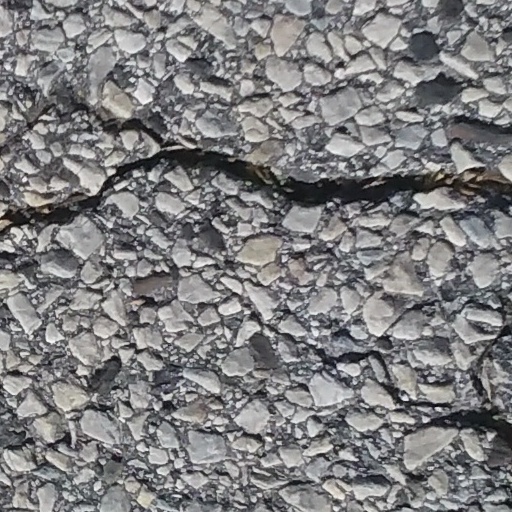} & 
    \includegraphics[width=0.08\linewidth]{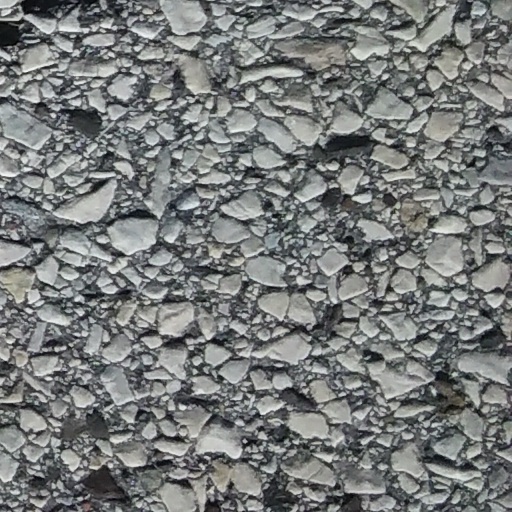} & 
    \includegraphics[width=0.08\linewidth]{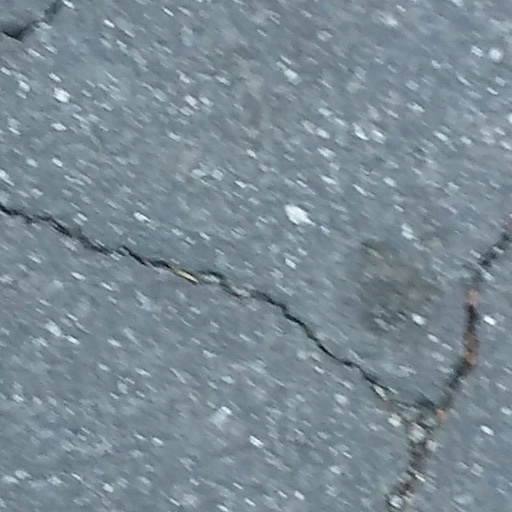} &
    \includegraphics[width=0.08\linewidth]{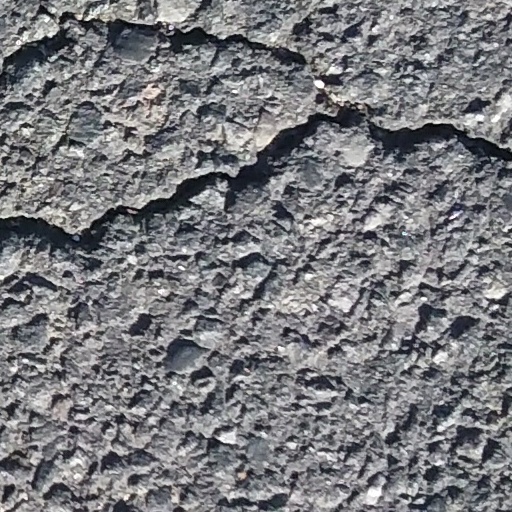} & 
    \includegraphics[width=0.08\linewidth]{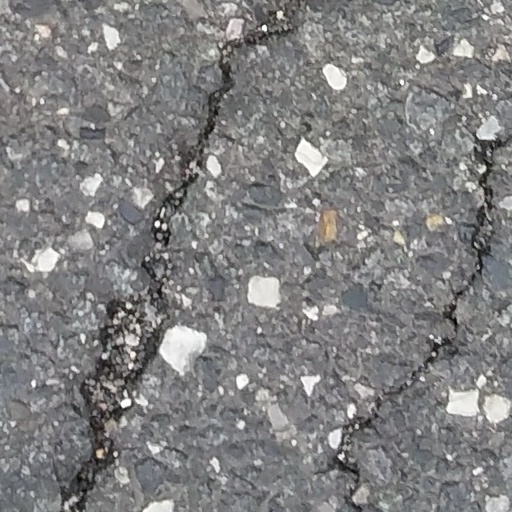} &
    \includegraphics[width=0.08\linewidth]{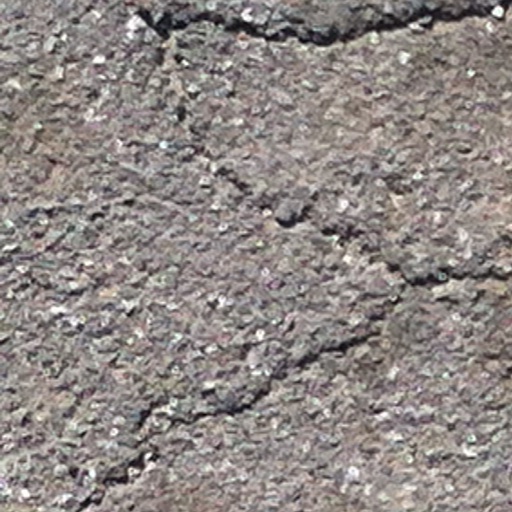} &
    \includegraphics[width=0.08\linewidth]{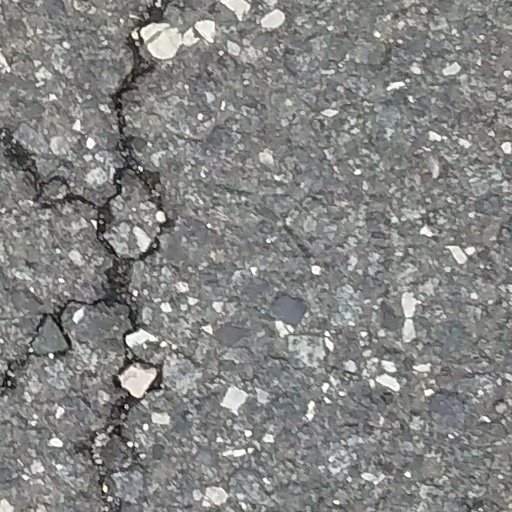} &
    \includegraphics[width=0.08\linewidth]{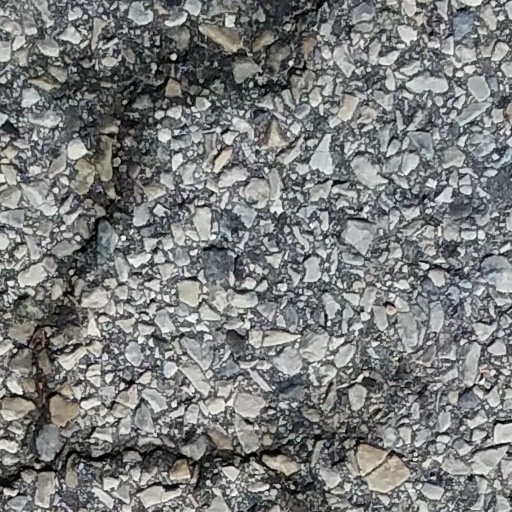} &
    \includegraphics[width=0.08\linewidth]{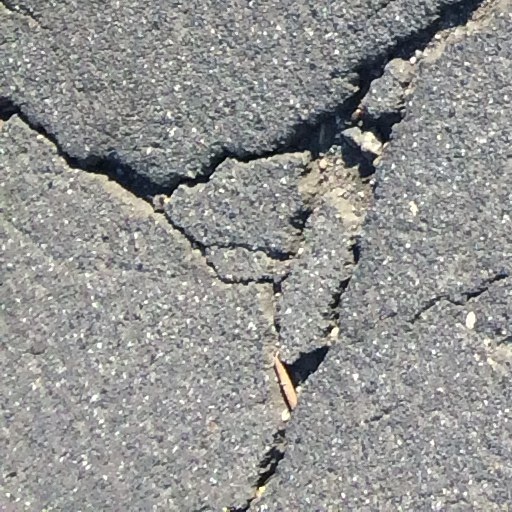} 
    \\

    \raisebox{2.5\normalbaselineskip}[0pt][0pt]{\rotatebox[origin=c]{0}{(b)}} &  
    \includegraphics[width=0.08\linewidth]{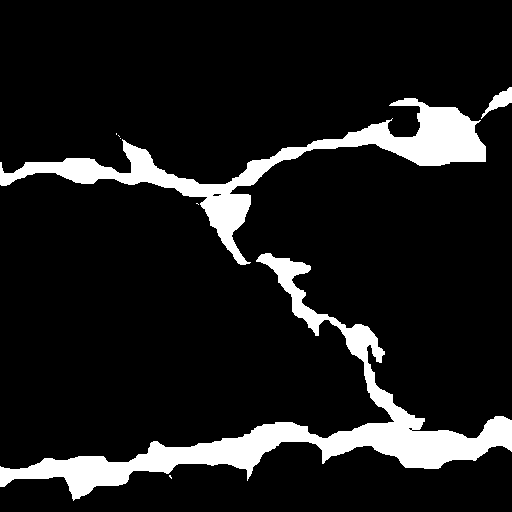} &
    \includegraphics[width=0.08\linewidth]{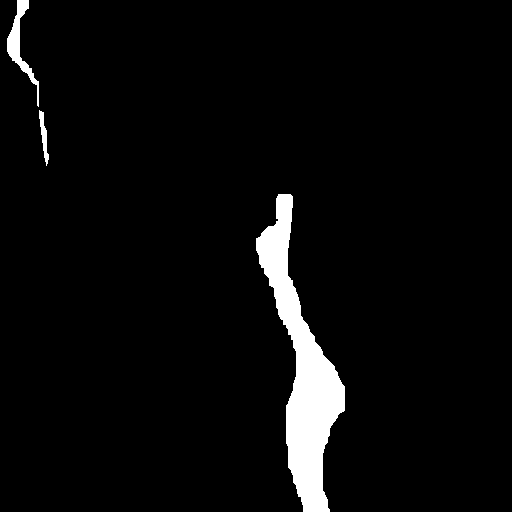} & 
    \includegraphics[width=0.08\linewidth]{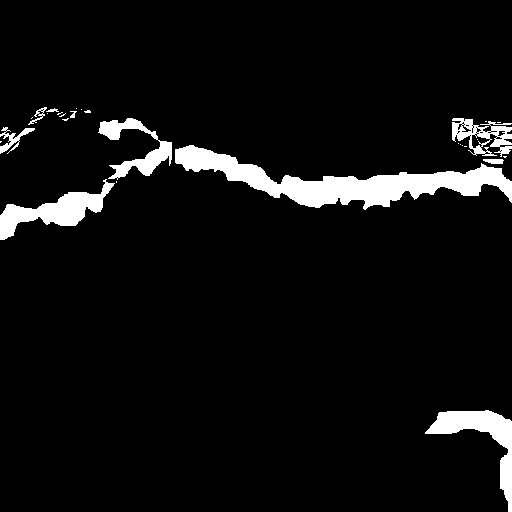} & 
    \includegraphics[width=0.08\linewidth]{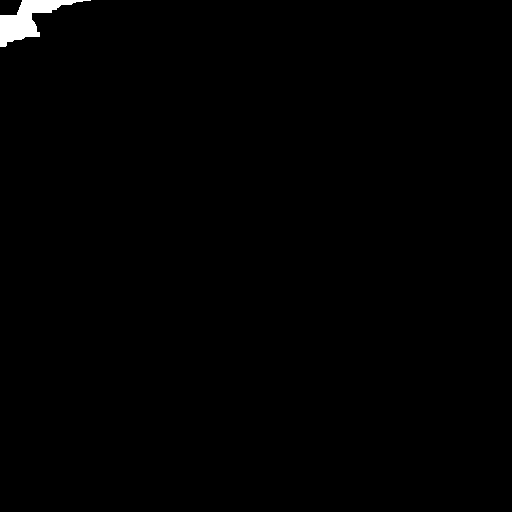} & 
    \includegraphics[width=0.08\linewidth]{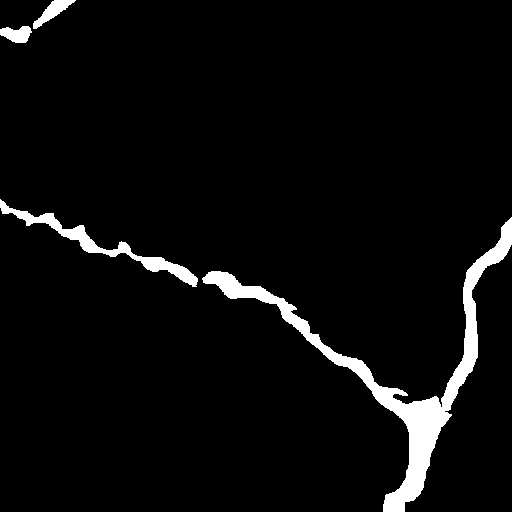} &
    \includegraphics[width=0.08\linewidth]{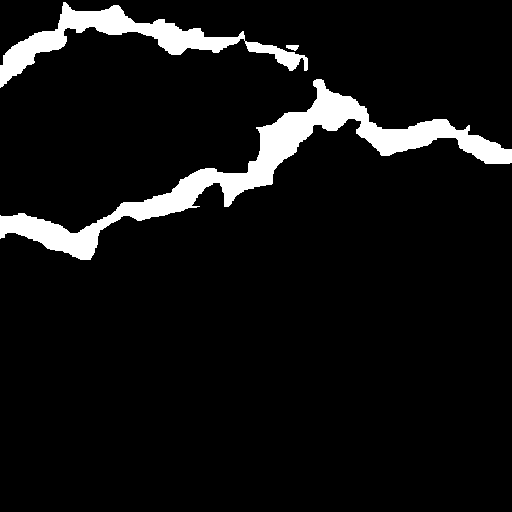} & 
    \includegraphics[width=0.08\linewidth]{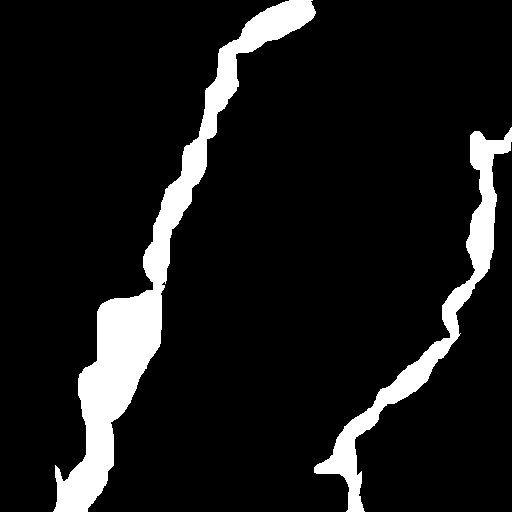} &
    \includegraphics[width=0.08\linewidth]{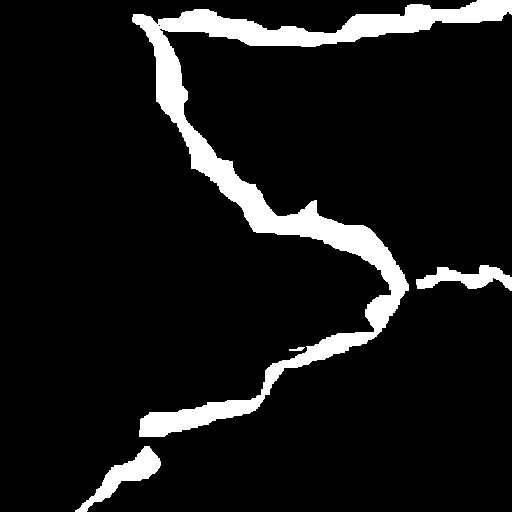} &
    \includegraphics[width=0.08\linewidth]{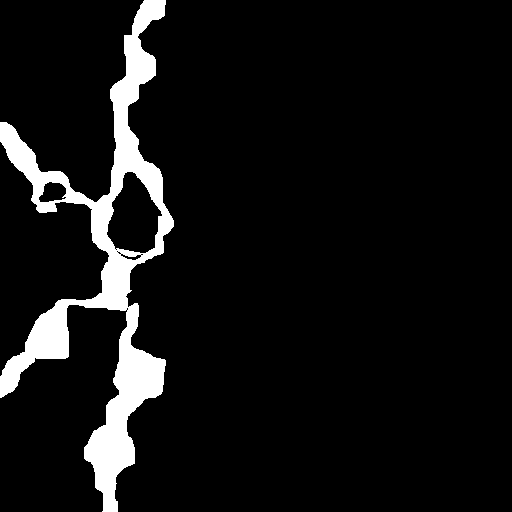} &
    \includegraphics[width=0.08\linewidth]{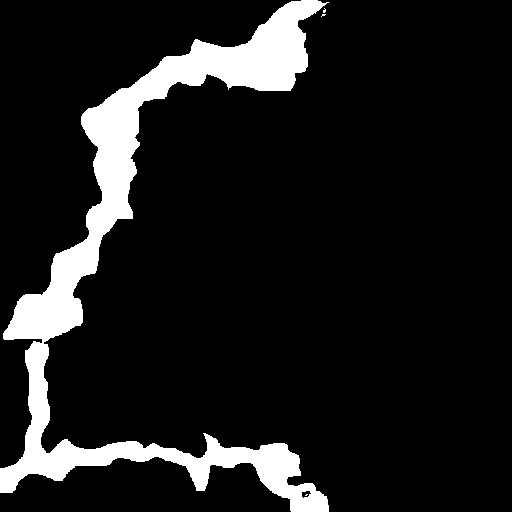} &
    \includegraphics[width=0.08\linewidth]{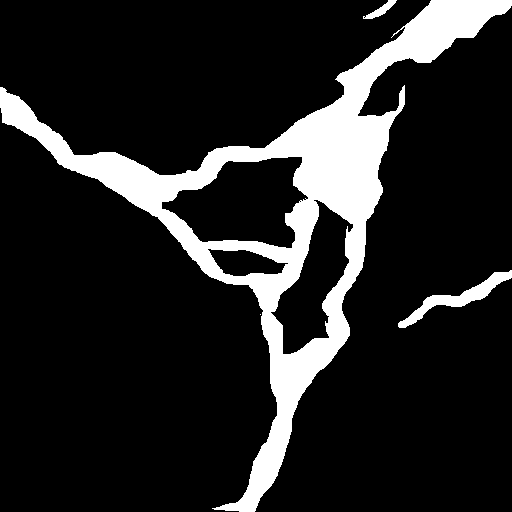} 
    \\

    \raisebox{2.5\normalbaselineskip}[0pt][0pt]{\rotatebox[origin=c]{0}{(c)}} &  
    \includegraphics[width=0.08\linewidth]{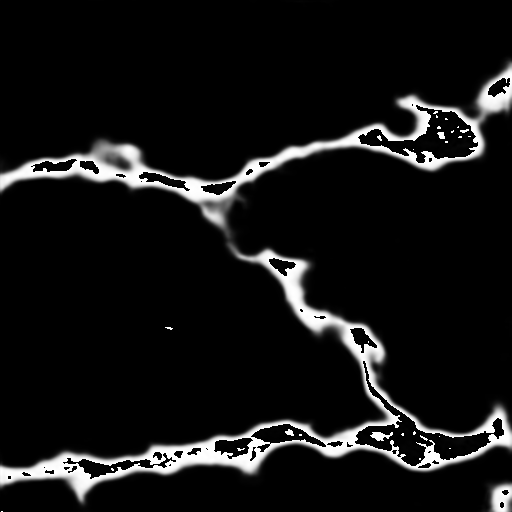} &
    \includegraphics[width=0.08\linewidth]{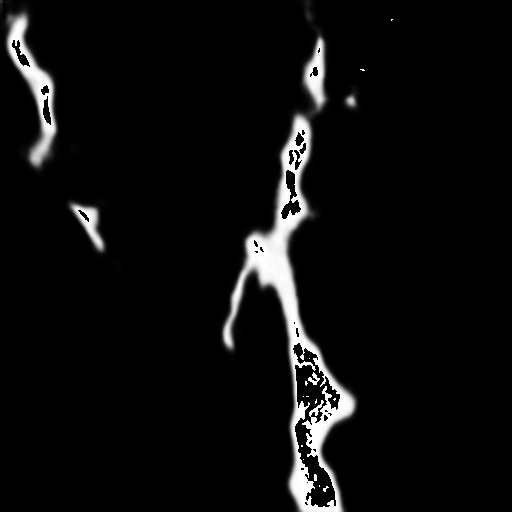} & 
    \includegraphics[width=0.08\linewidth]{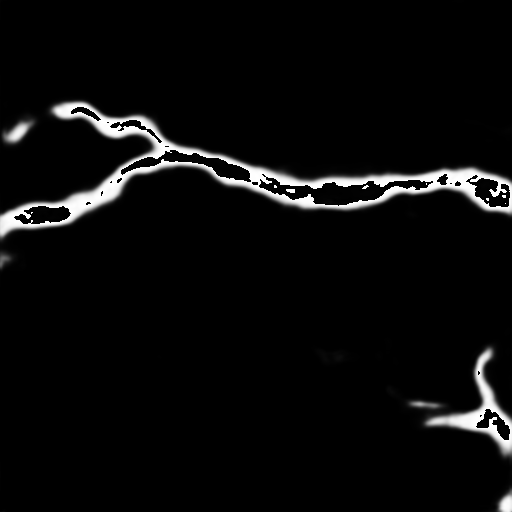} & 
    \includegraphics[width=0.08\linewidth]{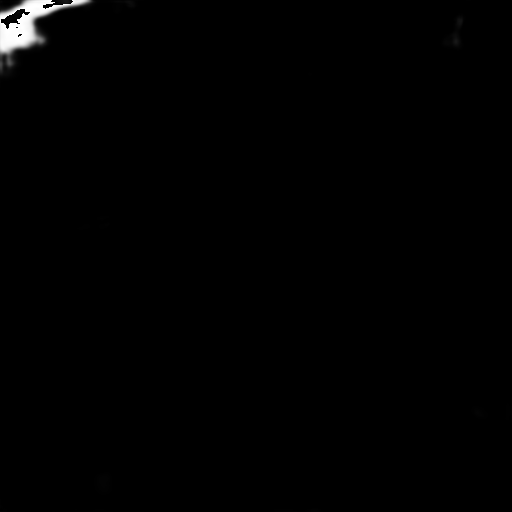} & 
    \includegraphics[width=0.08\linewidth]{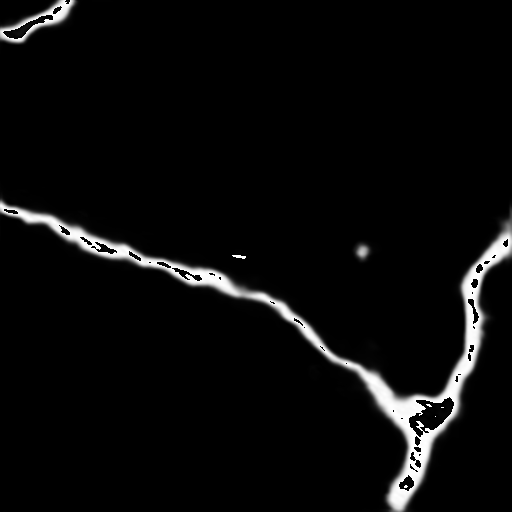} &
    \includegraphics[width=0.08\linewidth]{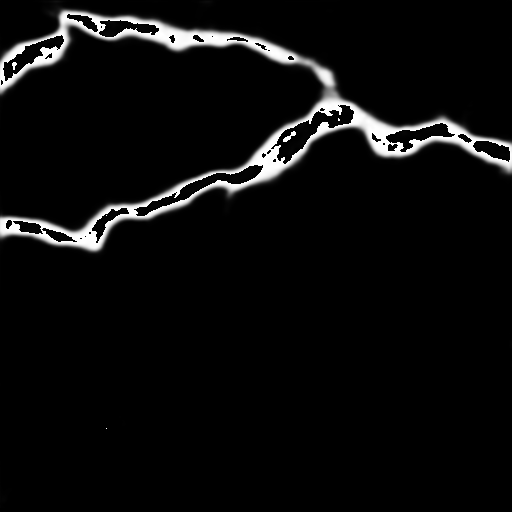} & 
    \includegraphics[width=0.08\linewidth]{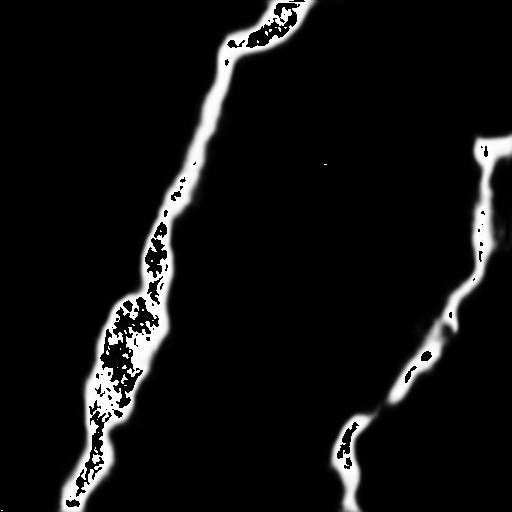} &
    \includegraphics[width=0.08\linewidth]{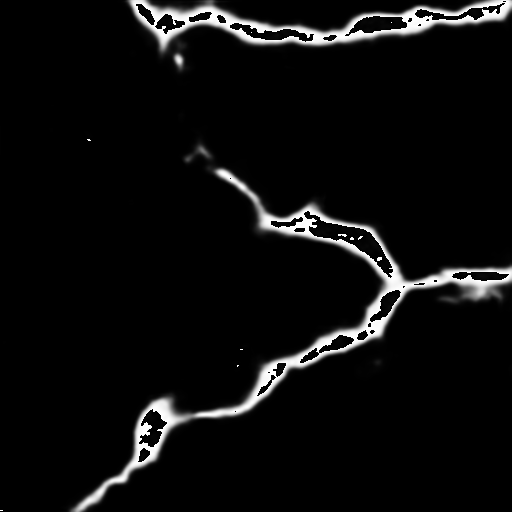} &
    \includegraphics[width=0.08\linewidth]{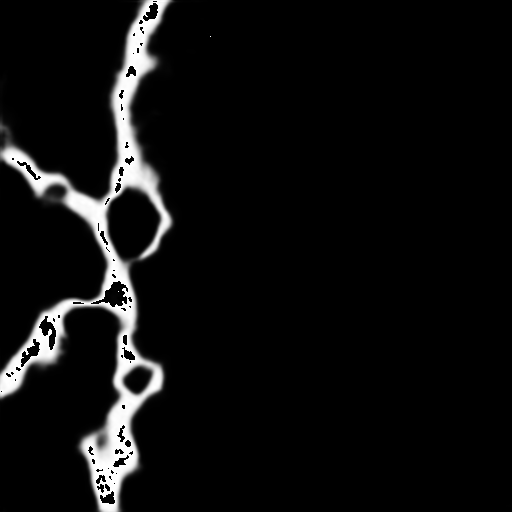}&
    \includegraphics[width=0.08\linewidth]{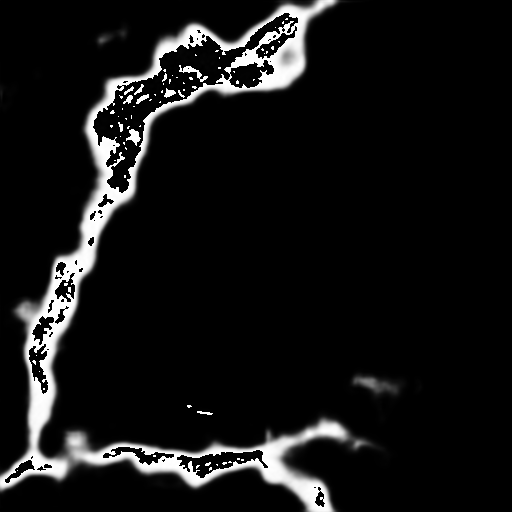} &
    \includegraphics[width=0.08\linewidth]{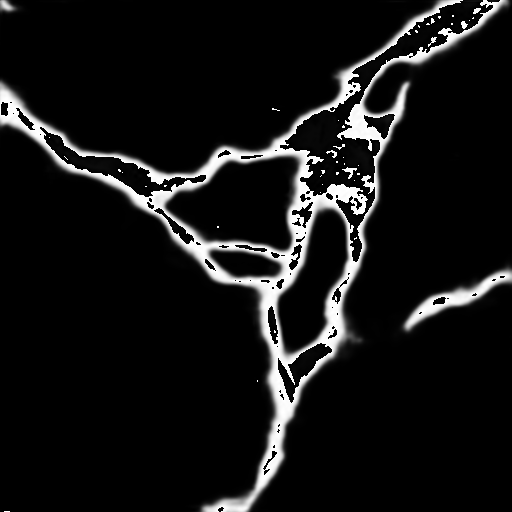}
    \\

    \raisebox{2.5\normalbaselineskip}[0pt][0pt]{\rotatebox[origin=c]{0}{(d)}} &  
    \includegraphics[width=0.08\linewidth]{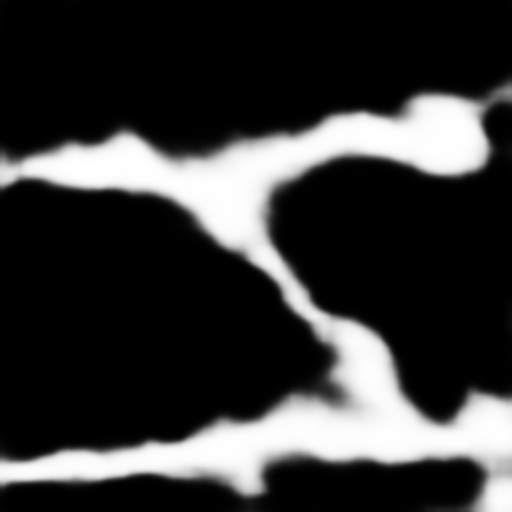} &
    \includegraphics[width=0.08\linewidth]{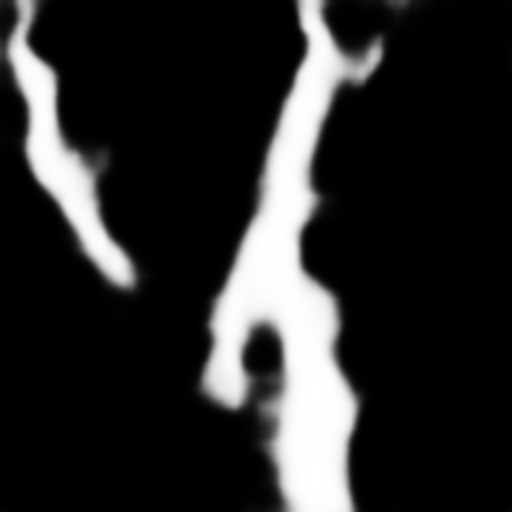} & 
    \includegraphics[width=0.08\linewidth]{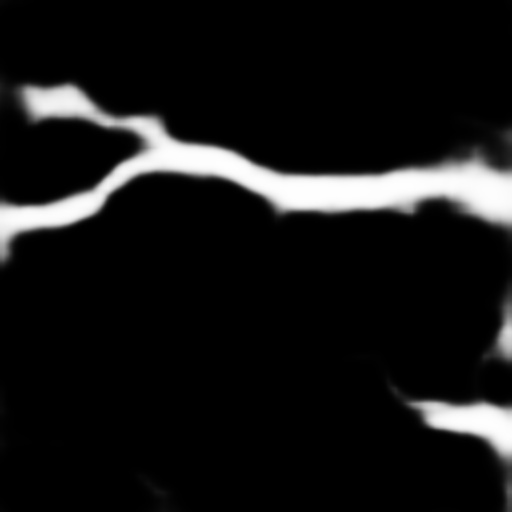} & 
    \includegraphics[width=0.08\linewidth]{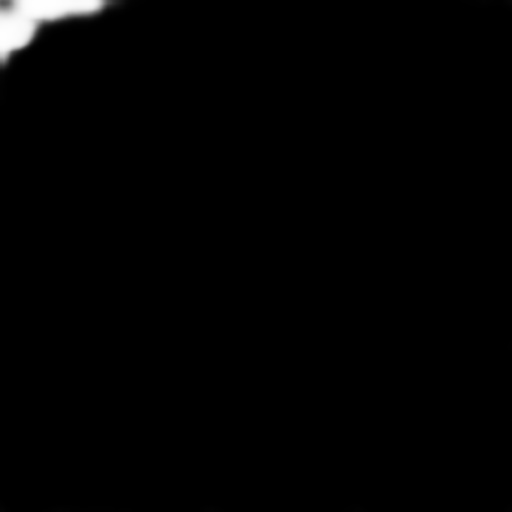} & 
    \includegraphics[width=0.08\linewidth]{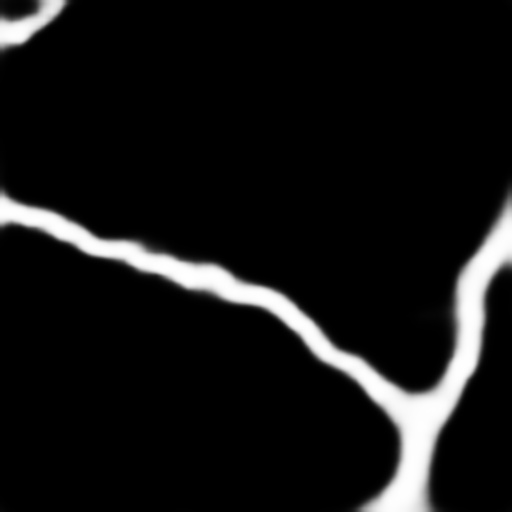} &
    \includegraphics[width=0.08\linewidth]{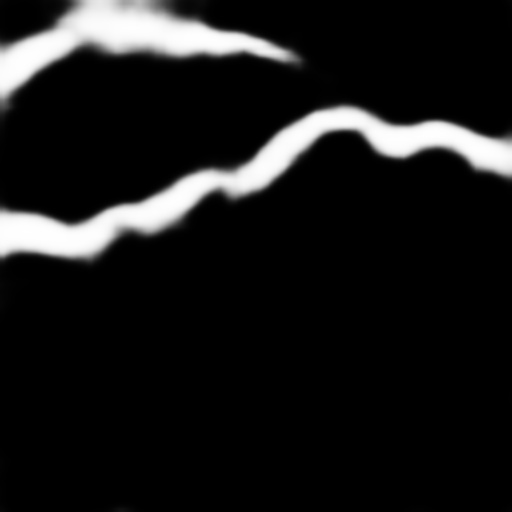} & 
    \includegraphics[width=0.08\linewidth]{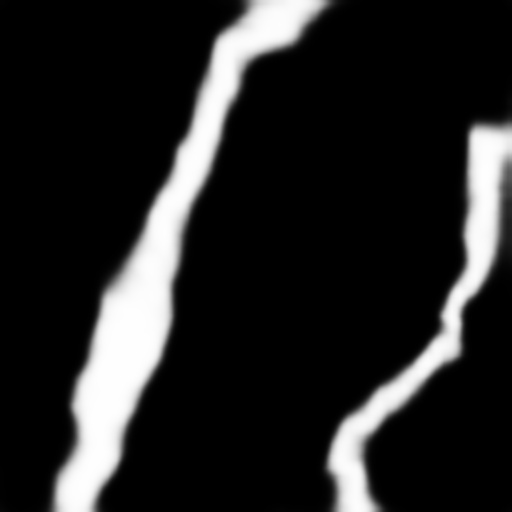} &
    \includegraphics[width=0.08\linewidth]{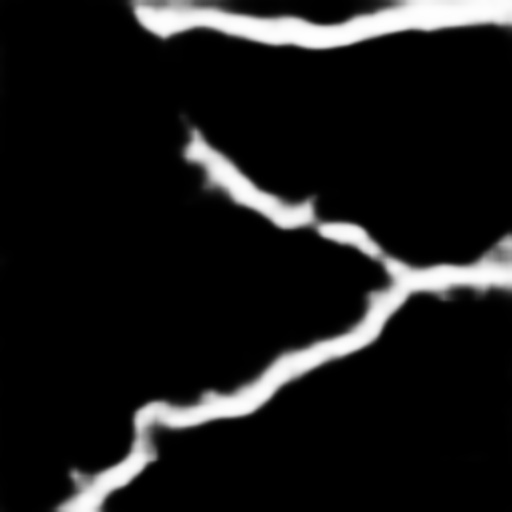} &
    \includegraphics[width=0.08\linewidth]{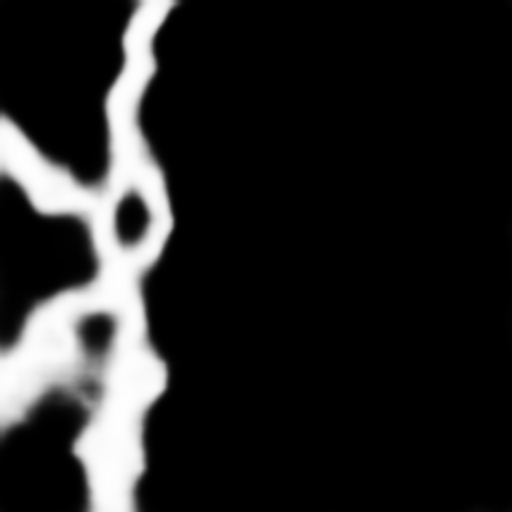} &
    \includegraphics[width=0.08\linewidth]{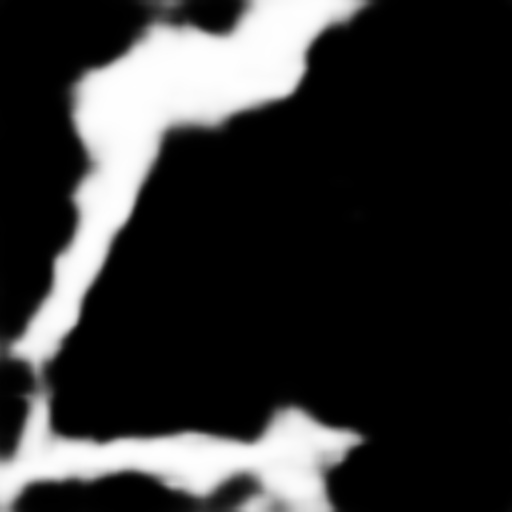} &
    \includegraphics[width=0.08\linewidth]{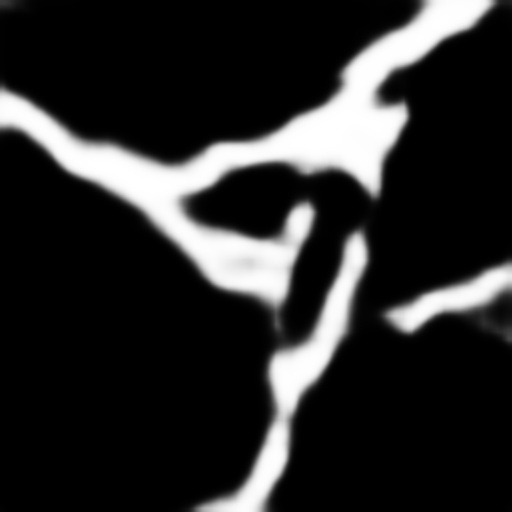} 
    \\

    \raisebox{2.5\normalbaselineskip}[0pt][0pt]{\rotatebox[origin=c]{0}{(e)}} &  
    \includegraphics[width=0.08\linewidth]{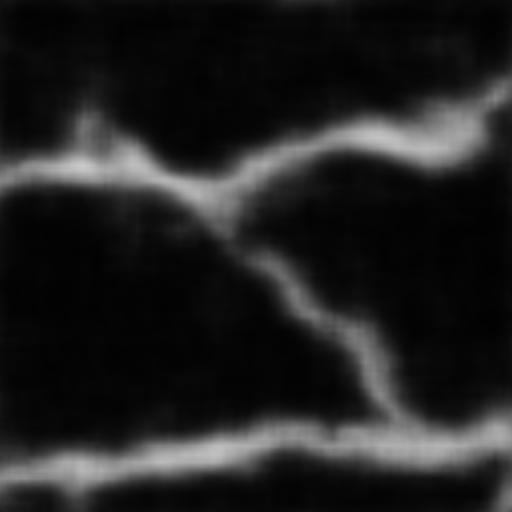} &
    \includegraphics[width=0.08\linewidth]{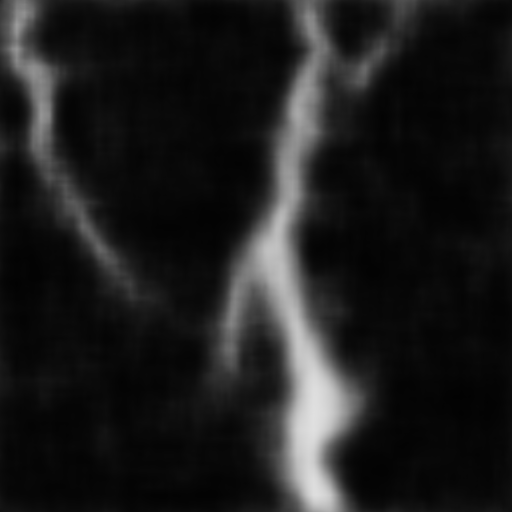} & 
    \includegraphics[width=0.08\linewidth]{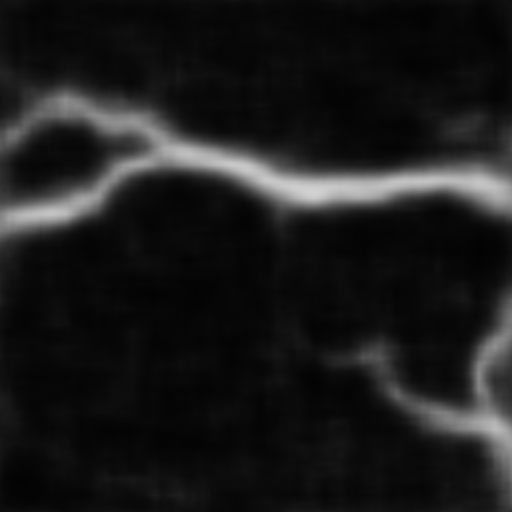} & 
    \includegraphics[width=0.08\linewidth]{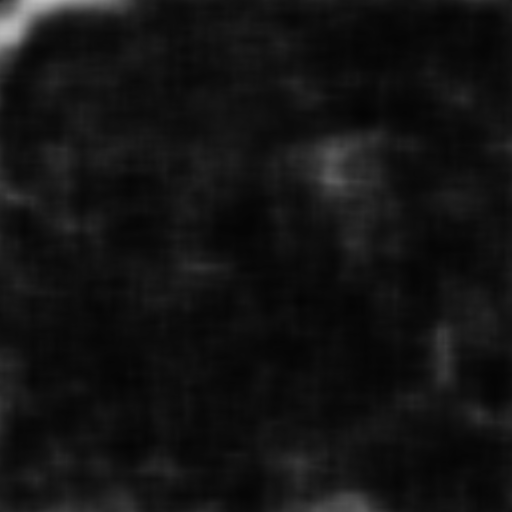} & 
    \includegraphics[width=0.08\linewidth]{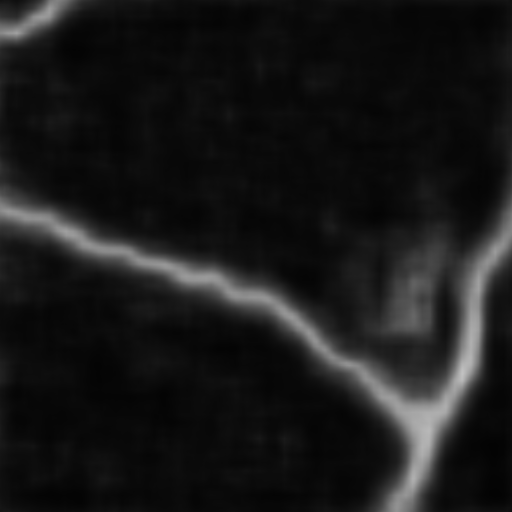} &
    \includegraphics[width=0.08\linewidth]{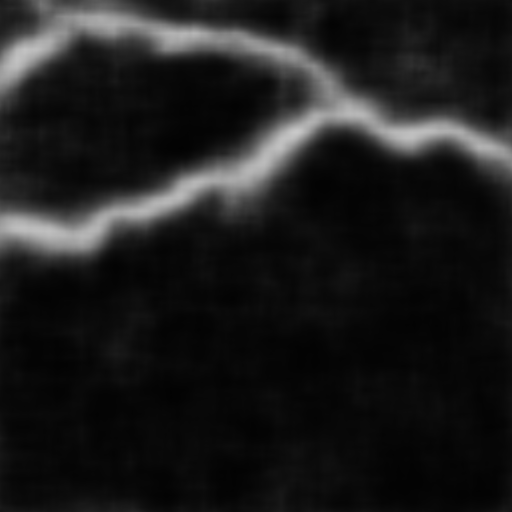} & 
    \includegraphics[width=0.08\linewidth]{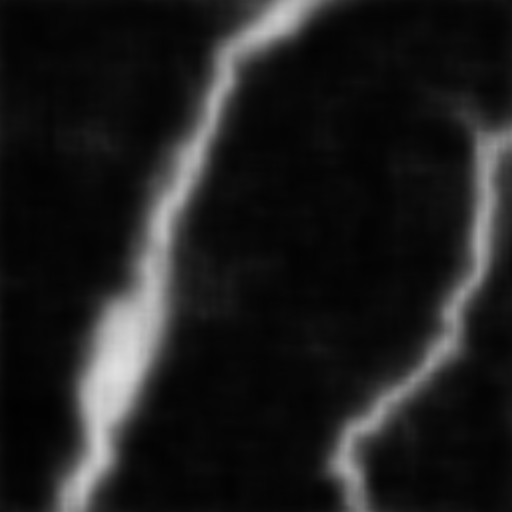} &
    \includegraphics[width=0.08\linewidth]{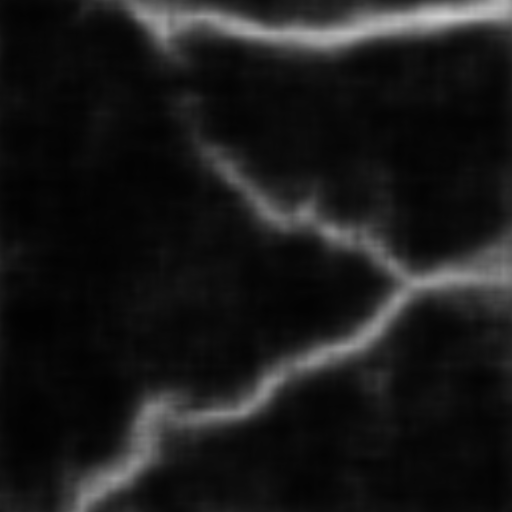} &
    \includegraphics[width=0.08\linewidth]{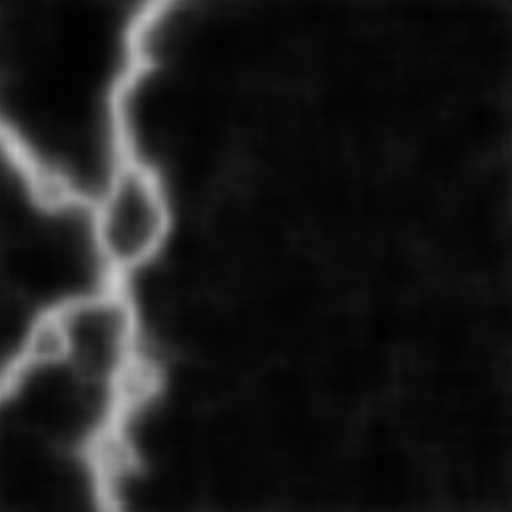}& 
    \includegraphics[width=0.08\linewidth]{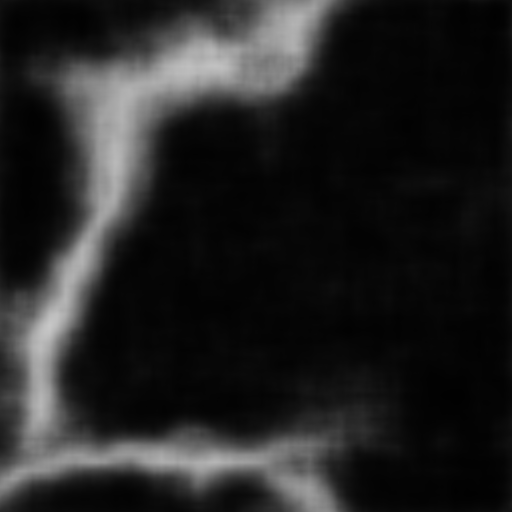} &
    \includegraphics[width=0.08\linewidth]{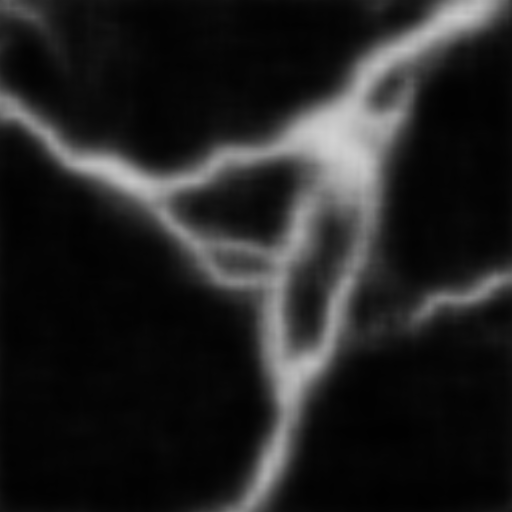}
    \\

    \raisebox{2.5\normalbaselineskip}[0pt][0pt]{\rotatebox[origin=c]{0}{(f)}} &  
    \includegraphics[width=0.08\linewidth]{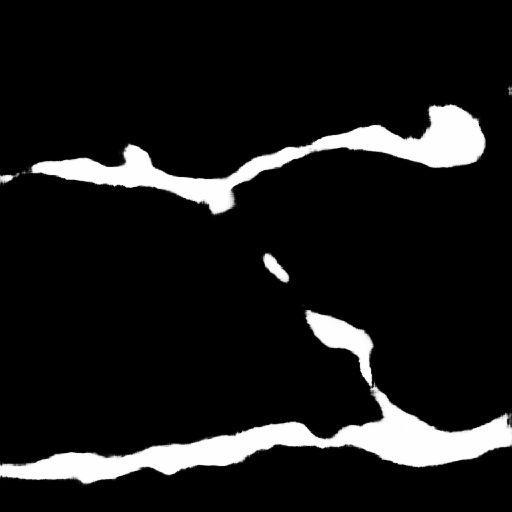} &
    \includegraphics[width=0.08\linewidth]{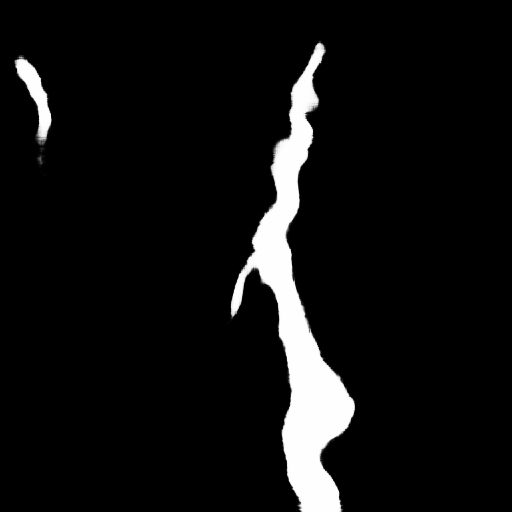} & 
    \includegraphics[width=0.08\linewidth]{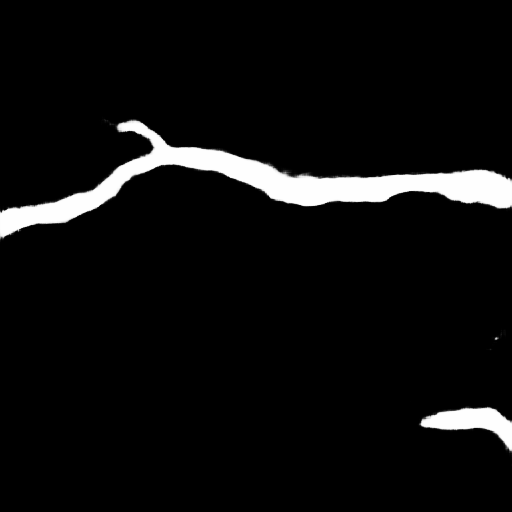} & 
    \includegraphics[width=0.08\linewidth]{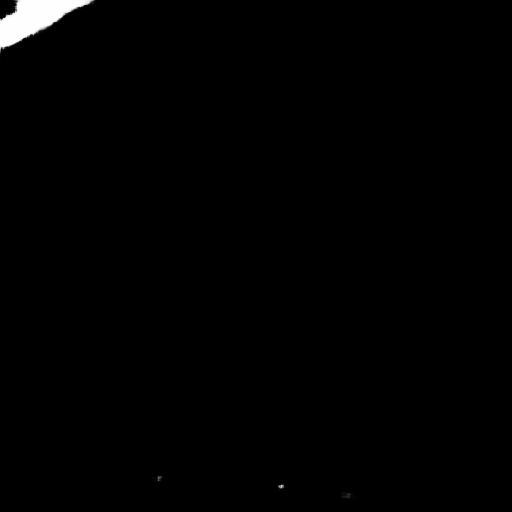} & 
    \includegraphics[width=0.08\linewidth]{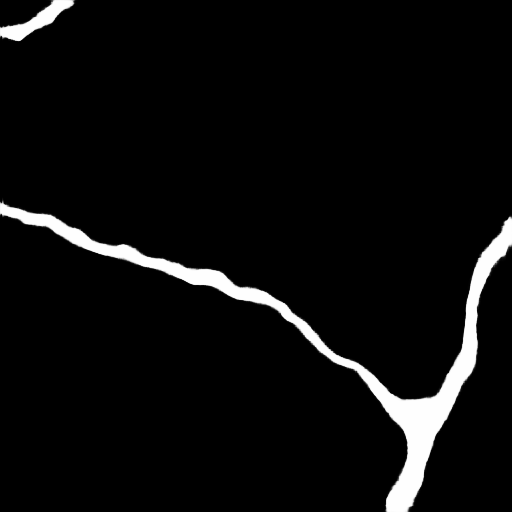} &
    \includegraphics[width=0.08\linewidth]{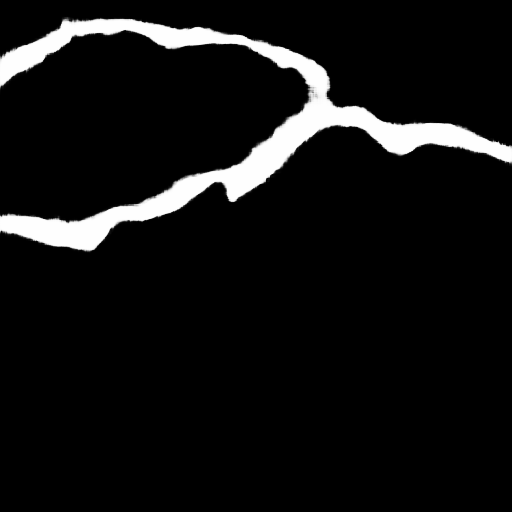} & 
    \includegraphics[width=0.08\linewidth]{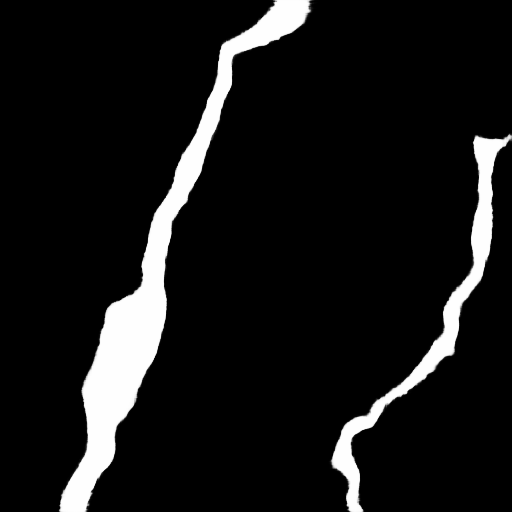} &
    \includegraphics[width=0.08\linewidth]{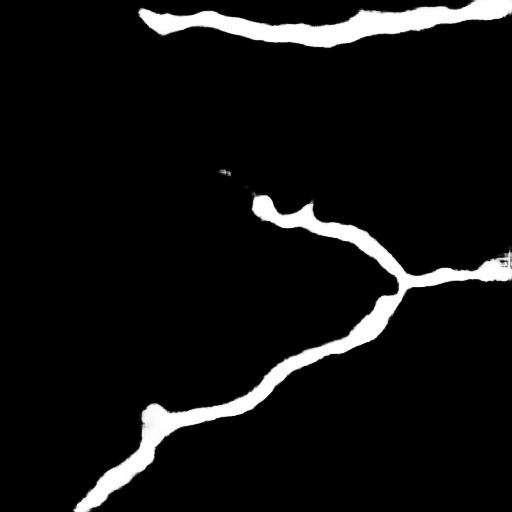} &
    \includegraphics[width=0.08\linewidth]{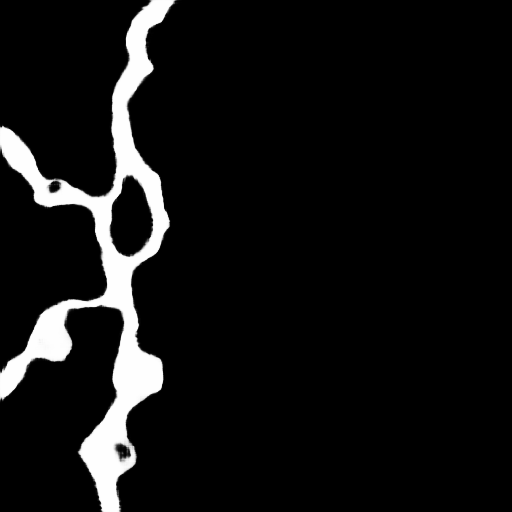} &
    \includegraphics[width=0.08\linewidth]{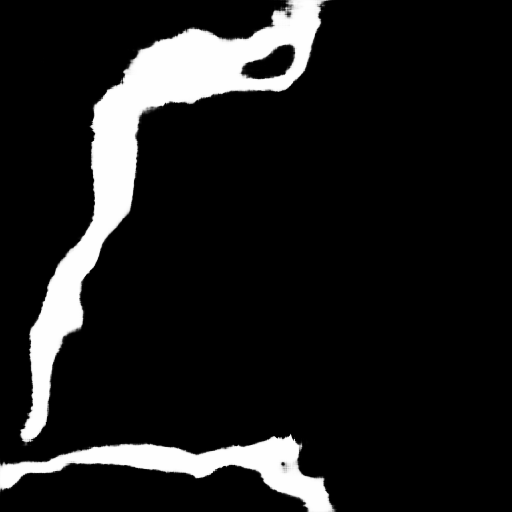} &
    \includegraphics[width=0.08\linewidth]{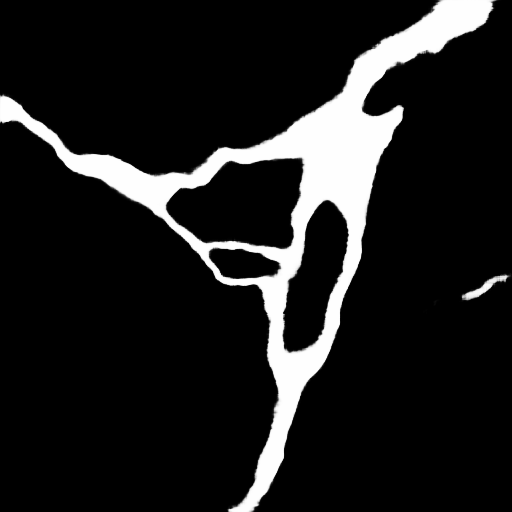} 
    \\

    \raisebox{2.5\normalbaselineskip}[0pt][0pt]{\rotatebox[origin=c]{0}{(g)}} &  
    \includegraphics[width=0.08\linewidth]{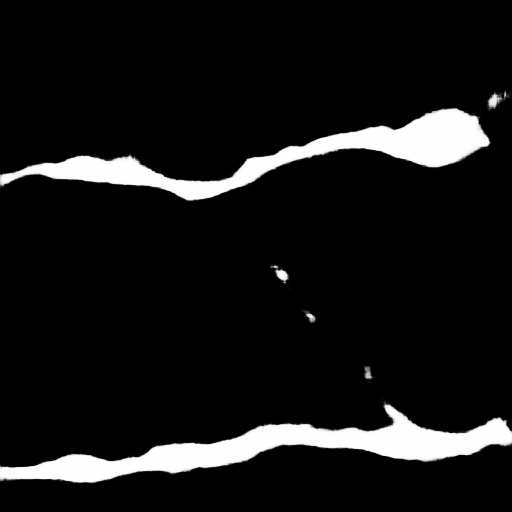} &
    \includegraphics[width=0.08\linewidth]{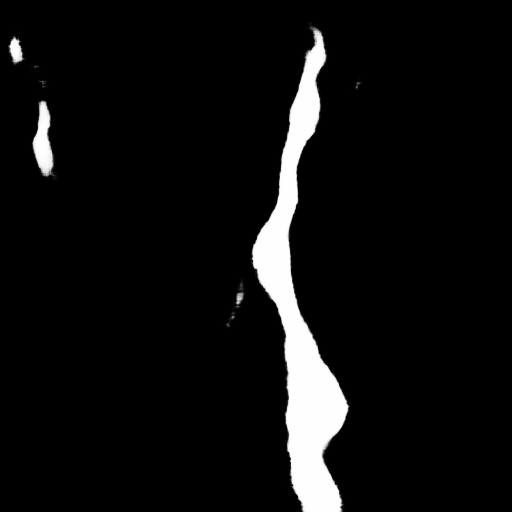} & 
    \includegraphics[width=0.08\linewidth]{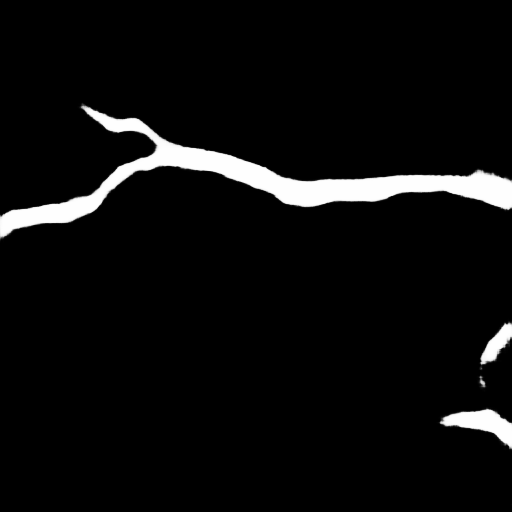} & 
    \includegraphics[width=0.08\linewidth]{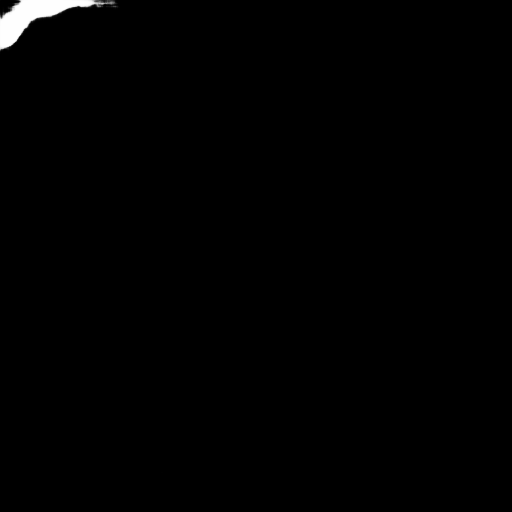} & 
    \includegraphics[width=0.08\linewidth]{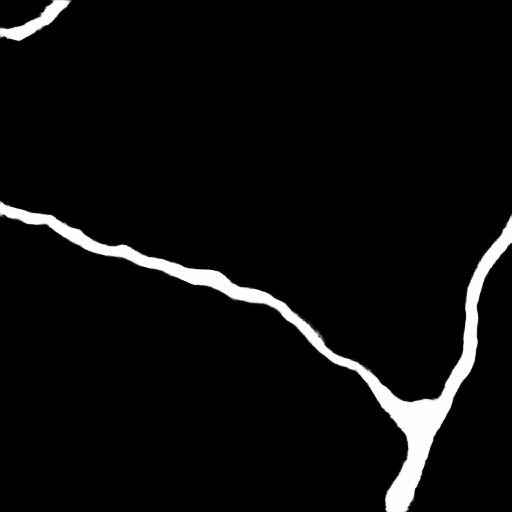} &
    \includegraphics[width=0.08\linewidth]{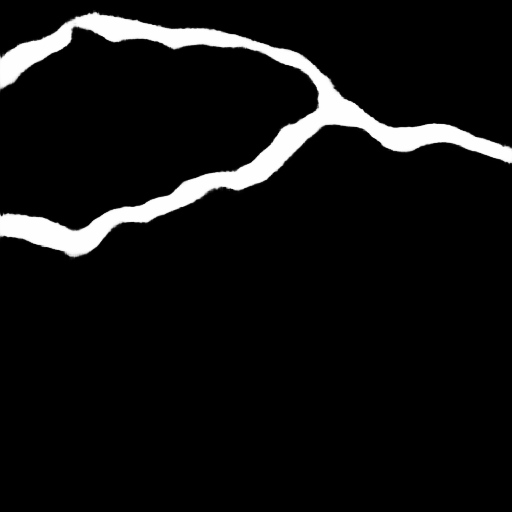} & 
    \includegraphics[width=0.08\linewidth]{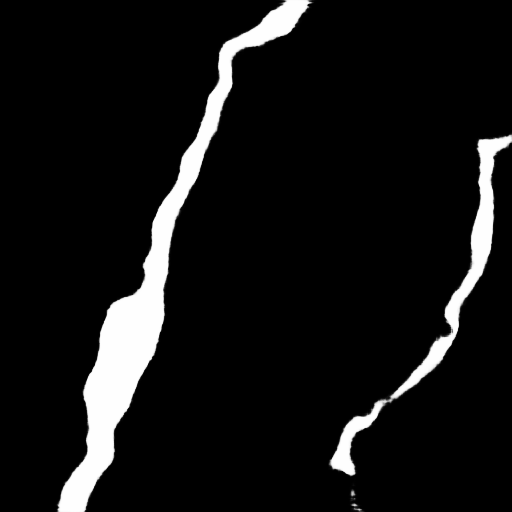} &
    \includegraphics[width=0.08\linewidth]{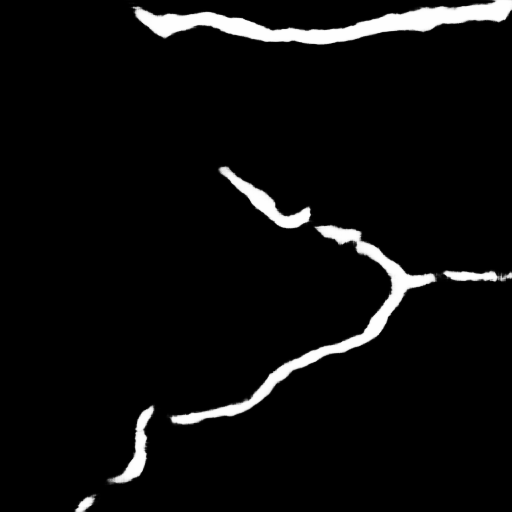} &
    \includegraphics[width=0.08\linewidth]{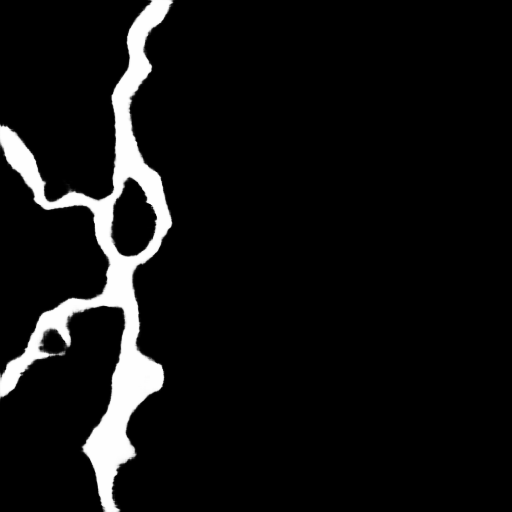} &
    \includegraphics[width=0.08\linewidth]{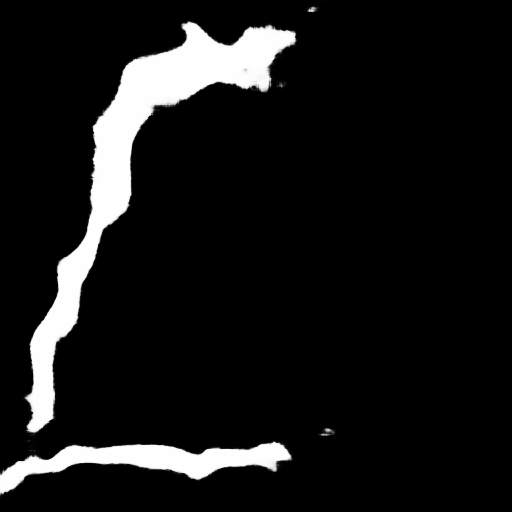} &
    \includegraphics[width=0.08\linewidth]{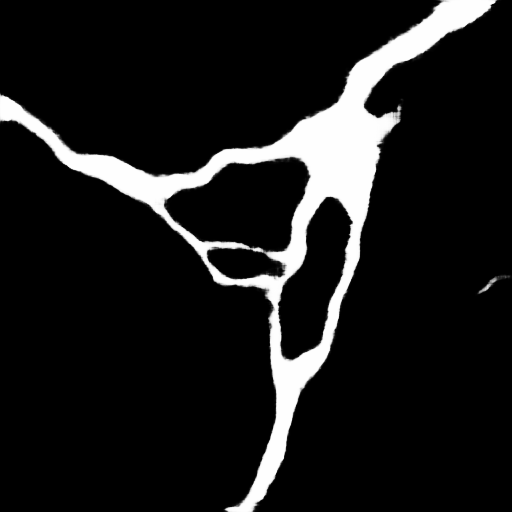}
    
    \\

    \raisebox{2.5\normalbaselineskip}[0pt][0pt]{\rotatebox[origin=c]{0}{(h)}} &  
    \includegraphics[width=0.08\linewidth]{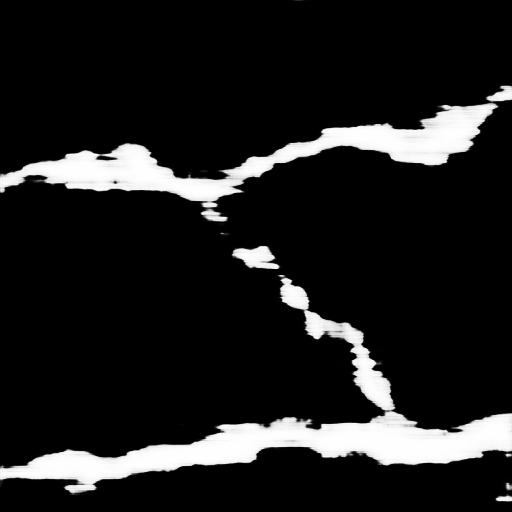} &
    \includegraphics[width=0.08\linewidth]{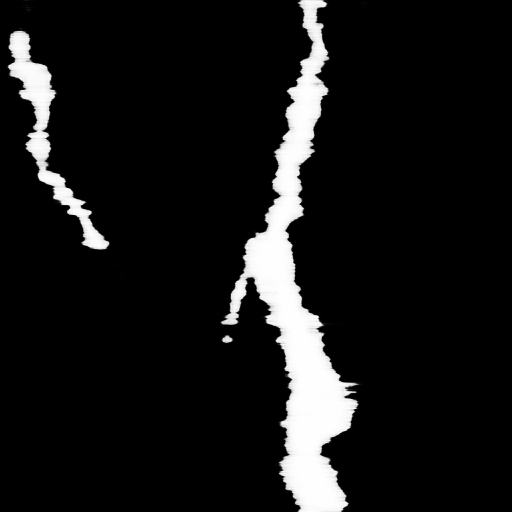} & 
    \includegraphics[width=0.08\linewidth]{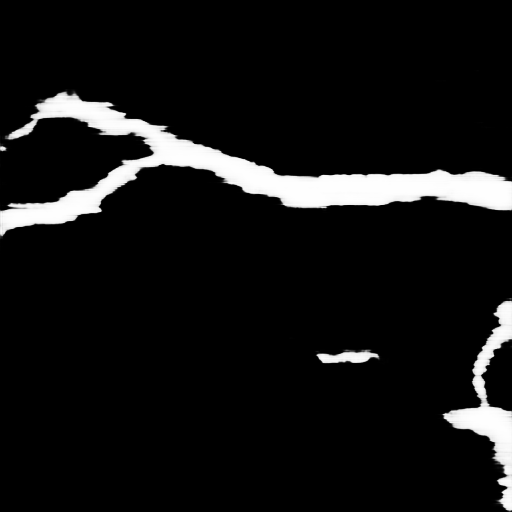} & 
    \includegraphics[width=0.08\linewidth]{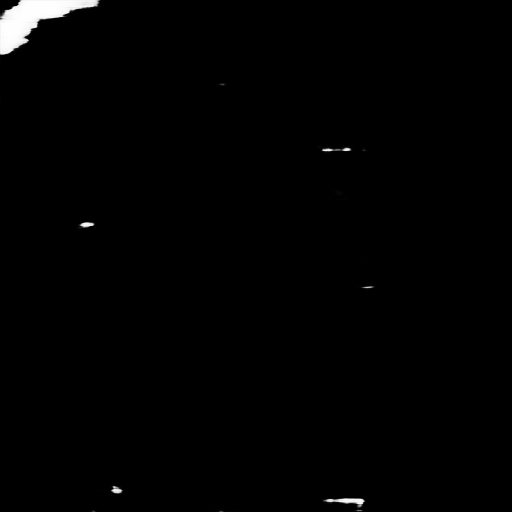} & 
    \includegraphics[width=0.08\linewidth]{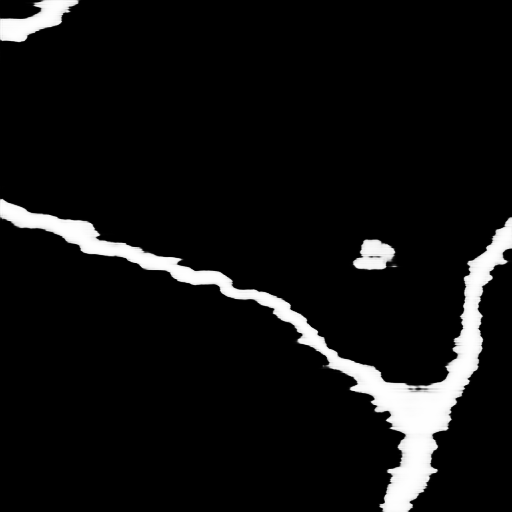} &
    \includegraphics[width=0.08\linewidth]{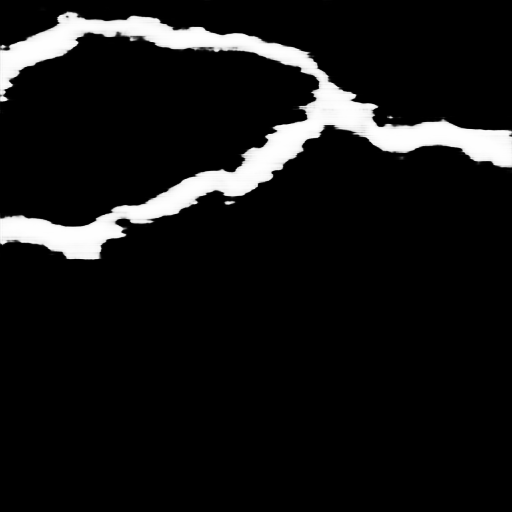} & 
    \includegraphics[width=0.08\linewidth]{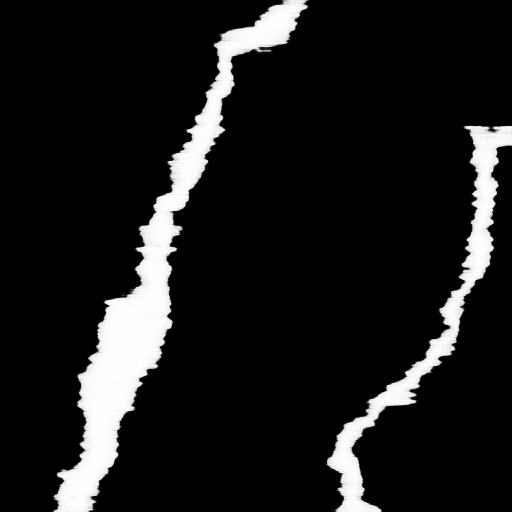} &
    \includegraphics[width=0.08\linewidth]{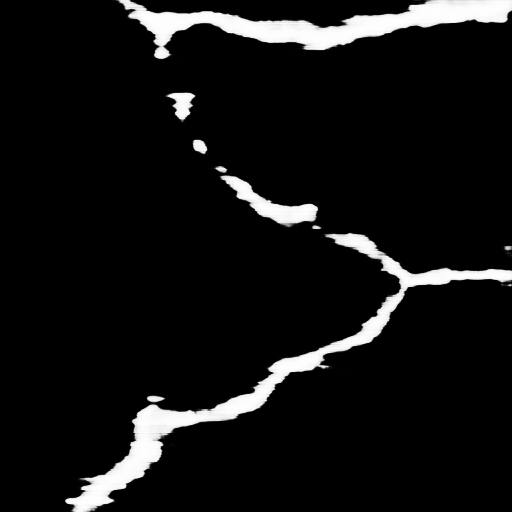} &
    \includegraphics[width=0.08\linewidth]{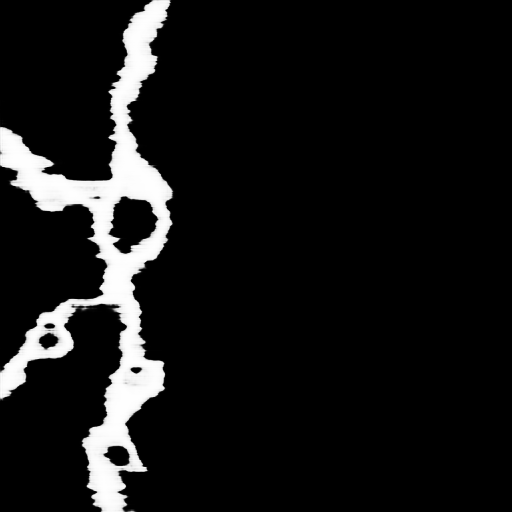} &
    \includegraphics[width=0.08\linewidth]{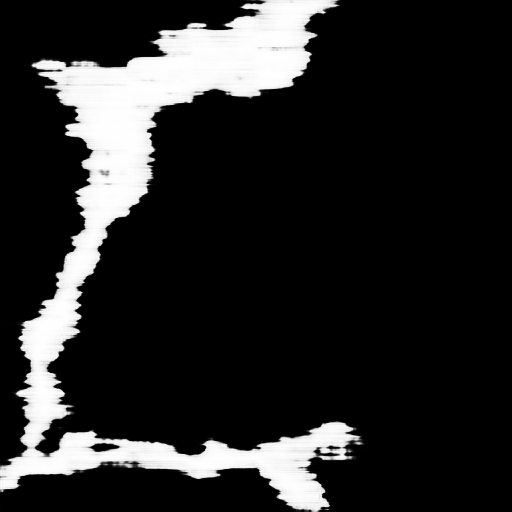} &
    \includegraphics[width=0.08\linewidth]{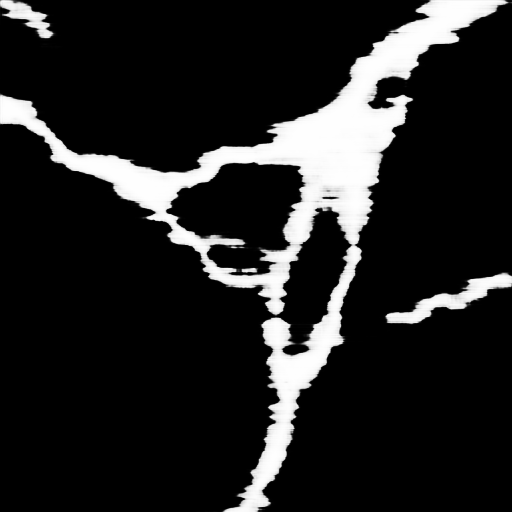} 
    \\
\end{tabular}
\caption{ \footnotesize {\textbf{Visual results on CRACK500.} (a) Input image. (b) Ground truth. (c) UNet. (d) CrackformerII. (e) DeepCrack. (f) cGAN\_CBAM (pixel). (g) cGAN\_CBAM\_Ig (pixel). (h)cGAN\_LSA (pixel).}
}
\label{fig:visual_crack500}
\end{figure*}

\begin{figure*}
\centering
\footnotesize
\renewcommand{\tabcolsep}{1pt} 
\renewcommand{\arraystretch}{0.2} 
\begin{tabular}{cccccccccccc}
    \raisebox{2.5\normalbaselineskip}[0pt][0pt]{\rotatebox[origin=c]{0}{(a)}} &  
    \includegraphics[width=0.08\linewidth]{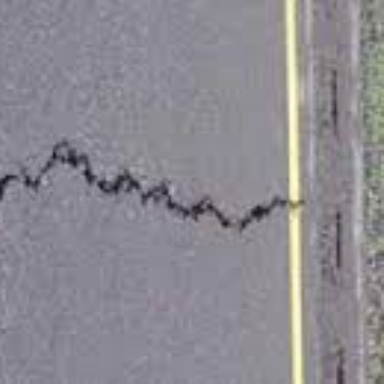} & \includegraphics[width=0.08\linewidth]{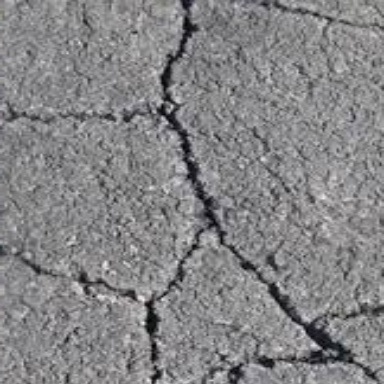} & 
    \includegraphics[width=0.08\linewidth]{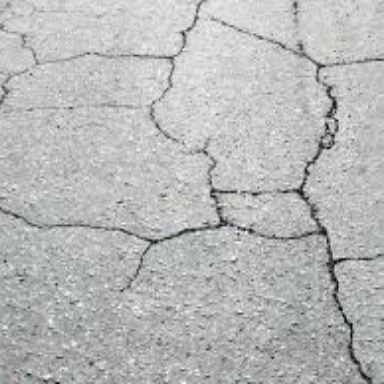} & 
    \includegraphics[width=0.08\linewidth]{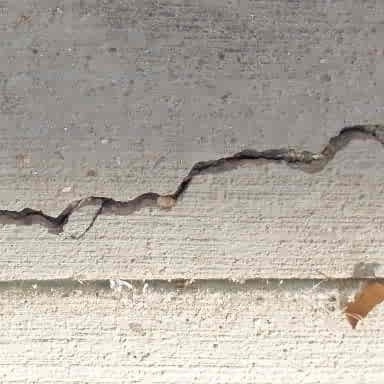} & 
    \includegraphics[width=0.08\linewidth]{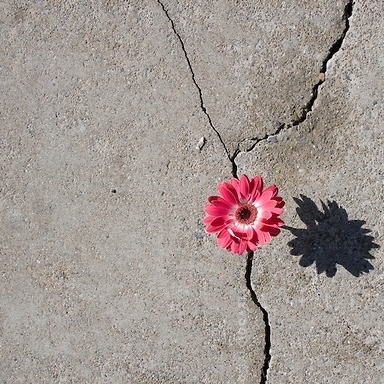} &
    \includegraphics[width=0.08\linewidth]{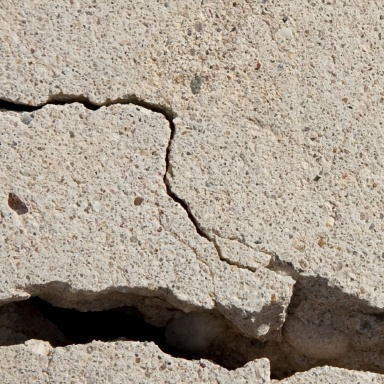} & 
    \includegraphics[width=0.08\linewidth]{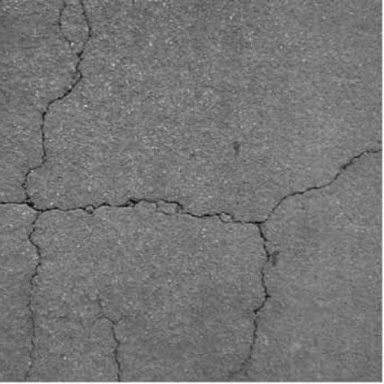} &
    \includegraphics[width=0.08\linewidth]{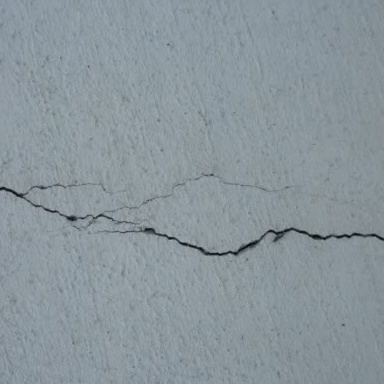} &
    \includegraphics[width=0.08\linewidth]{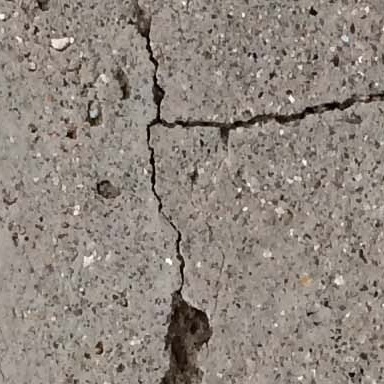} &
    \includegraphics[width=0.08\linewidth]{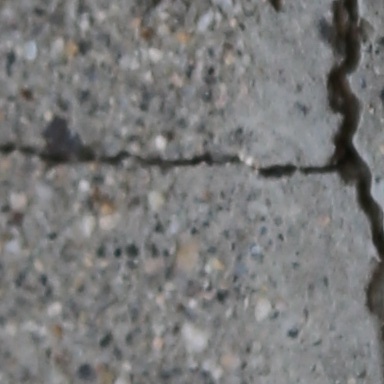} &
    \includegraphics[width=0.08\linewidth]{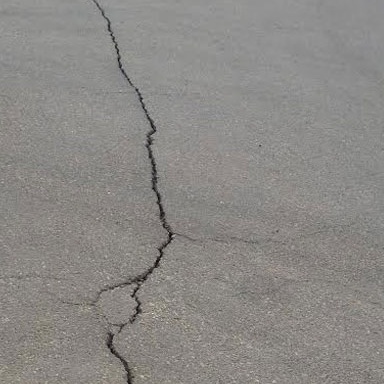} 
    \\

    \raisebox{2.5\normalbaselineskip}[0pt][0pt]{\rotatebox[origin=c]{0}{(b)}} &  
    \includegraphics[width=0.08\linewidth]{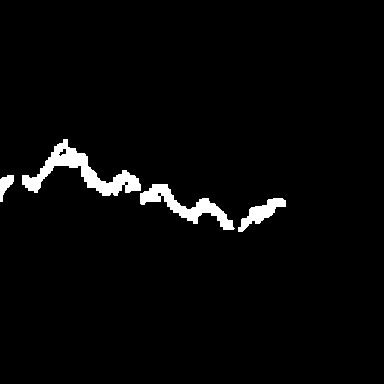} &
    \includegraphics[width=0.08\linewidth]{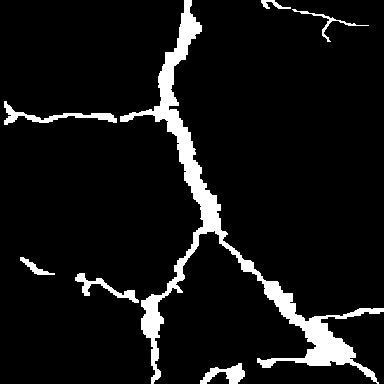} & 
    \includegraphics[width=0.08\linewidth]{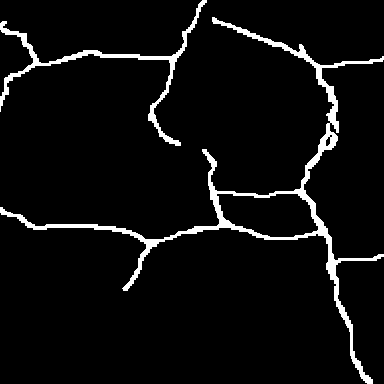} & 
    \includegraphics[width=0.08\linewidth]{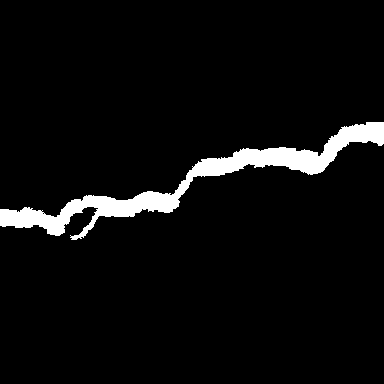} & 
    \includegraphics[width=0.08\linewidth]{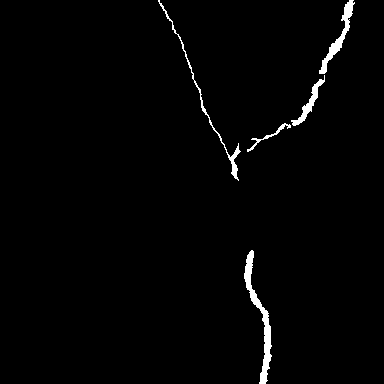} &
    \includegraphics[width=0.08\linewidth]{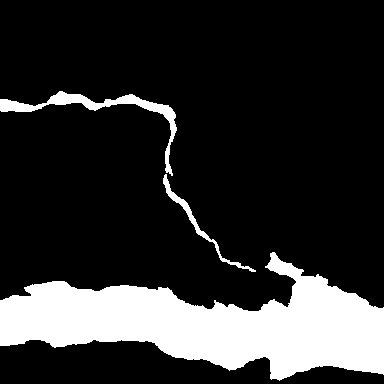} & 
    \includegraphics[width=0.08\linewidth]{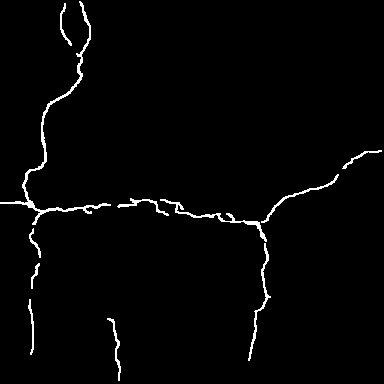} &
    \includegraphics[width=0.08\linewidth]{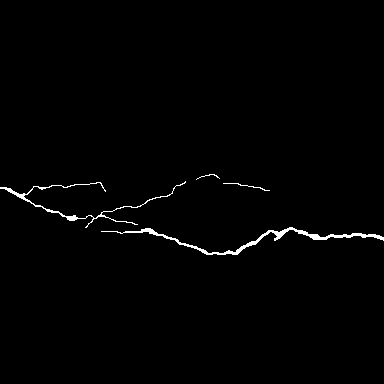} &
    \includegraphics[width=0.08\linewidth]{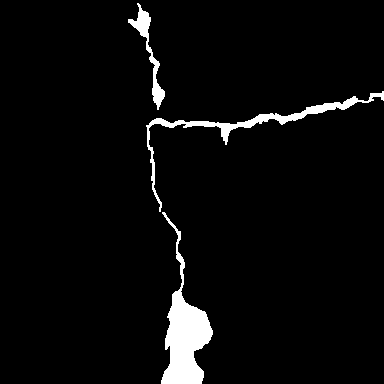} &
    \includegraphics[width=0.08\linewidth]{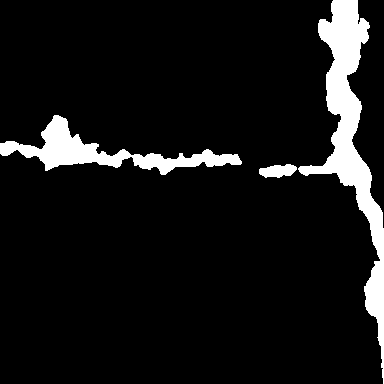} &
    \includegraphics[width=0.08\linewidth]{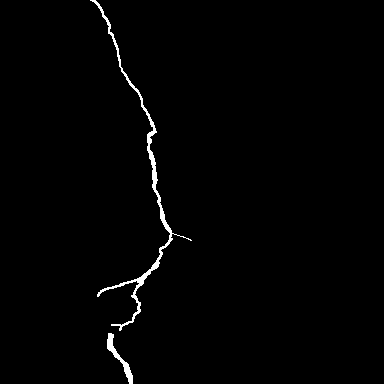} 
    \\

    \raisebox{2.5\normalbaselineskip}[0pt][0pt]{\rotatebox[origin=c]{0}{(c)}} &  
    \includegraphics[width=0.08\linewidth]{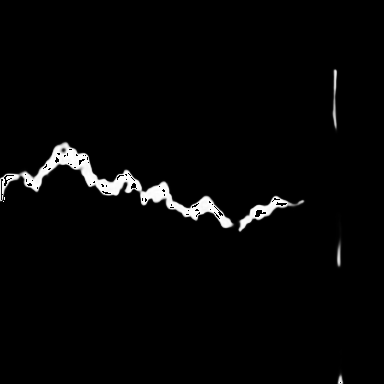} &
    \includegraphics[width=0.08\linewidth]{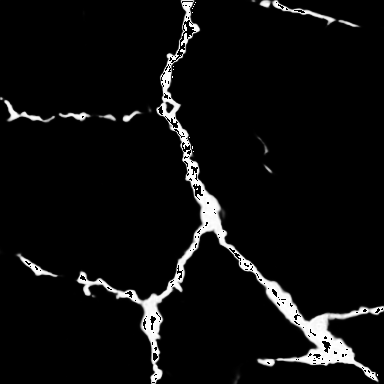} & 
    \includegraphics[width=0.08\linewidth]{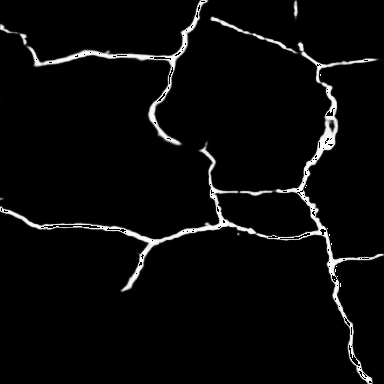} & 
    \includegraphics[width=0.08\linewidth]{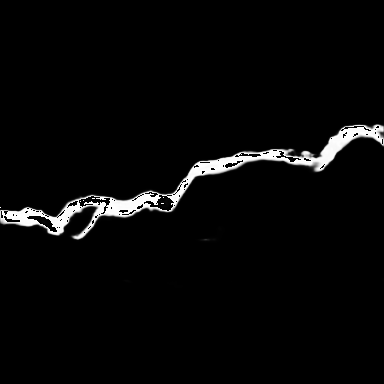} & 
    \includegraphics[width=0.08\linewidth]{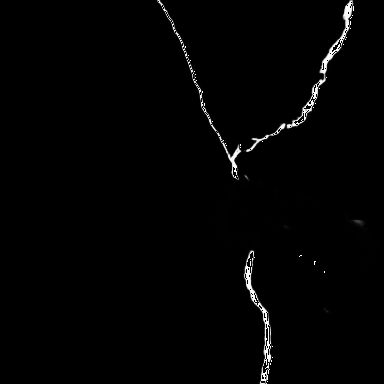} &
    \includegraphics[width=0.08\linewidth]{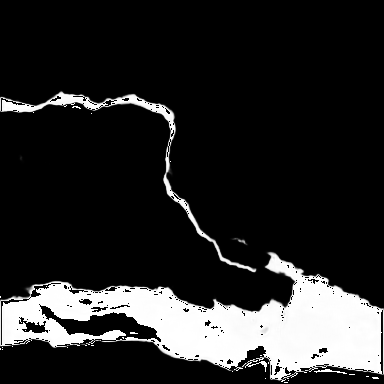} & 
    \includegraphics[width=0.08\linewidth]{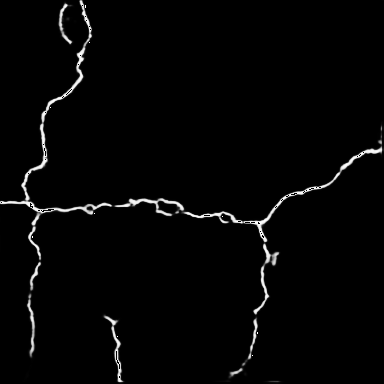} &
    \includegraphics[width=0.08\linewidth]{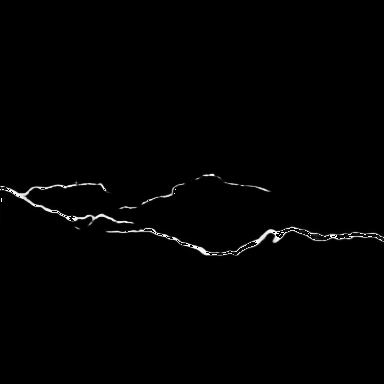} &
    \includegraphics[width=0.08\linewidth]{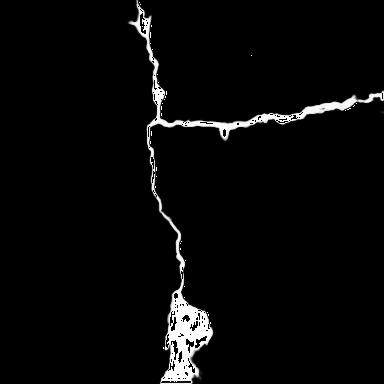}&
    \includegraphics[width=0.08\linewidth]{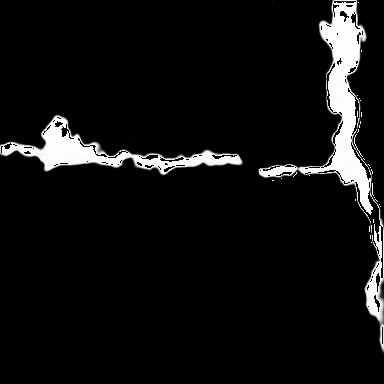} &
    \includegraphics[width=0.08\linewidth]{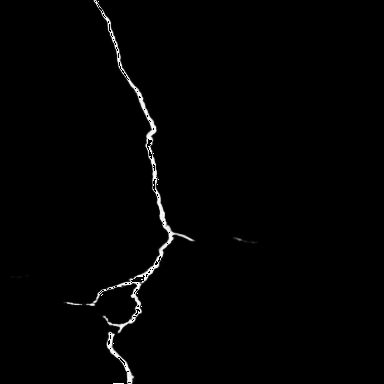}
    \\

    \raisebox{2.5\normalbaselineskip}[0pt][0pt]{\rotatebox[origin=c]{0}{(d)}} &  
    \includegraphics[width=0.08\linewidth]{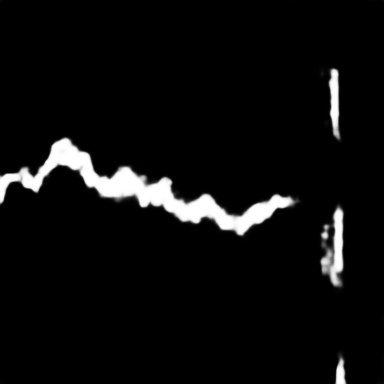} &
    \includegraphics[width=0.08\linewidth]{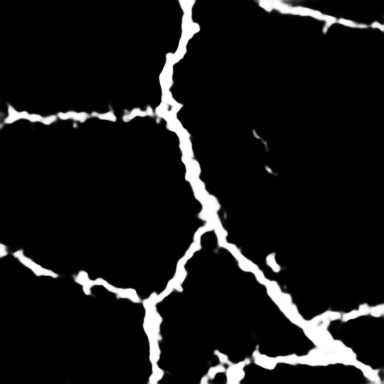} & 
    \includegraphics[width=0.08\linewidth]{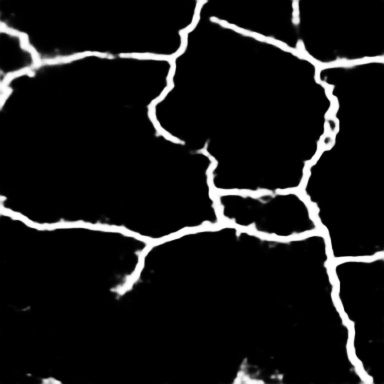} & 
    \includegraphics[width=0.08\linewidth]{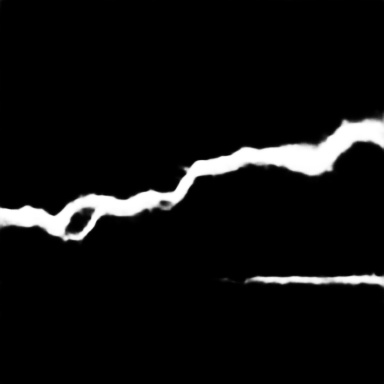} & 
    \includegraphics[width=0.08\linewidth]{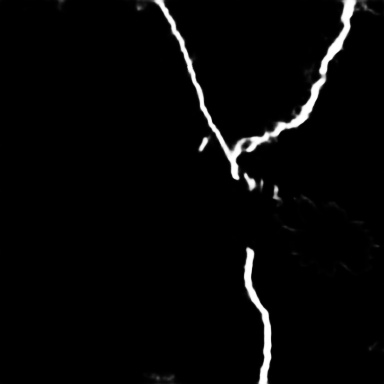} &
    \includegraphics[width=0.08\linewidth]{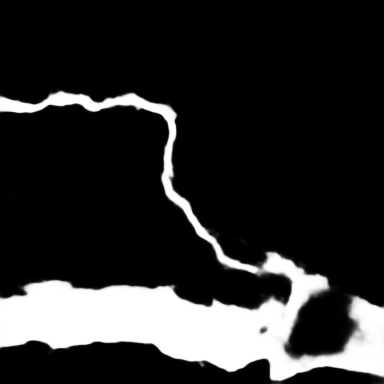} & 
    \includegraphics[width=0.08\linewidth]{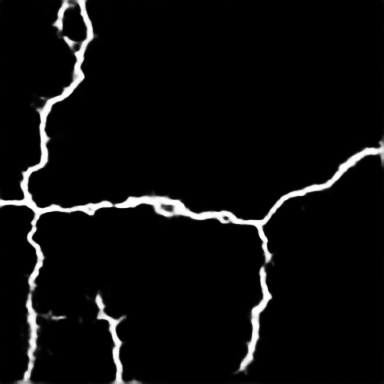} &
    \includegraphics[width=0.08\linewidth]{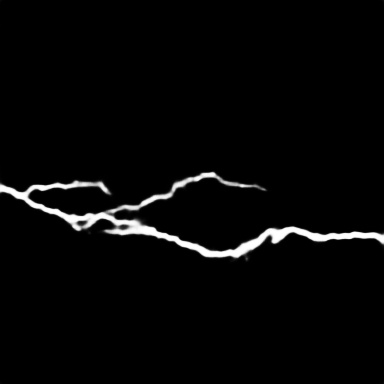} &
    \includegraphics[width=0.08\linewidth]{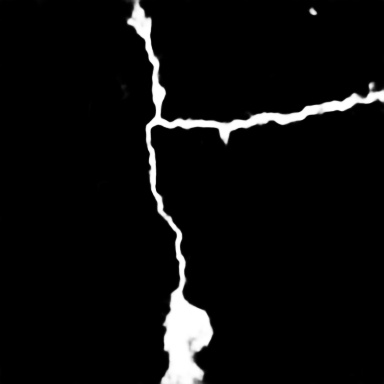} &
    \includegraphics[width=0.08\linewidth]{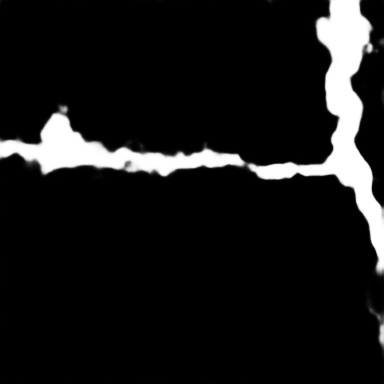} &
    \includegraphics[width=0.08\linewidth]{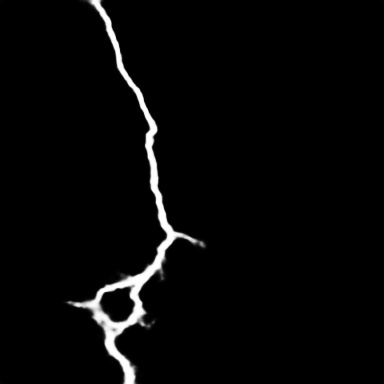} 
    \\

    \raisebox{2.5\normalbaselineskip}[0pt][0pt]{\rotatebox[origin=c]{0}{(e)}} &  
    \includegraphics[width=0.08\linewidth]{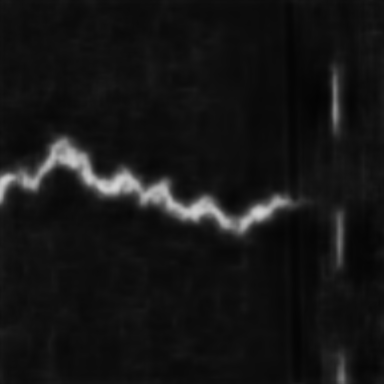} &
    \includegraphics[width=0.08\linewidth]{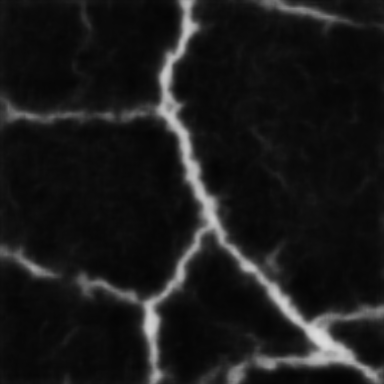} & 
    \includegraphics[width=0.08\linewidth]{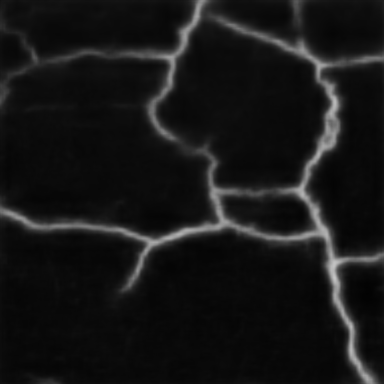} & 
    \includegraphics[width=0.08\linewidth]{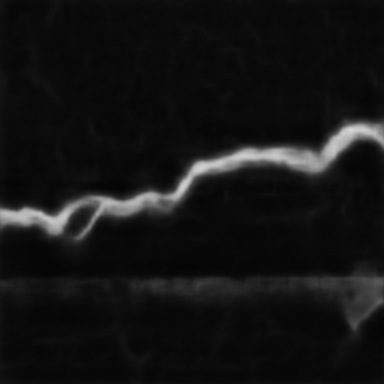} & 
    \includegraphics[width=0.08\linewidth]{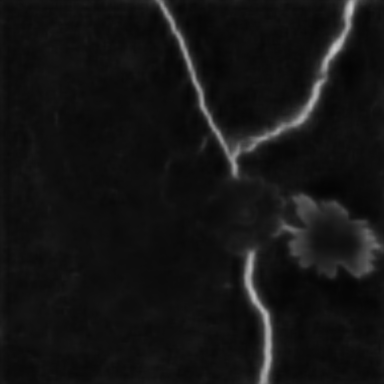} &
    \includegraphics[width=0.08\linewidth]{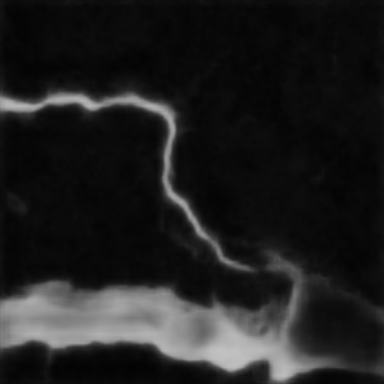} & 
    \includegraphics[width=0.08\linewidth]{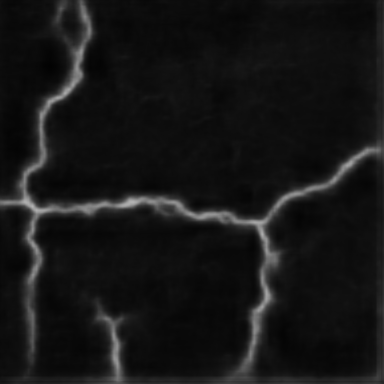} &
    \includegraphics[width=0.08\linewidth]{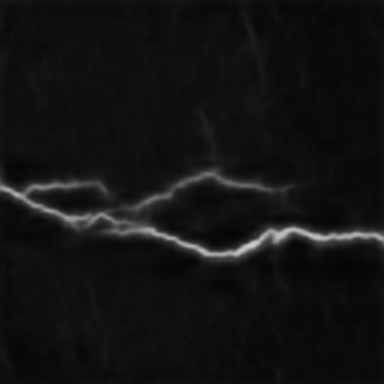} &
    \includegraphics[width=0.08\linewidth]{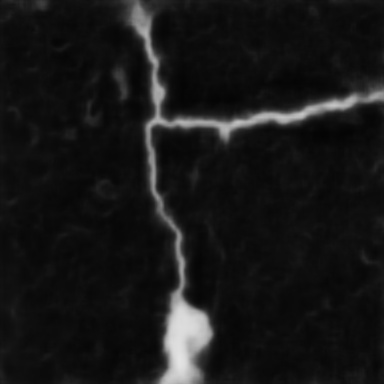}& 
    \includegraphics[width=0.08\linewidth]{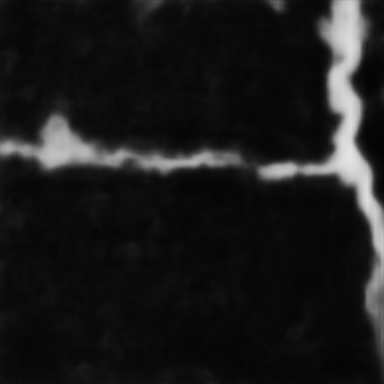} &
    \includegraphics[width=0.08\linewidth]{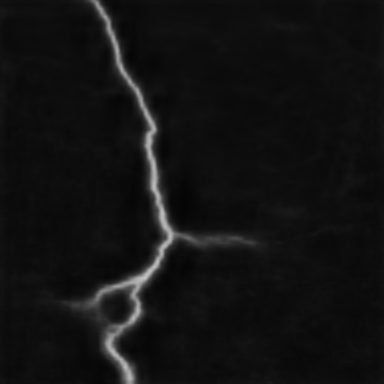}
    \\

    \raisebox{2.5\normalbaselineskip}[0pt][0pt]{\rotatebox[origin=c]{0}{(f)}} &  
    \includegraphics[width=0.08\linewidth]{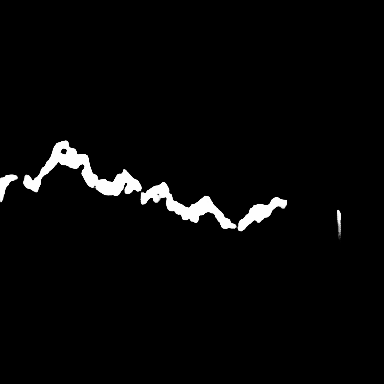} &
    \includegraphics[width=0.08\linewidth]{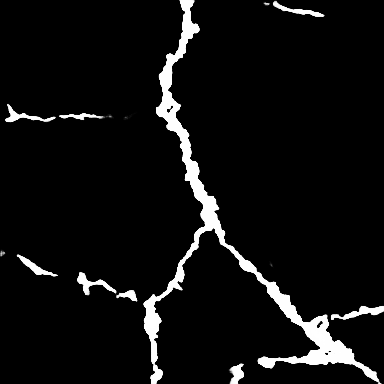} & 
    \includegraphics[width=0.08\linewidth]{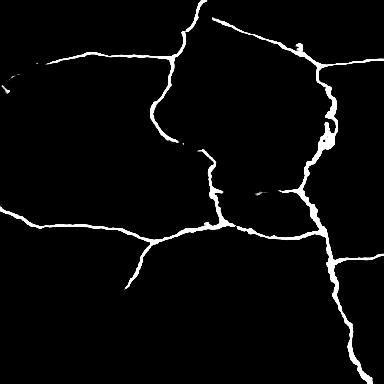} & 
    \includegraphics[width=0.08\linewidth]{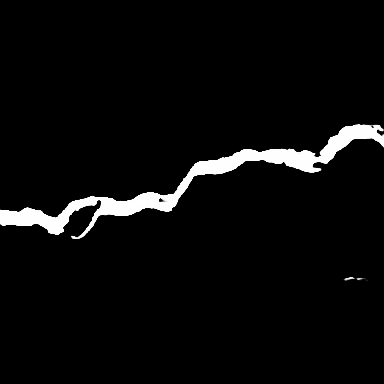} & 
    \includegraphics[width=0.08\linewidth]{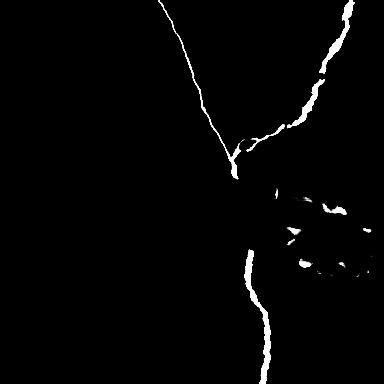} &
    \includegraphics[width=0.08\linewidth]{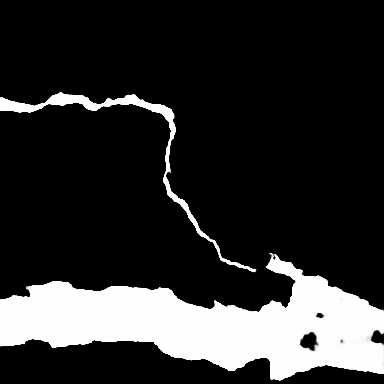} & 
    \includegraphics[width=0.08\linewidth]{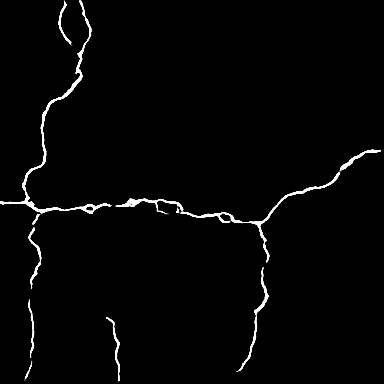} &
    \includegraphics[width=0.08\linewidth]{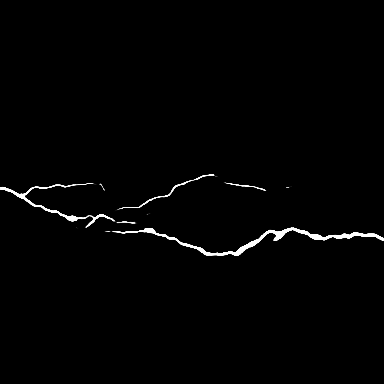} &
    \includegraphics[width=0.08\linewidth]{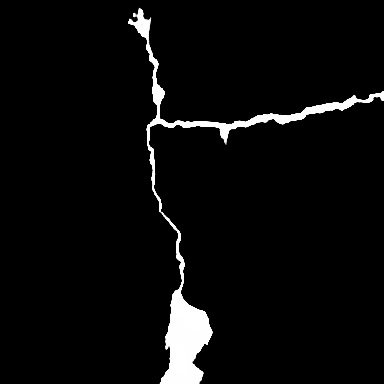} &
    \includegraphics[width=0.08\linewidth]{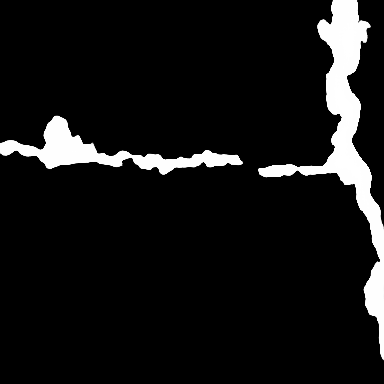} &
    \includegraphics[width=0.08\linewidth]{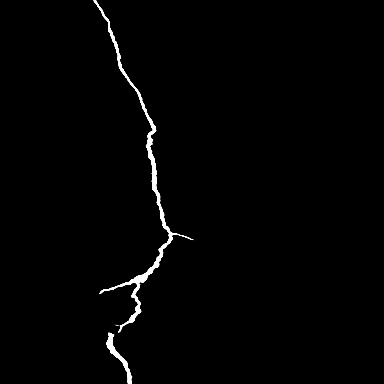} 
    \\

    \raisebox{2.5\normalbaselineskip}[0pt][0pt]{\rotatebox[origin=c]{0}{(g)}} &  
    \includegraphics[width=0.08\linewidth]{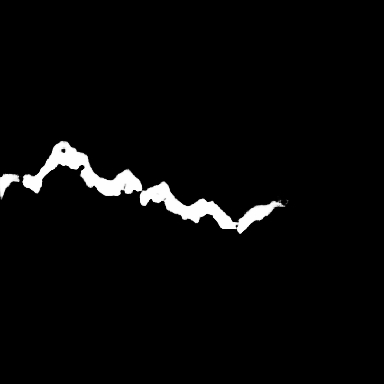} &
    \includegraphics[width=0.08\linewidth]{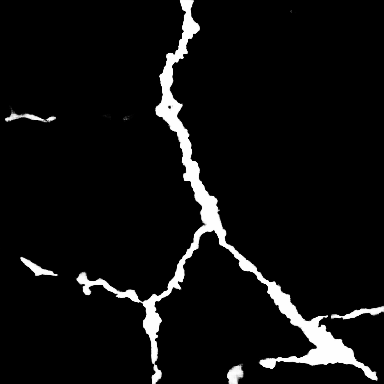} & 
    \includegraphics[width=0.08\linewidth]{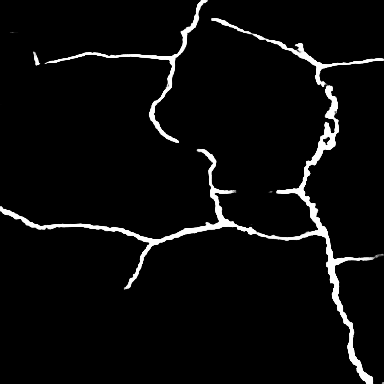} & 
    \includegraphics[width=0.08\linewidth]{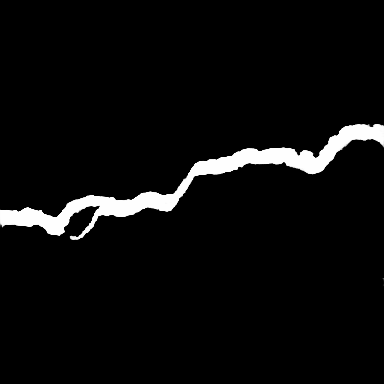} & 
    \includegraphics[width=0.08\linewidth]{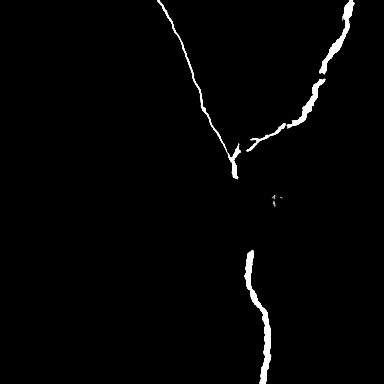} &
    \includegraphics[width=0.08\linewidth]{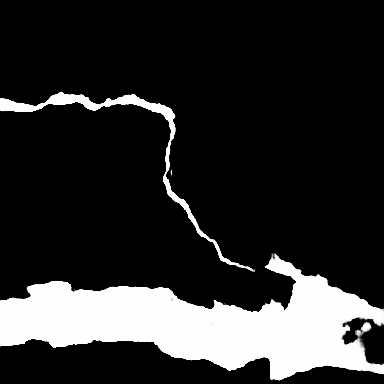} & 
    \includegraphics[width=0.08\linewidth]{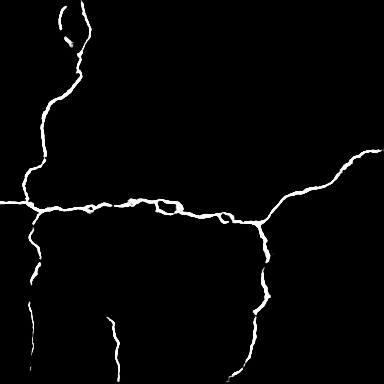} &
    \includegraphics[width=0.08\linewidth]{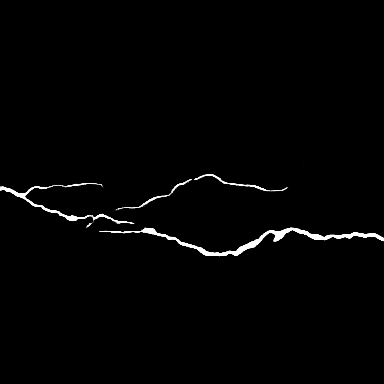} &
    \includegraphics[width=0.08\linewidth]{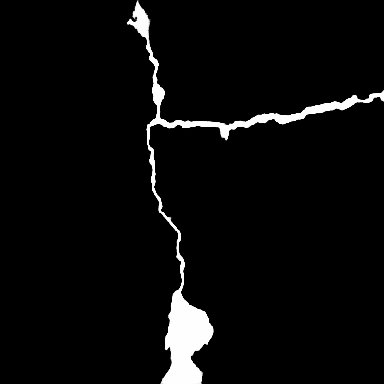} &
    \includegraphics[width=0.08\linewidth]{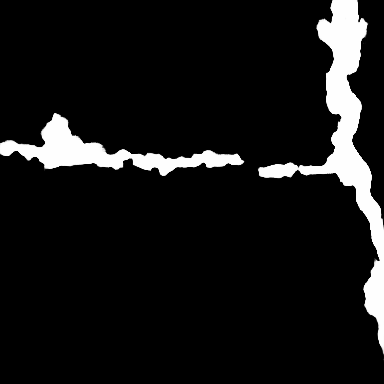} &
    \includegraphics[width=0.08\linewidth]{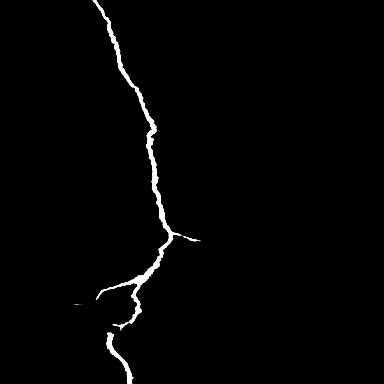}
    
    \\

    \raisebox{2.5\normalbaselineskip}[0pt][0pt]{\rotatebox[origin=c]{0}{(h)}} &  
    \includegraphics[width=0.08\linewidth]{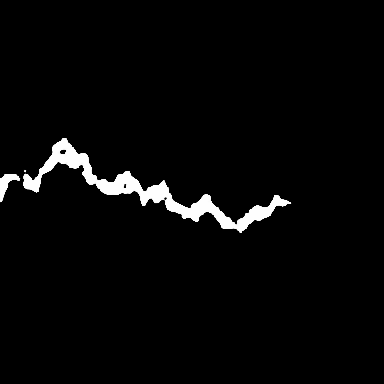} &
    \includegraphics[width=0.08\linewidth]{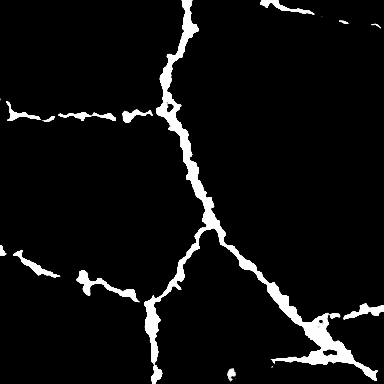} & 
    \includegraphics[width=0.08\linewidth]{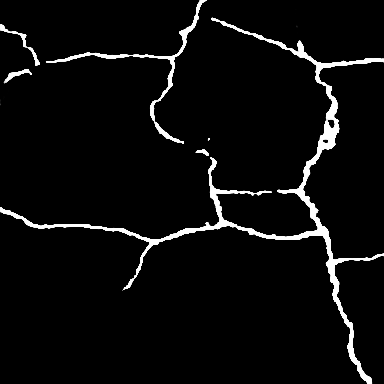} & 
    \includegraphics[width=0.08\linewidth]{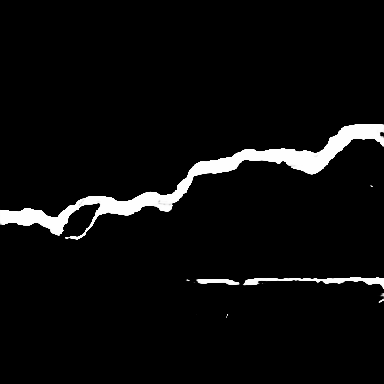} & 
    \includegraphics[width=0.08\linewidth]{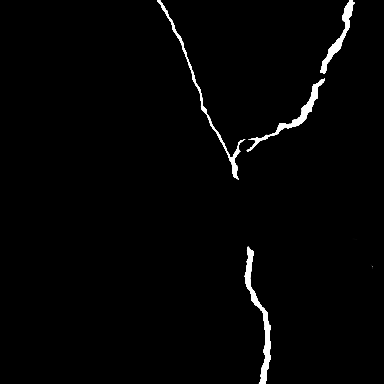} &
    \includegraphics[width=0.08\linewidth]{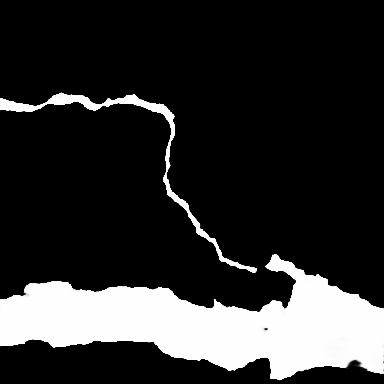} & 
    \includegraphics[width=0.08\linewidth]{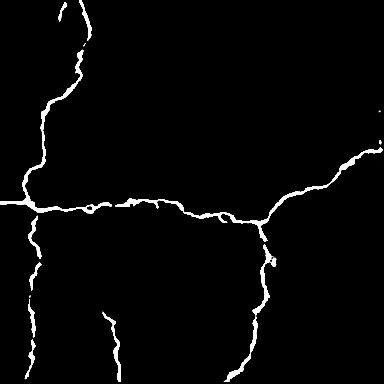} &
    \includegraphics[width=0.08\linewidth]{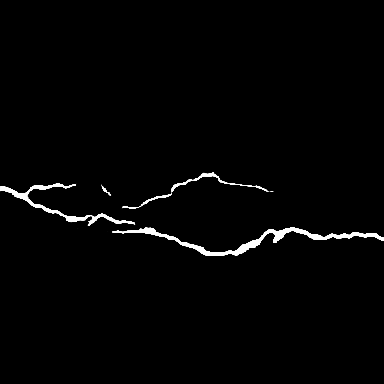} &
    \includegraphics[width=0.08\linewidth]{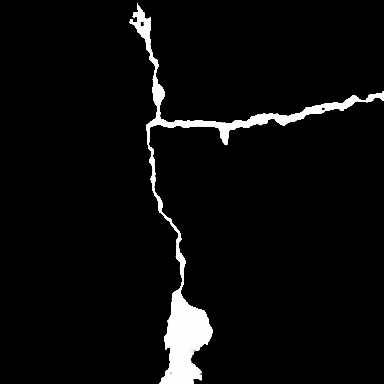} &
    \includegraphics[width=0.08\linewidth]{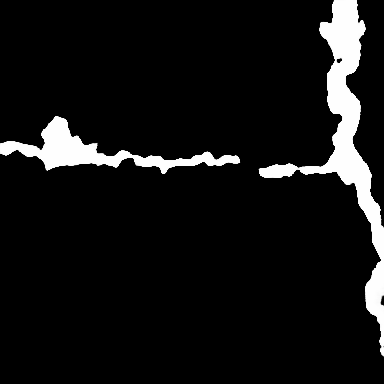} &
    \includegraphics[width=0.08\linewidth]{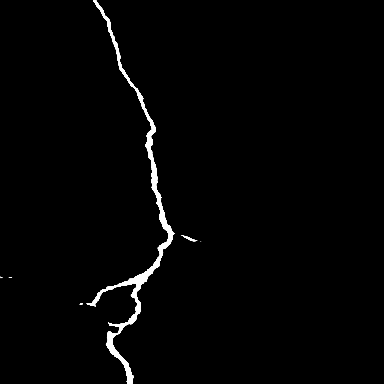} 
    \\
\end{tabular}
\caption{ \footnotesize {\textbf{Visual results on DeepCrack-DB.} (a) Input image. (b) Ground truth. (c) UNet. (d) CrackformerII. (e) DeepCrack. (f) cGAN\_CBAM (pixel). (g) cGAN\_CBAM\_Ig (pixel). (h)cGAN\_LSA (pixel).}
}
\label{fig:visual_deepcrack}
\end{figure*}

\begin{figure}
\centering
\footnotesize
\renewcommand{\tabcolsep}{1pt} 
\renewcommand{\arraystretch}{0.2} 
\begin{tabular}{cccccccccccc}
    \raisebox{1.5\normalbaselineskip}[0pt][0pt]{\rotatebox[origin=c]{0}{(a)}} &  
    \includegraphics[width=0.15\linewidth]{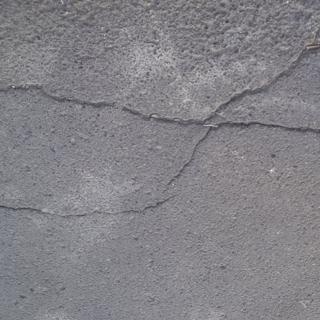} & \includegraphics[width=0.15\linewidth]{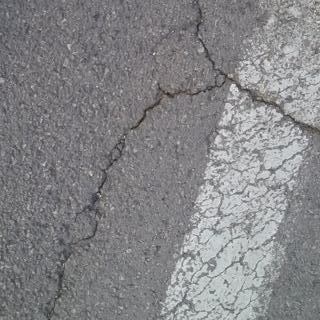} & 
    \includegraphics[width=0.15\linewidth]{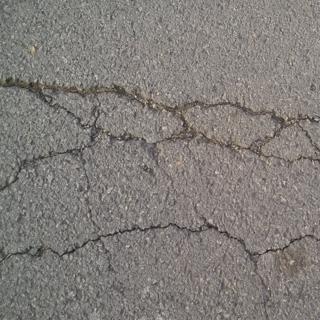} & 
    \includegraphics[width=0.15\linewidth]{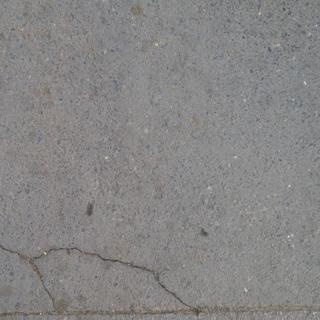} & 
    \includegraphics[width=0.15\linewidth]{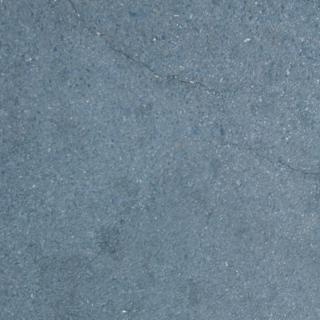} &
    \includegraphics[width=0.15\linewidth]{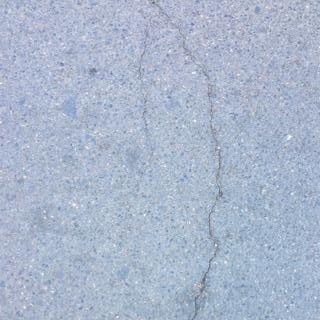} &
    \\

    \raisebox{1.5\normalbaselineskip}[0pt][0pt]{\rotatebox[origin=c]{0}{(b)}} &  
    \includegraphics[width=0.15\linewidth]{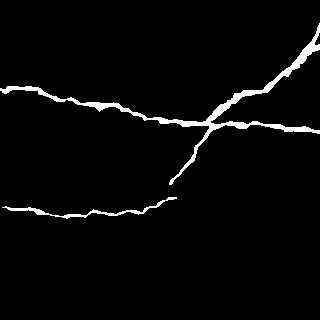} &
    \includegraphics[width=0.15\linewidth]{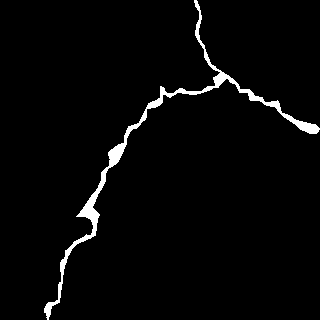} & 
    \includegraphics[width=0.15\linewidth]{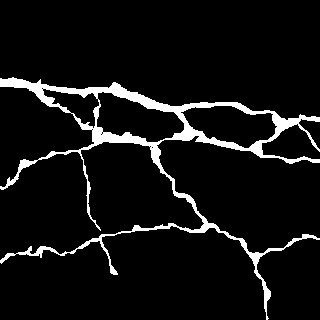} & 
    \includegraphics[width=0.15\linewidth]{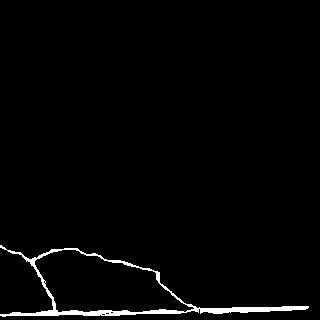} & 
    \includegraphics[width=0.15\linewidth]{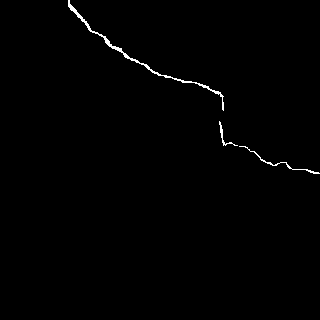} &
    \includegraphics[width=0.15\linewidth]{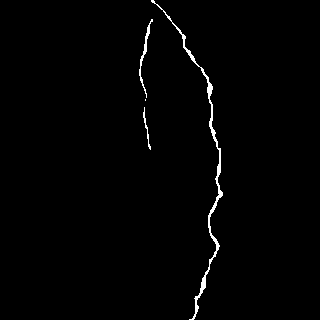} &

    \\

    \raisebox{1.5\normalbaselineskip}[0pt][0pt]{\rotatebox[origin=c]{0}{(c)}} &  
    \includegraphics[width=0.15\linewidth]{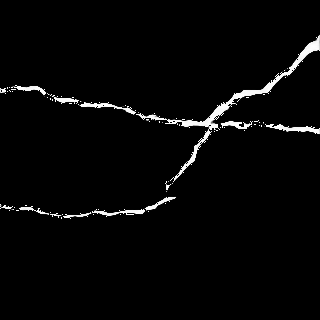} &
    \includegraphics[width=0.15\linewidth]{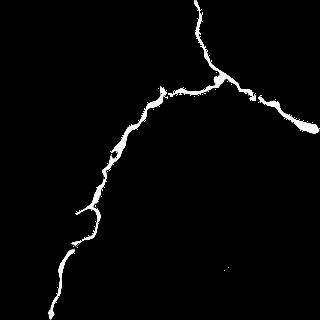} & 
    \includegraphics[width=0.15\linewidth]{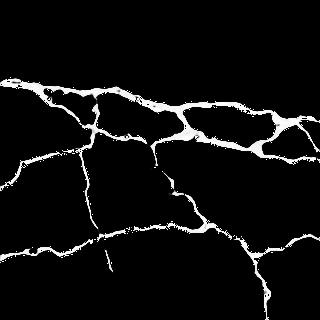} & 
    \includegraphics[width=0.15\linewidth]{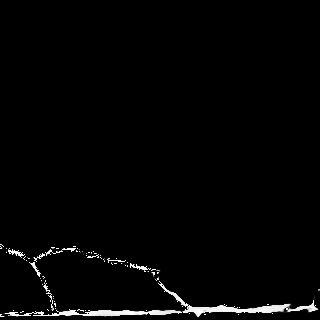} & 
    \includegraphics[width=0.15\linewidth]{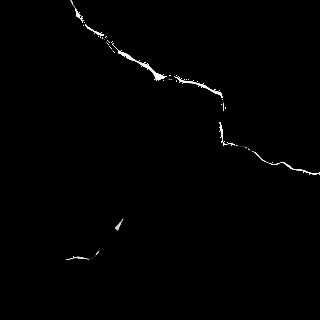} &
    \includegraphics[width=0.15\linewidth]{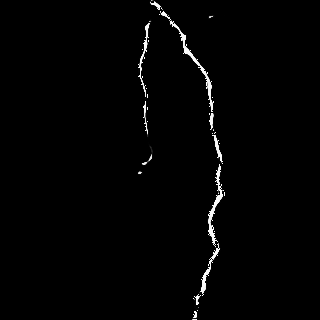}&
    \\

    \raisebox{1.5\normalbaselineskip}[0pt][0pt]{\rotatebox[origin=c]{0}{(d)}} &  
    \includegraphics[width=0.15\linewidth]{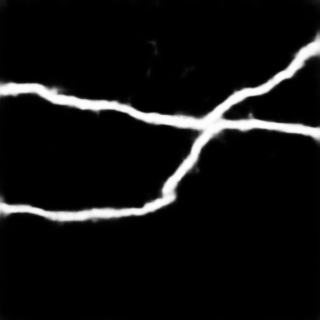} &
    \includegraphics[width=0.15\linewidth]{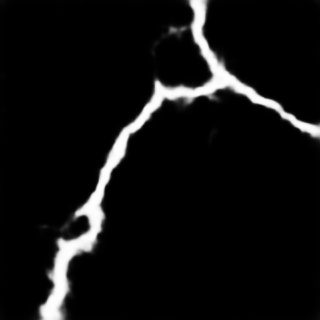} & 
    \includegraphics[width=0.15\linewidth]{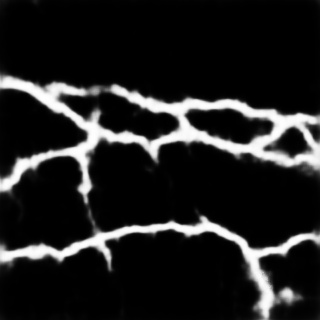} & 
    \includegraphics[width=0.15\linewidth]{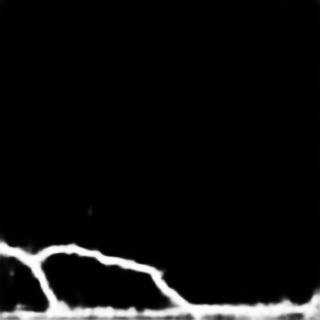} & 
    \includegraphics[width=0.15\linewidth]{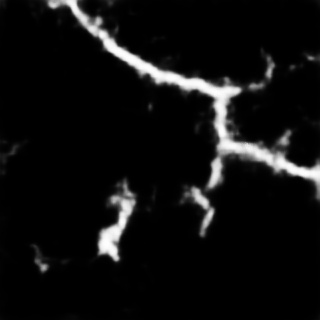} &
    \includegraphics[width=0.15\linewidth]{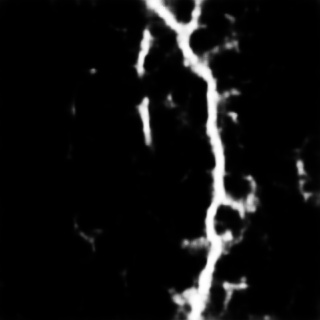} &

    \\

    \raisebox{1.5\normalbaselineskip}[0pt][0pt]{\rotatebox[origin=c]{0}{(e)}} &  
    \includegraphics[width=0.15\linewidth]{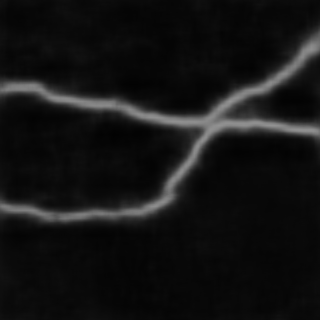} &
    \includegraphics[width=0.15\linewidth]{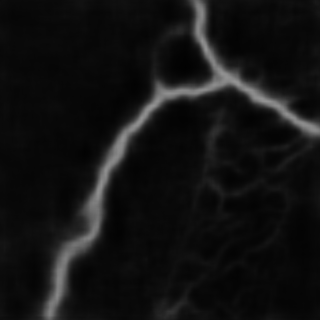} & 
    \includegraphics[width=0.15\linewidth]{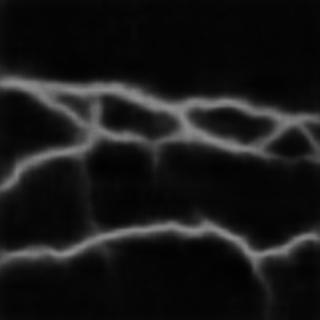} & 
    \includegraphics[width=0.15\linewidth]{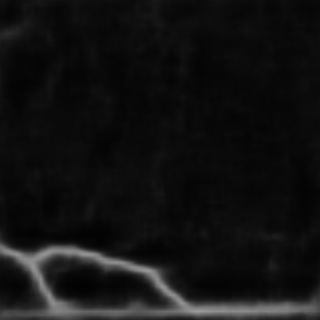} & 
    \includegraphics[width=0.15\linewidth]{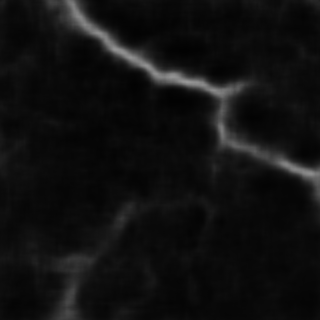} &
    \includegraphics[width=0.15\linewidth]{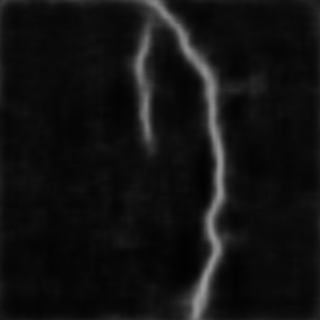}& 
    \\

    \raisebox{1.5\normalbaselineskip}[0pt][0pt]{\rotatebox[origin=c]{0}{(f)}} &  
    \includegraphics[width=0.15\linewidth]{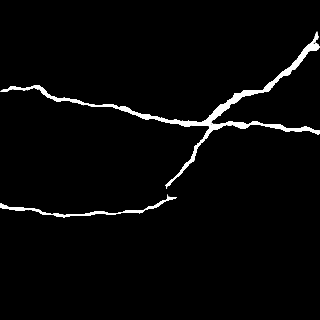} &
    \includegraphics[width=0.15\linewidth]{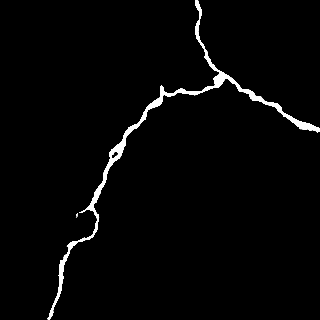} & 
    \includegraphics[width=0.15\linewidth]{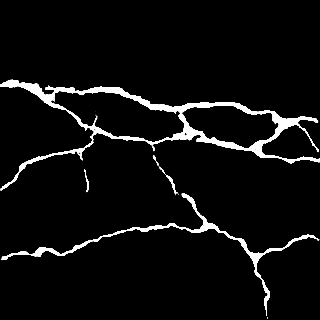} & 
    \includegraphics[width=0.15\linewidth]{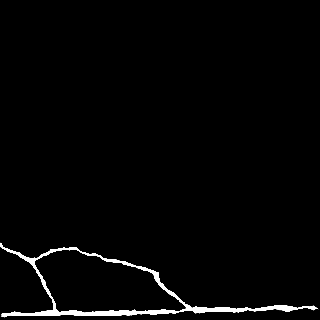} & 
    \includegraphics[width=0.15\linewidth]{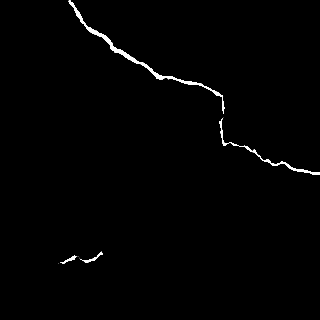} &
    \includegraphics[width=0.15\linewidth]{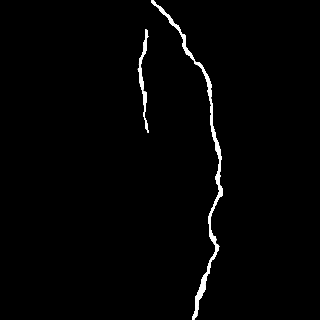} &
    \\

    \raisebox{1.5\normalbaselineskip}[0pt][0pt]{\rotatebox[origin=c]{0}{(g)}} &  
    \includegraphics[width=0.15\linewidth]{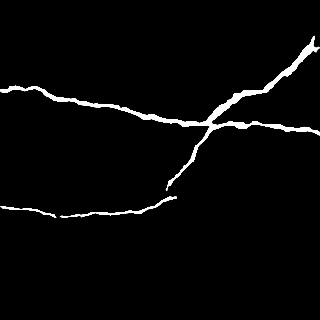} &
    \includegraphics[width=0.15\linewidth]{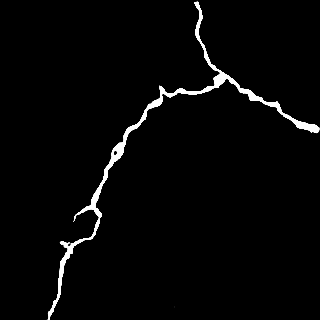} & 
    \includegraphics[width=0.15\linewidth]{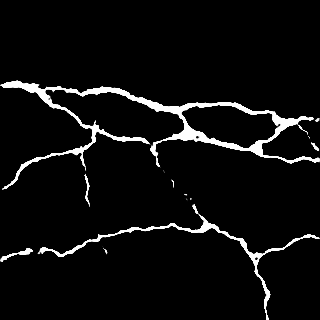} & 
    \includegraphics[width=0.15\linewidth]{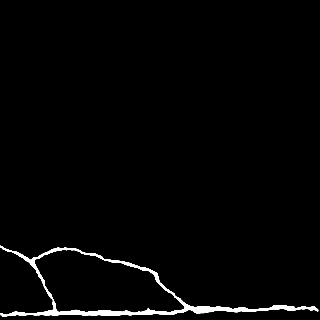} & 
    \includegraphics[width=0.15\linewidth]{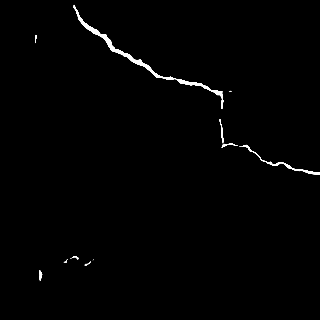} &
    \includegraphics[width=0.15\linewidth]{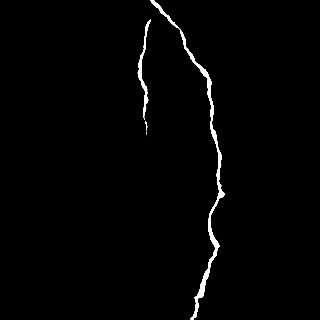} &
    
    \\

    \raisebox{1.5\normalbaselineskip}[0pt][0pt]{\rotatebox[origin=c]{0}{(h)}} &  
    \includegraphics[width=0.15\linewidth]{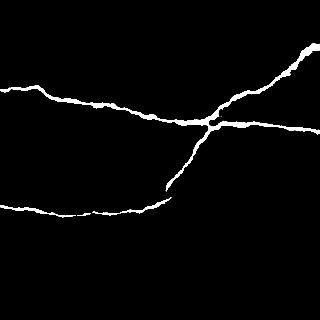} &
    \includegraphics[width=0.15\linewidth]{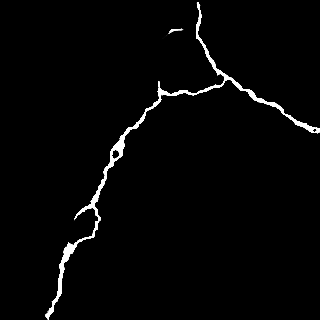} & 
    \includegraphics[width=0.15\linewidth]{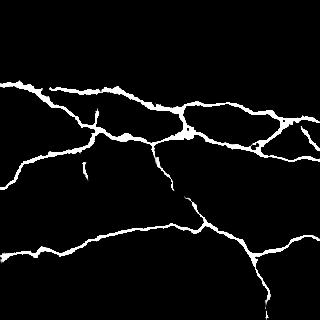} & 
    \includegraphics[width=0.15\linewidth]{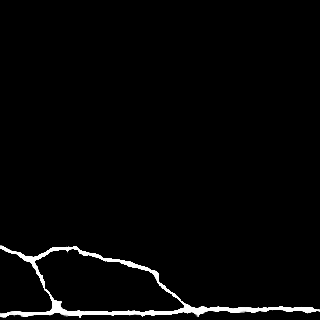} & 
    \includegraphics[width=0.15\linewidth]{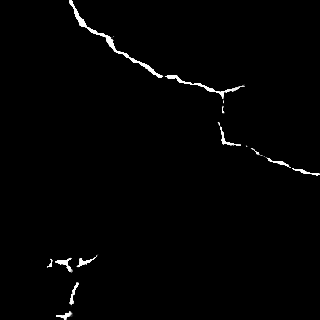} &
    \includegraphics[width=0.15\linewidth]{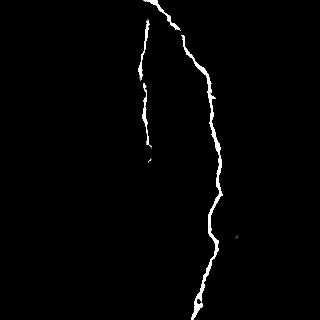} & 
    \\
\end{tabular}
\caption{\footnotesize {\textbf{Visual results on CFD.} (a) Input image. (b) Ground truth. (c) UNet. (d) CrackformerII. (e) DeepCrack. (f) cGAN\_CBAM (pixel). (g) cGAN\_CBAM\_Ig (pixel). (h)cGAN\_LSA (pixel).}
}
\label{fig:visual_cfd}
\end{figure}

\begin{figure}
\centering
\footnotesize
\renewcommand{\tabcolsep}{1pt} 
\renewcommand{\arraystretch}{0.2} 
\begin{tabular}{cccccccccccc}
    \raisebox{1.5\normalbaselineskip}[0pt][0pt]{\rotatebox[origin=c]{0}{(a)}} &  
    \includegraphics[width=0.15\linewidth]{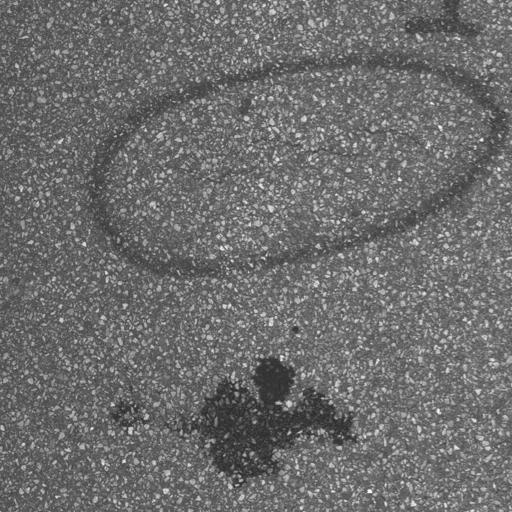} &
     \includegraphics[width=0.15\linewidth]{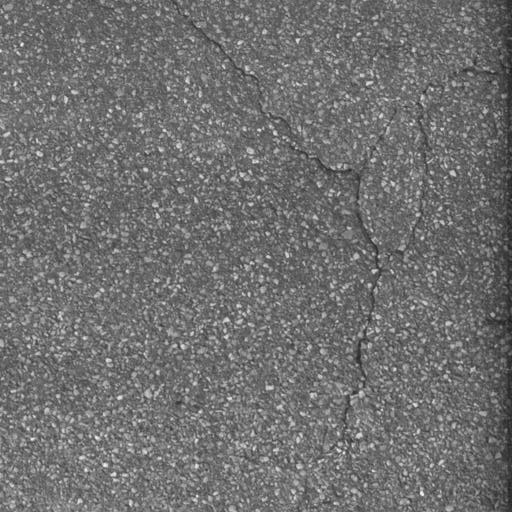} & 
  
    \includegraphics[width=0.15\linewidth]{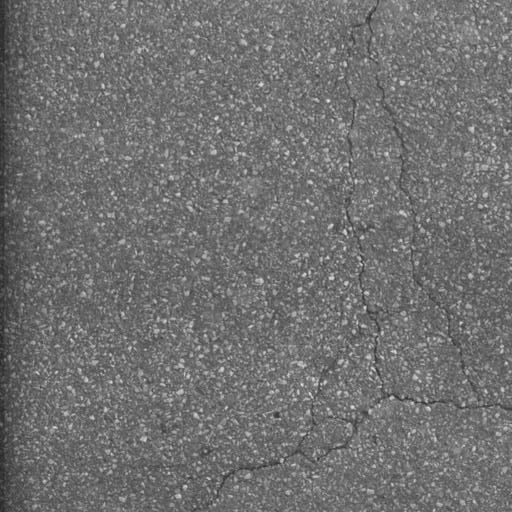} &
  
    \includegraphics[width=0.15\linewidth]{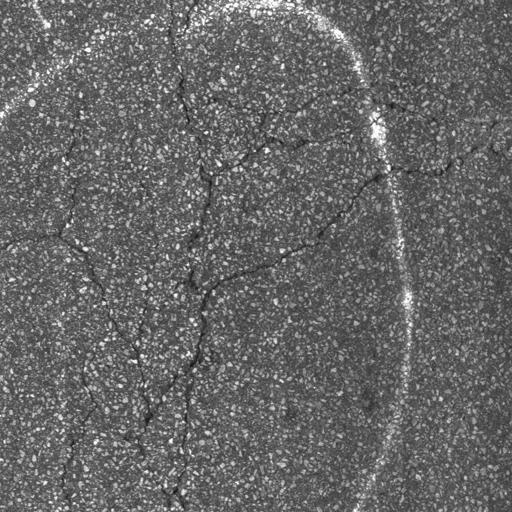} &
 
    \includegraphics[width=0.15\linewidth]{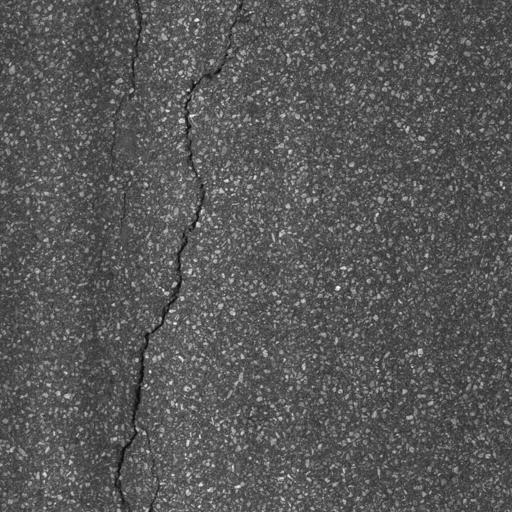} &

    \includegraphics[width=0.15\linewidth]{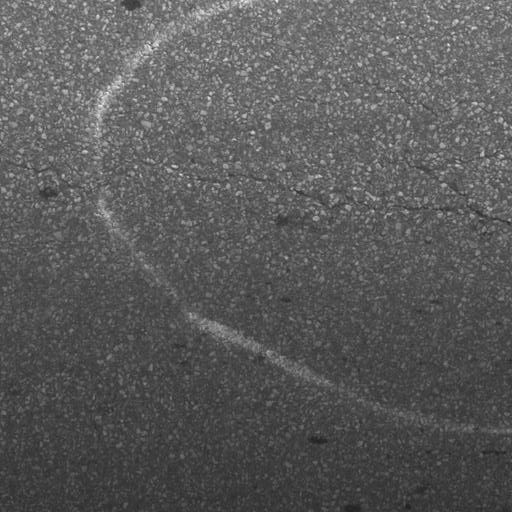} 
    \\

    \raisebox{1.5\normalbaselineskip}[0pt][0pt]{\rotatebox[origin=c]{0}{(b)}} &  
    \includegraphics[width=0.15\linewidth]{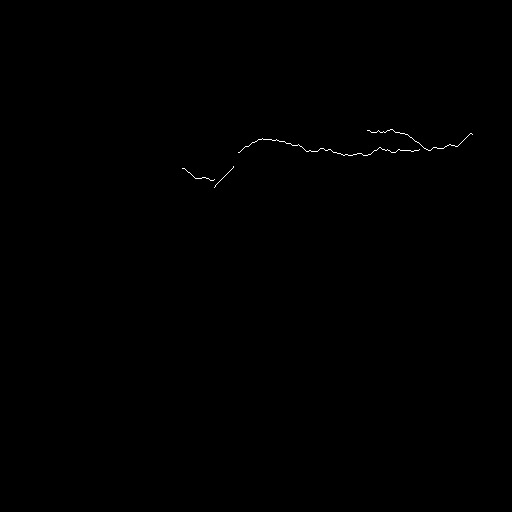} &
    \includegraphics[width=0.15\linewidth]{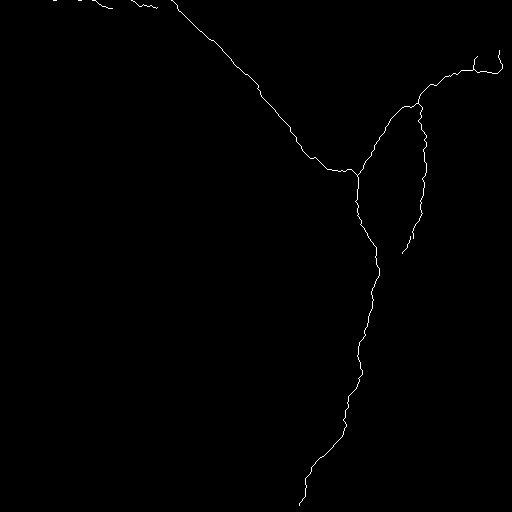} & 

    \includegraphics[width=0.15\linewidth]{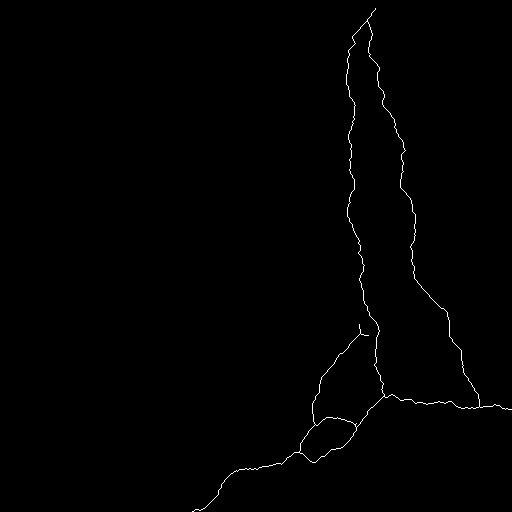} &
    
    \includegraphics[width=0.15\linewidth]{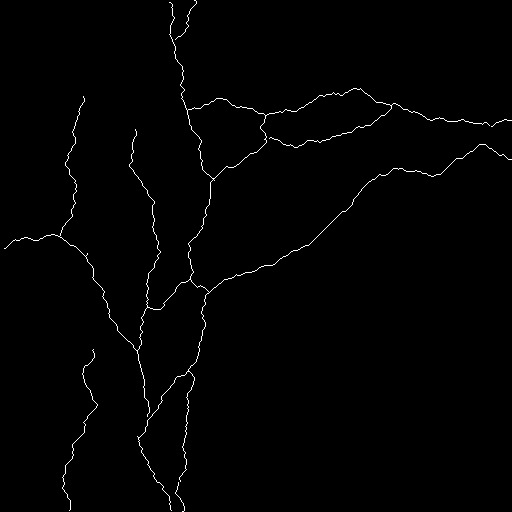} &
  
    \includegraphics[width=0.15\linewidth]{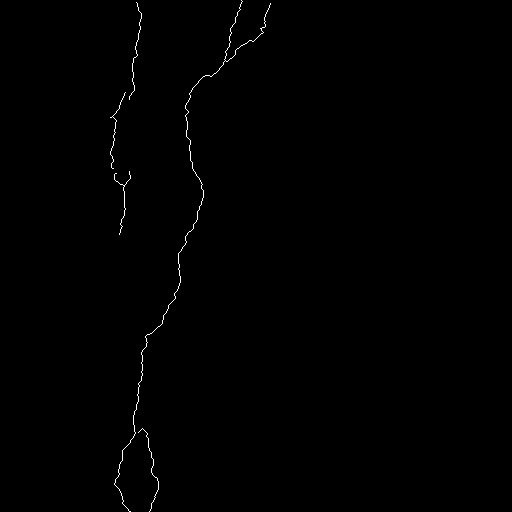} &

    \includegraphics[width=0.15\linewidth]{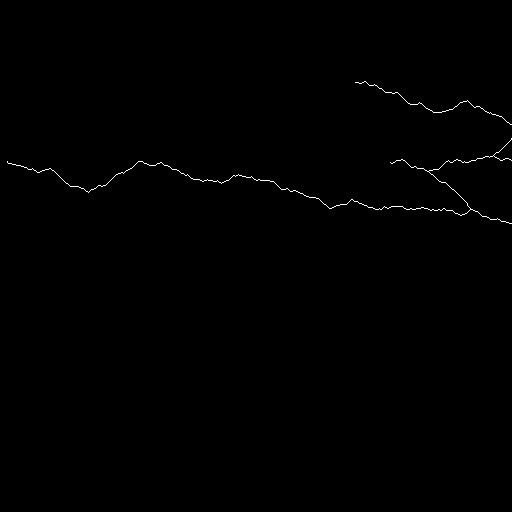} 
    \\

    \raisebox{1.5\normalbaselineskip}[0pt][0pt]{\rotatebox[origin=c]{0}{(c)}} &  
    \includegraphics[width=0.15\linewidth]{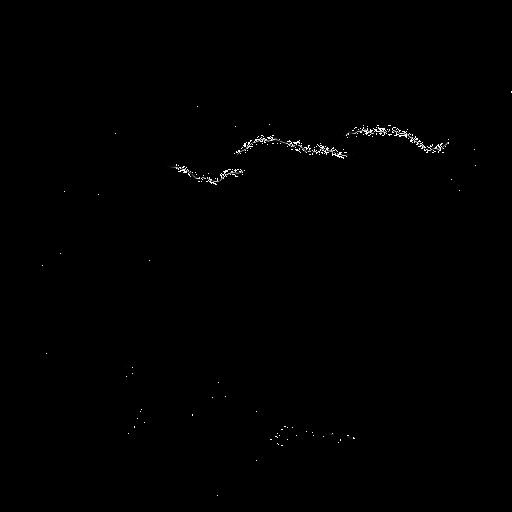} &
    \includegraphics[width=0.15\linewidth]{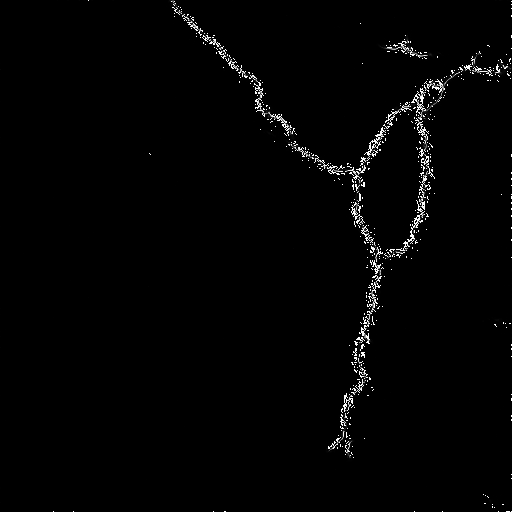} & 

    \includegraphics[width=0.15\linewidth]{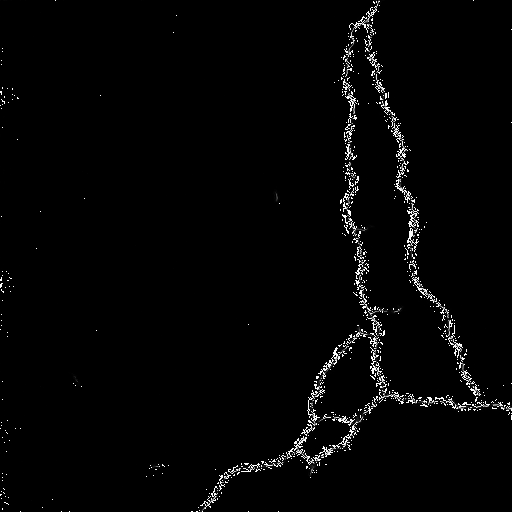} &

    \includegraphics[width=0.15\linewidth]{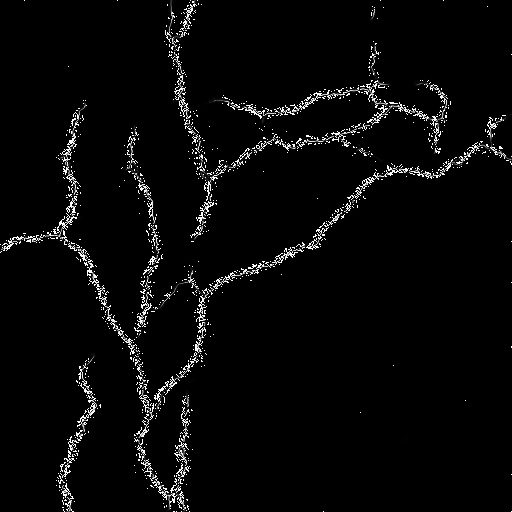} &

    \includegraphics[width=0.15\linewidth]{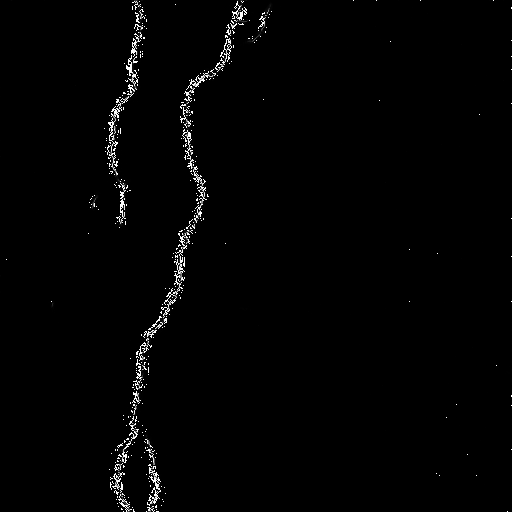}&

    \includegraphics[width=0.15\linewidth]{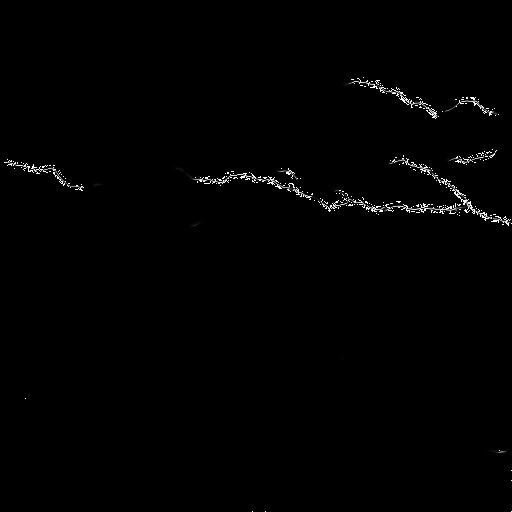}
    \\

    \raisebox{1.5\normalbaselineskip}[0pt][0pt]{\rotatebox[origin=c]{0}{(d)}} &  
    \includegraphics[width=0.15\linewidth]{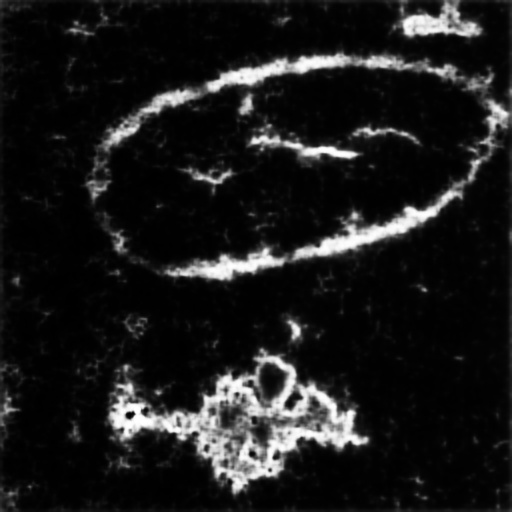} &
    \includegraphics[width=0.15\linewidth]{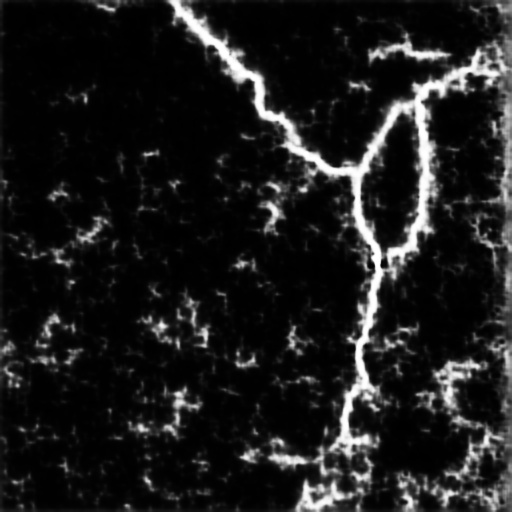} & 
  
    \includegraphics[width=0.15\linewidth]{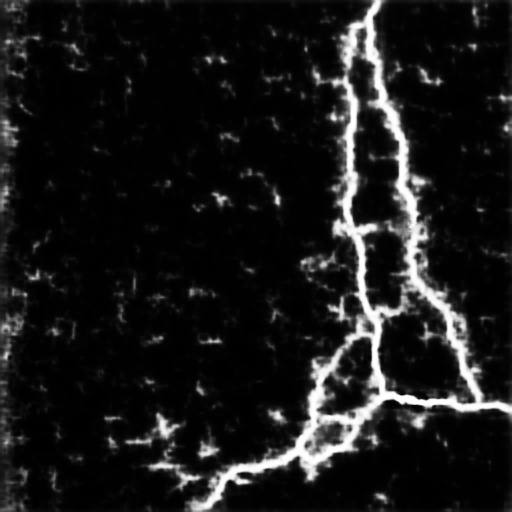} &

    \includegraphics[width=0.15\linewidth]{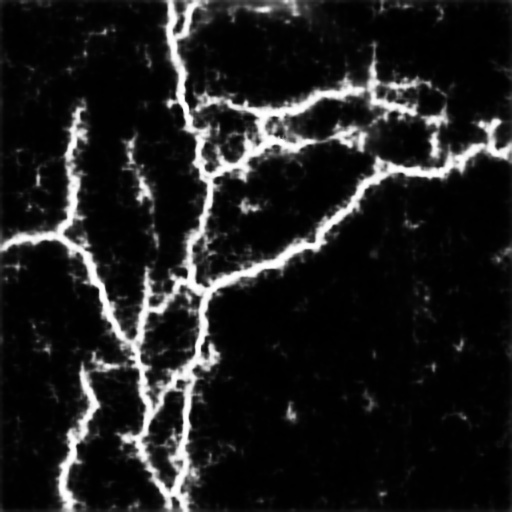} &

    \includegraphics[width=0.15\linewidth]{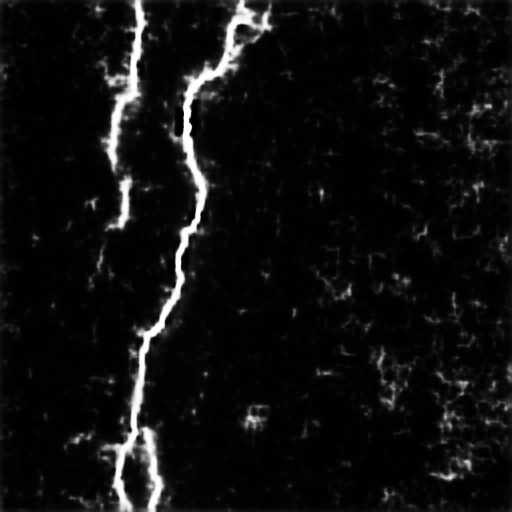} &

    \includegraphics[width=0.15\linewidth]{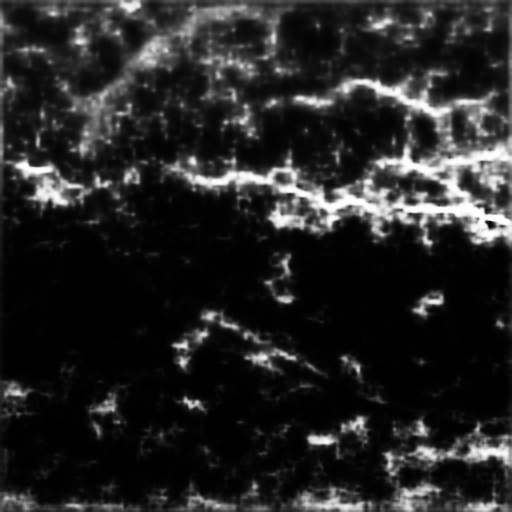} 
    \\

    \raisebox{1.5\normalbaselineskip}[0pt][0pt]{\rotatebox[origin=c]{0}{(e)}} &  
    \includegraphics[width=0.15\linewidth]{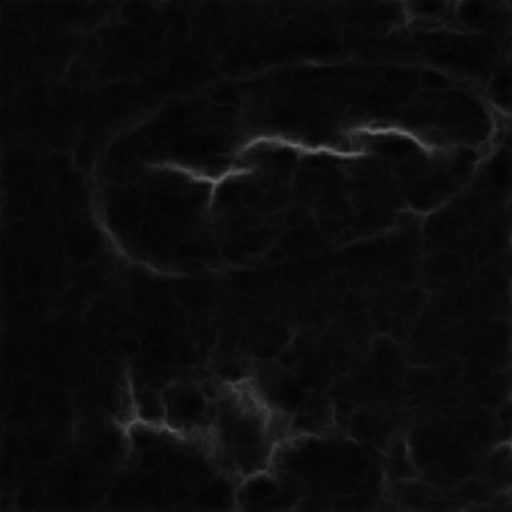} &
    \includegraphics[width=0.15\linewidth]{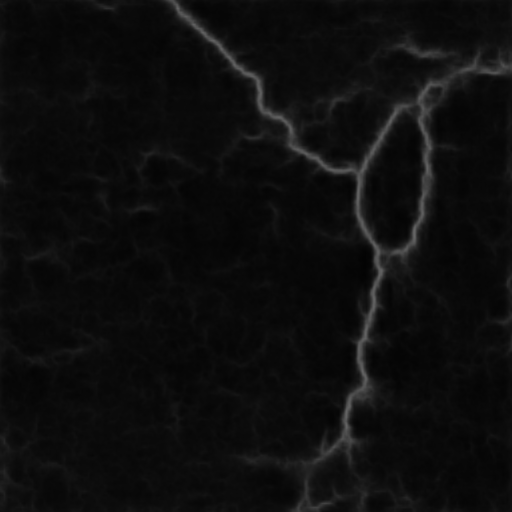} & 

    \includegraphics[width=0.15\linewidth]{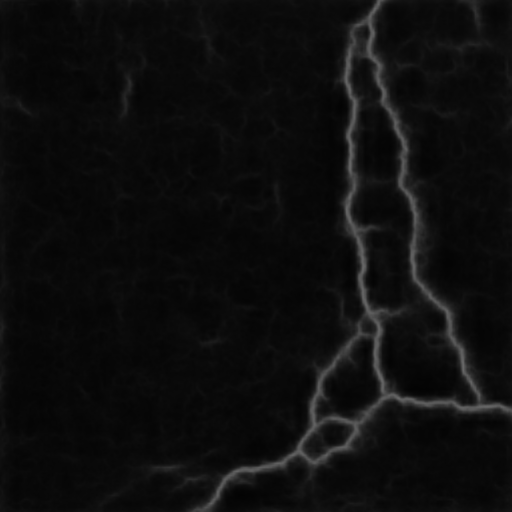} &

    \includegraphics[width=0.15\linewidth]{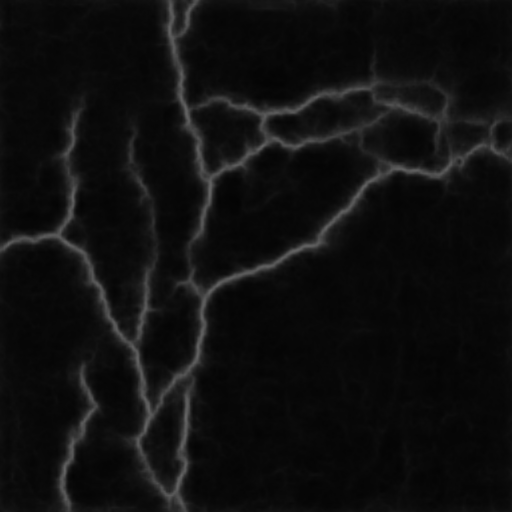} &

    \includegraphics[width=0.15\linewidth]{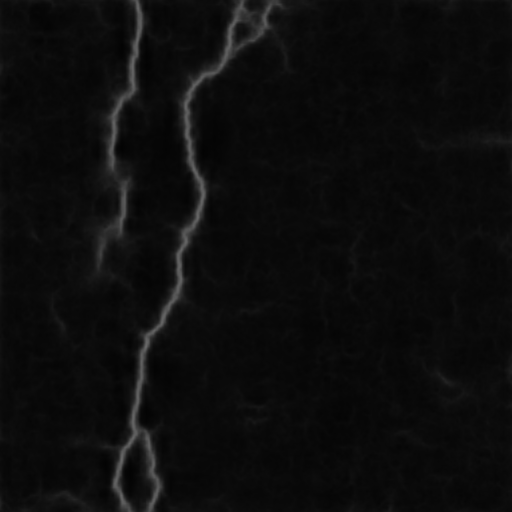}& 

    \includegraphics[width=0.15\linewidth]{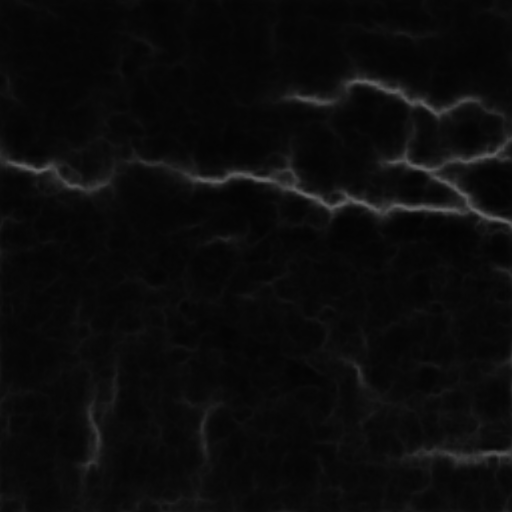}
    \\

    \raisebox{1.5\normalbaselineskip}[0pt][0pt]{\rotatebox[origin=c]{0}{(f)}} &  
    \includegraphics[width=0.15\linewidth]{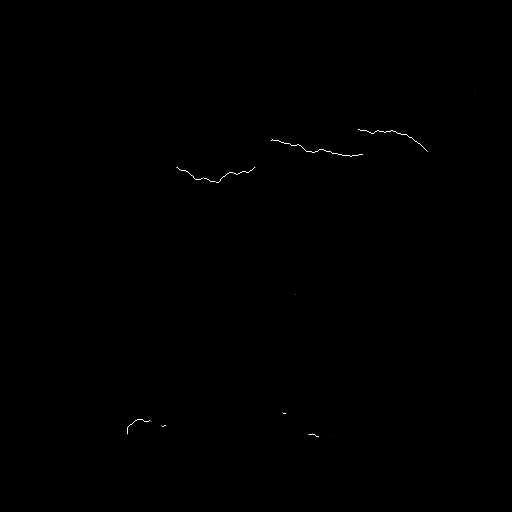} &
    \includegraphics[width=0.15\linewidth]{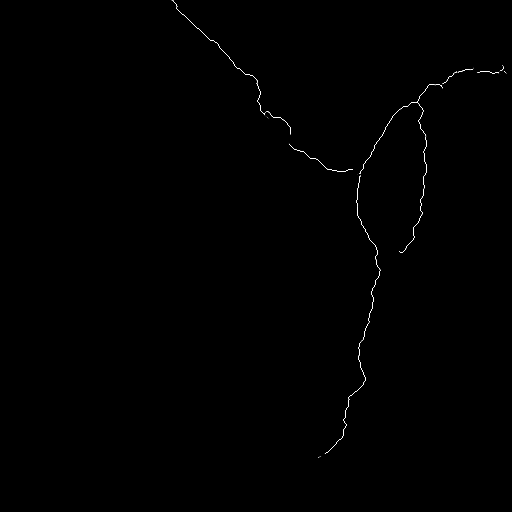} & 
 
    \includegraphics[width=0.15\linewidth]{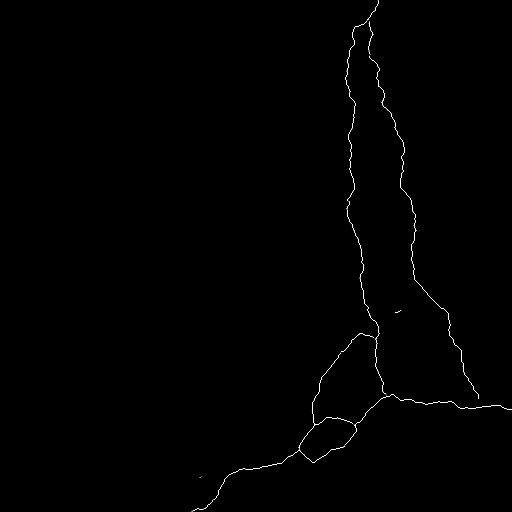} &
  
    \includegraphics[width=0.15\linewidth]{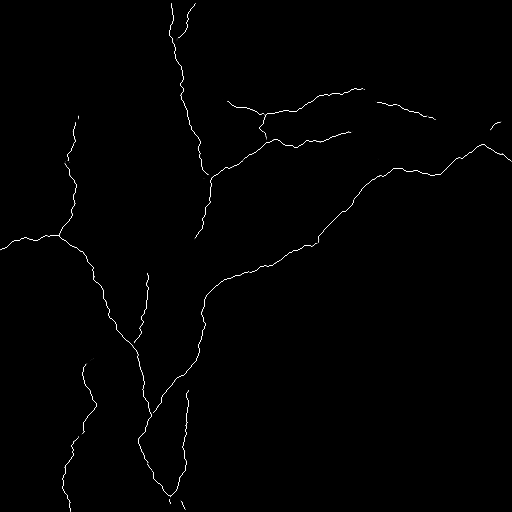} &
 
    \includegraphics[width=0.15\linewidth]{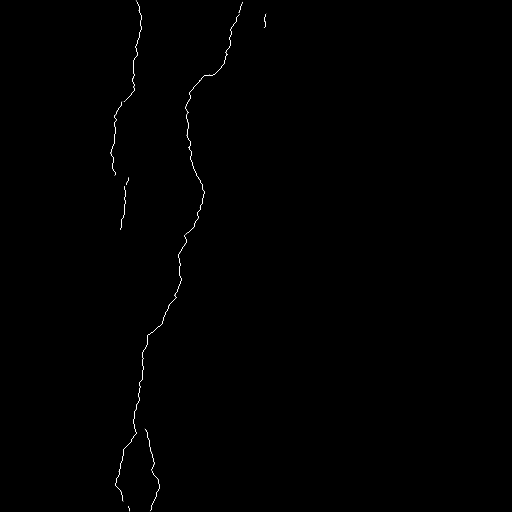} &

    \includegraphics[width=0.15\linewidth]{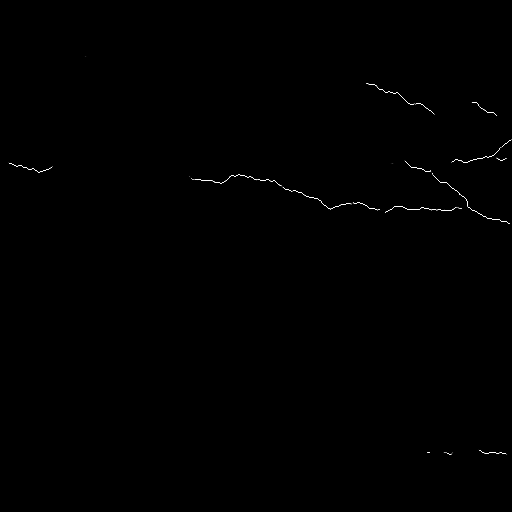} 
    \\

    \raisebox{1.5\normalbaselineskip}[0pt][0pt]{\rotatebox[origin=c]{0}{(g)}} &  
    \includegraphics[width=0.15\linewidth]{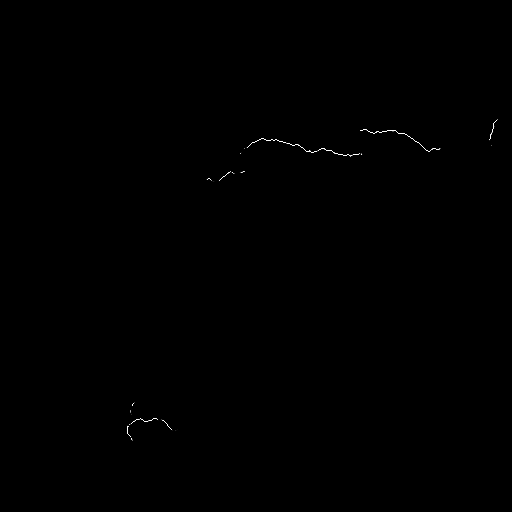} &
    \includegraphics[width=0.15\linewidth]{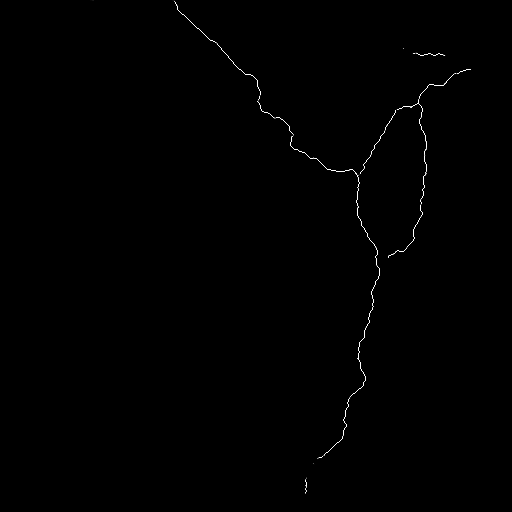} & 

    \includegraphics[width=0.15\linewidth]{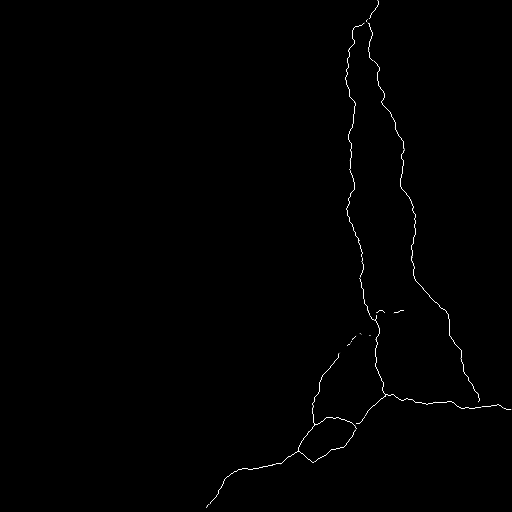} &

    \includegraphics[width=0.15\linewidth]{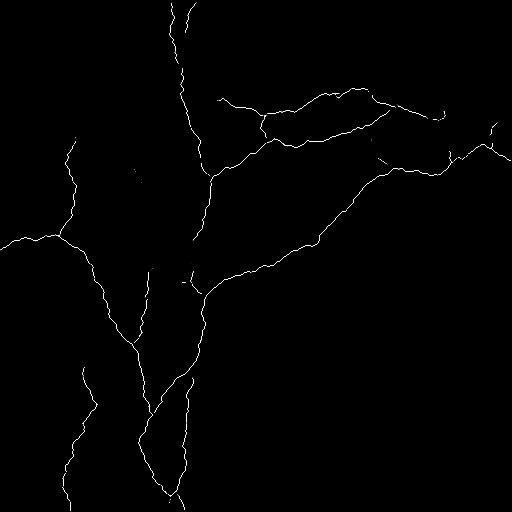} &

    \includegraphics[width=0.15\linewidth]{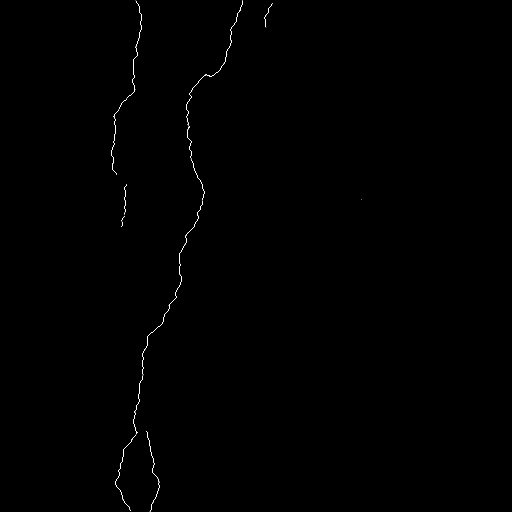} &
 
    \includegraphics[width=0.15\linewidth]{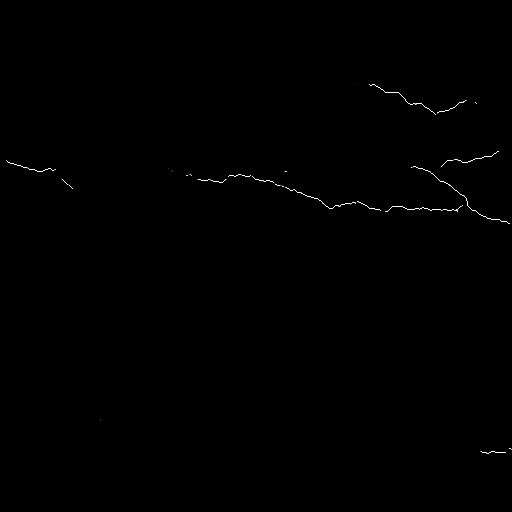}
    \\
    
    \raisebox{1.5\normalbaselineskip}[0pt][0pt]{\rotatebox[origin=c]{0}{(h)}} &  
    \includegraphics[width=0.15\linewidth]{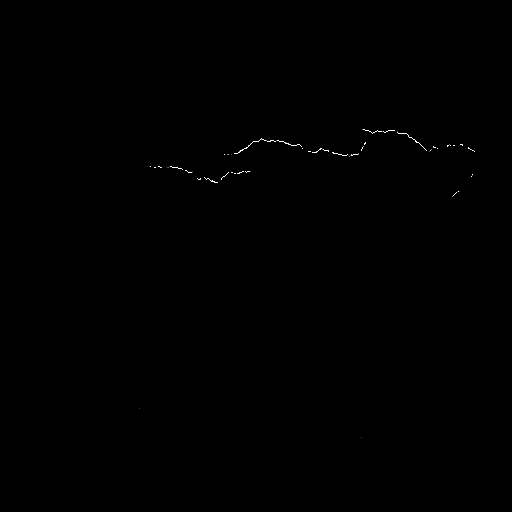} &
    \includegraphics[width=0.15\linewidth]{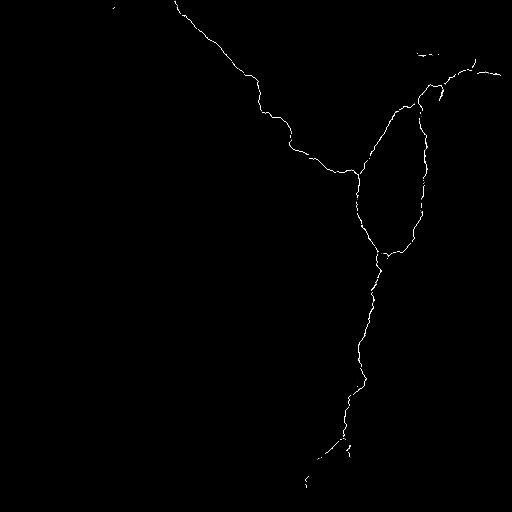} & 

    \includegraphics[width=0.15\linewidth]{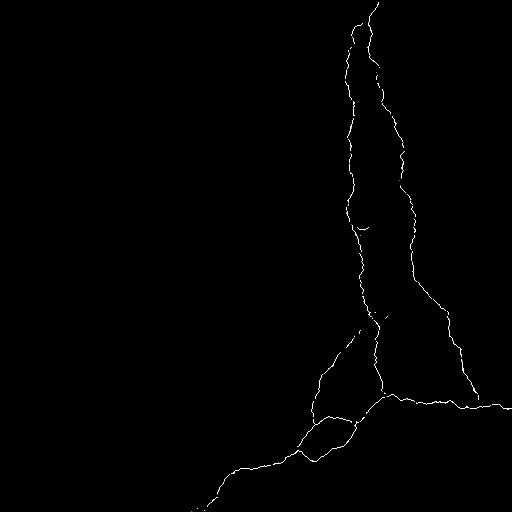} &

    \includegraphics[width=0.15\linewidth]{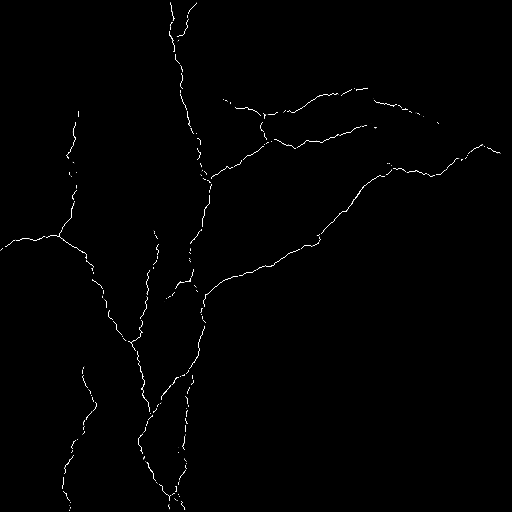} &
 
    \includegraphics[width=0.15\linewidth]{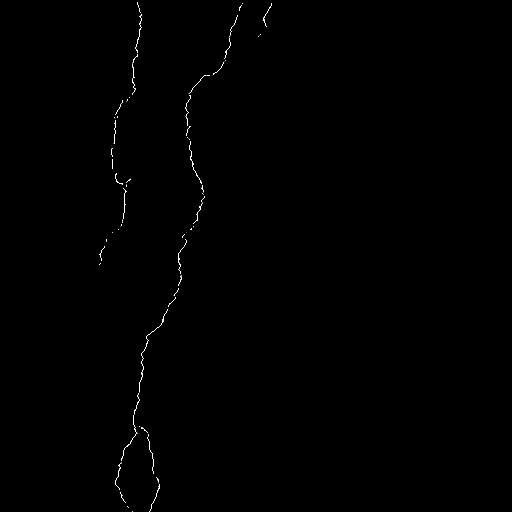} &
  
    \includegraphics[width=0.15\linewidth]{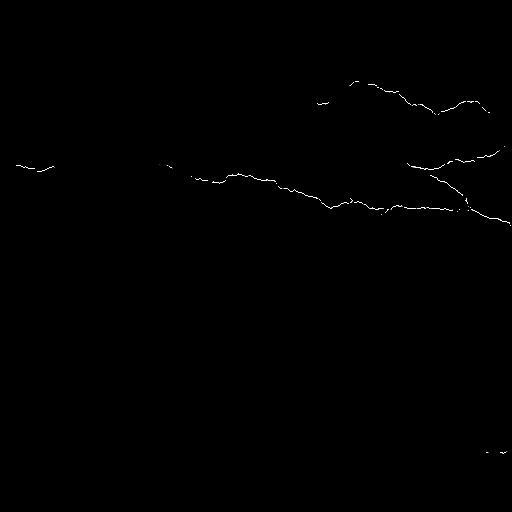} 
    \\
\end{tabular}
\caption{ \footnotesize {\textbf{Visual results on CrackLS315.} {\small (a) Input image. (b) Ground truth. (c) UNet. (d) CrackformerII. (e) DeepCrack. (f) cGAN\_CBAM (pixel). (g) cGAN\_CBAM\_Ig (pixel). (h)cGAN\_LSA (pixel).}}
}
\label{fig:visual_crackls315}
\end{figure}

\begin{figure}
\centering
\footnotesize
\renewcommand{\tabcolsep}{1pt} 
\renewcommand{\arraystretch}{0.2} 
\begin{tabular}{ccccccc}
    \raisebox{1.5\normalbaselineskip}[0pt][0pt]{\rotatebox[origin=c]{0}{(a)}} &  
  
    \includegraphics[width=0.15\linewidth]{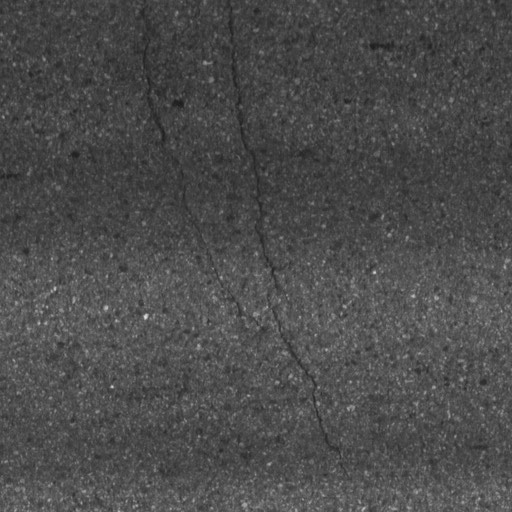} & 
    \includegraphics[width=0.15\linewidth]{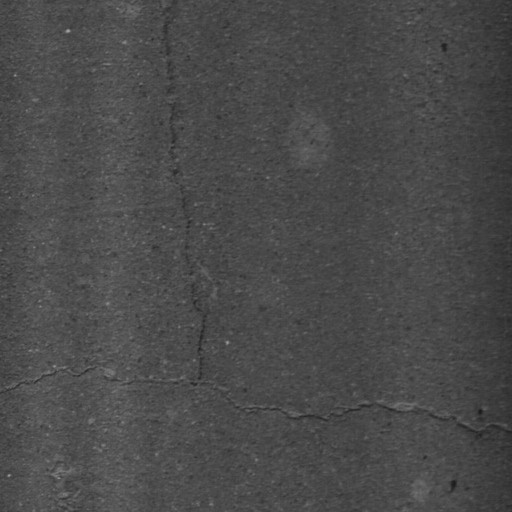} &
    \includegraphics[width=0.15\linewidth]{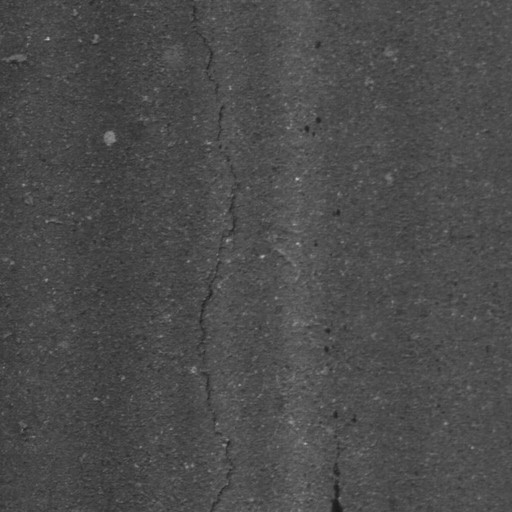} & 
    \includegraphics[width=0.15\linewidth]{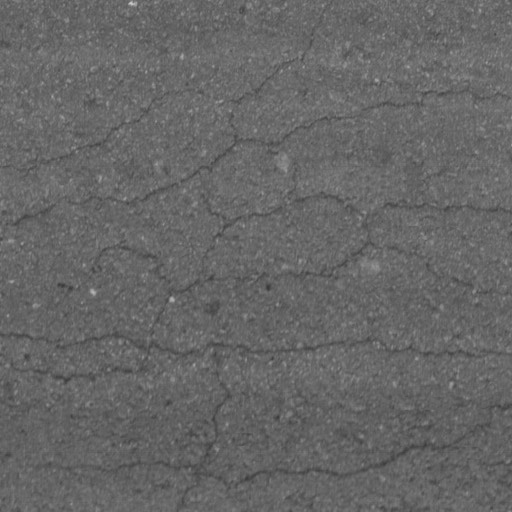} &
    \includegraphics[width=0.15\linewidth]{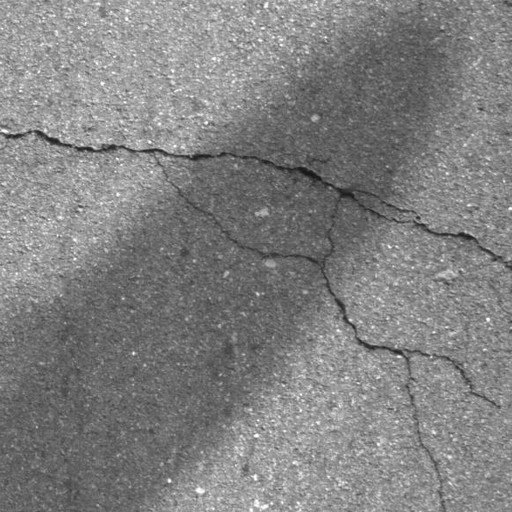} &
    \includegraphics[width=0.15\linewidth]{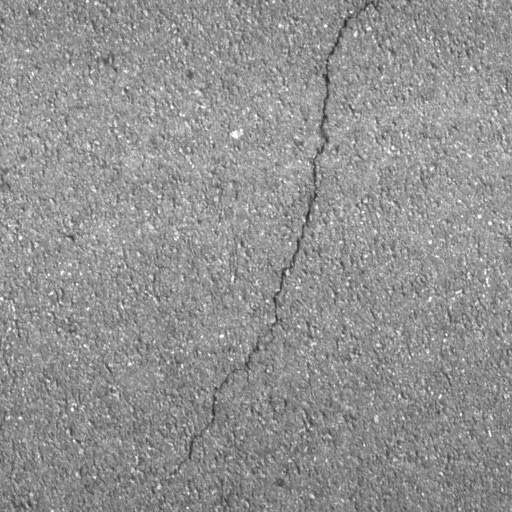} 
    \\

    \raisebox{1.5\normalbaselineskip}[0pt][0pt]{\rotatebox[origin=c]{0}{(b)}} &  
    \includegraphics[width=0.15\linewidth]{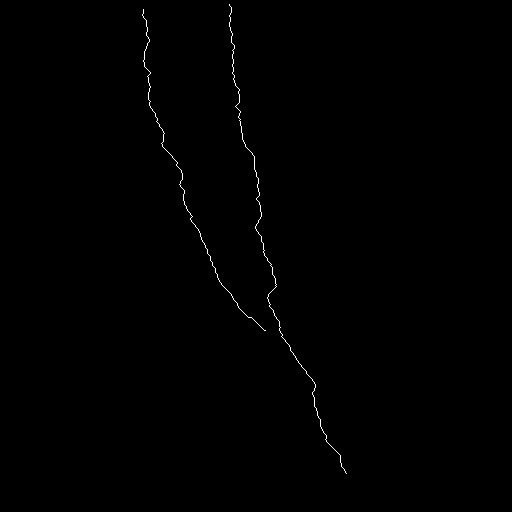} & 
    \includegraphics[width=0.15\linewidth]{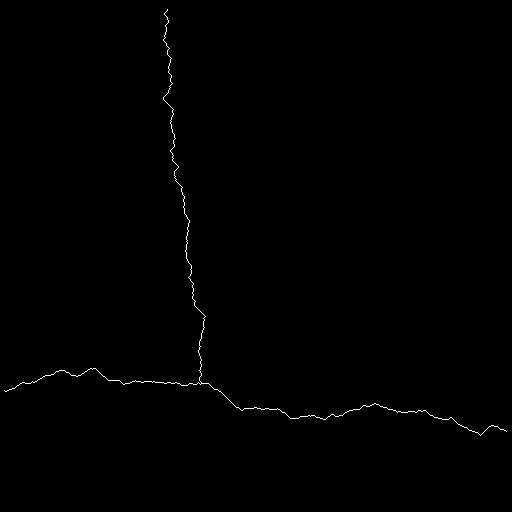} &
    \includegraphics[width=0.15\linewidth]{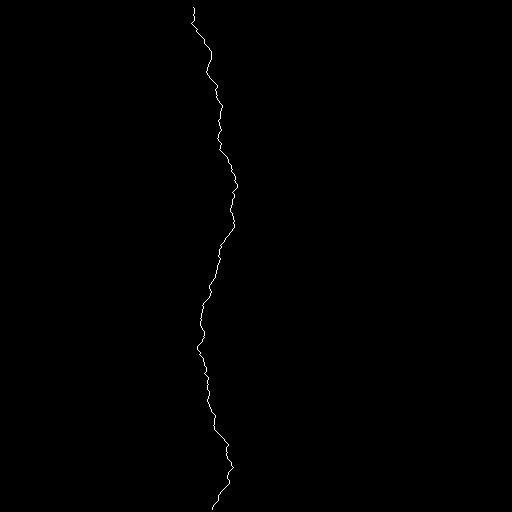} & 
    \includegraphics[width=0.15\linewidth]{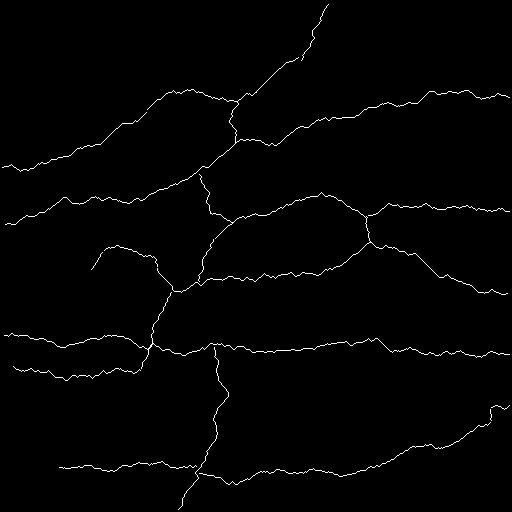} &
    \includegraphics[width=0.15\linewidth]{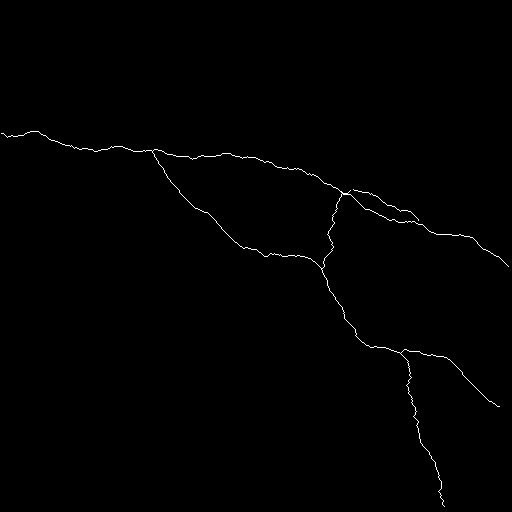} &
    \includegraphics[width=0.15\linewidth]{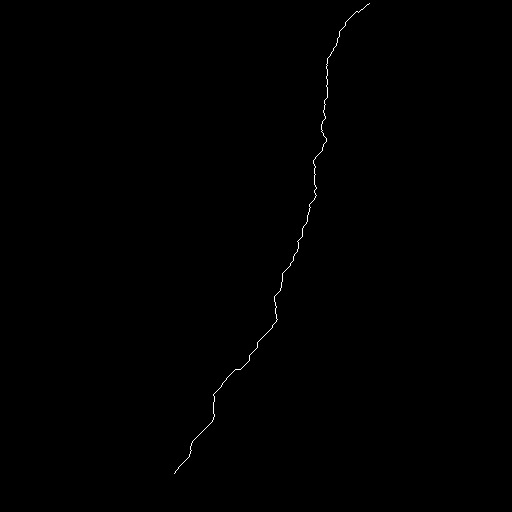} 
    \\

    \raisebox{1.5\normalbaselineskip}[0pt][0pt]{\rotatebox[origin=c]{0}{(c)}} &  
    \includegraphics[width=0.15\linewidth]{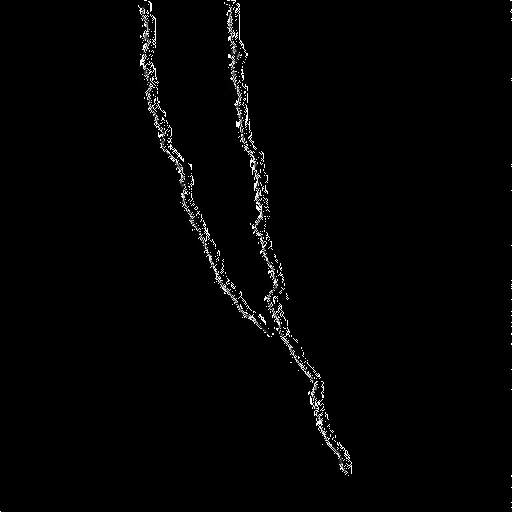} & 
    \includegraphics[width=0.15\linewidth]{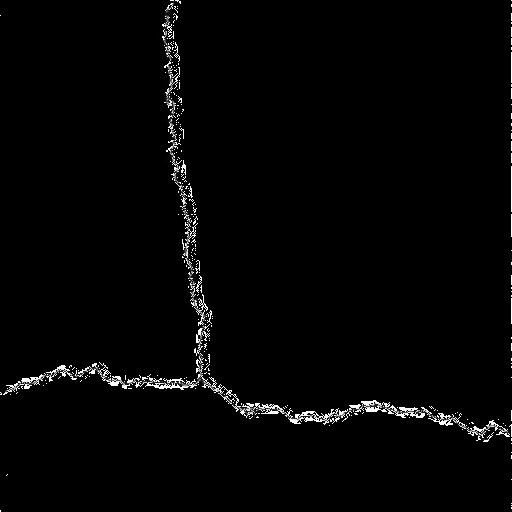} &
    \includegraphics[width=0.15\linewidth]{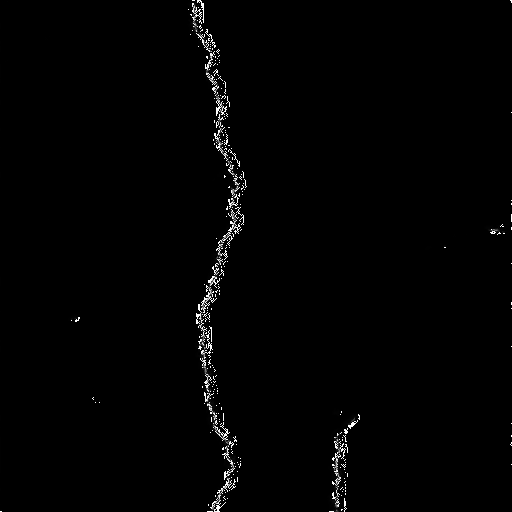} & 
    \includegraphics[width=0.15\linewidth]{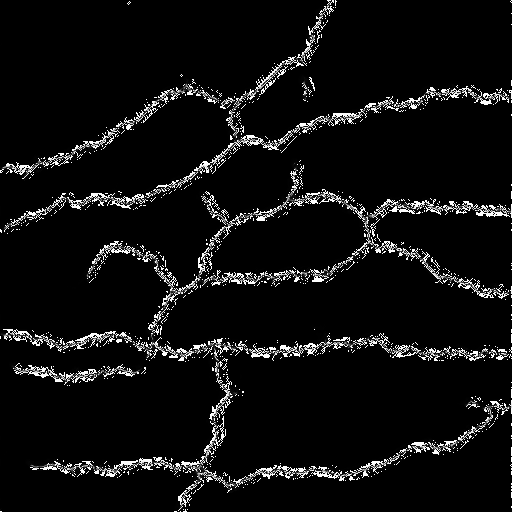}&
    \includegraphics[width=0.15\linewidth]{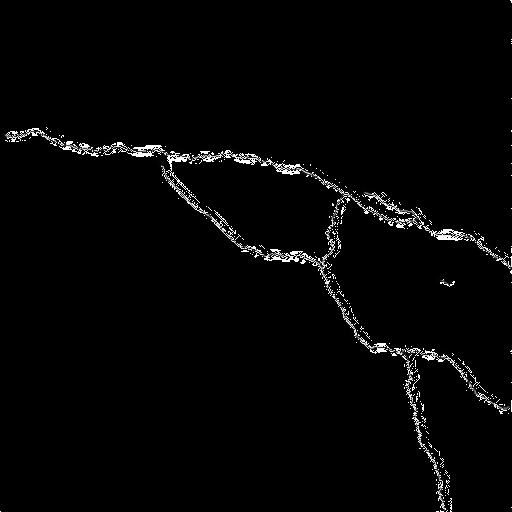} &
    \includegraphics[width=0.15\linewidth]{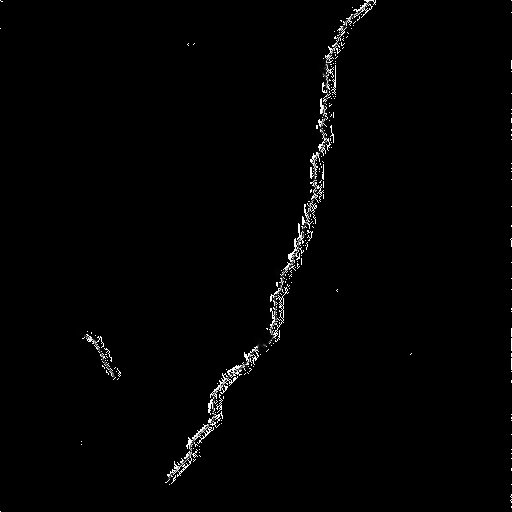}
    \\

    \raisebox{1.5\normalbaselineskip}[0pt][0pt]{\rotatebox[origin=c]{0}{(d)}} &  
    \includegraphics[width=0.15\linewidth]{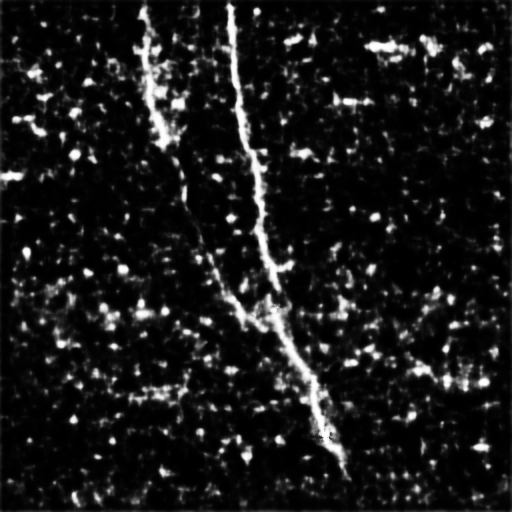} & 
    \includegraphics[width=0.15\linewidth]{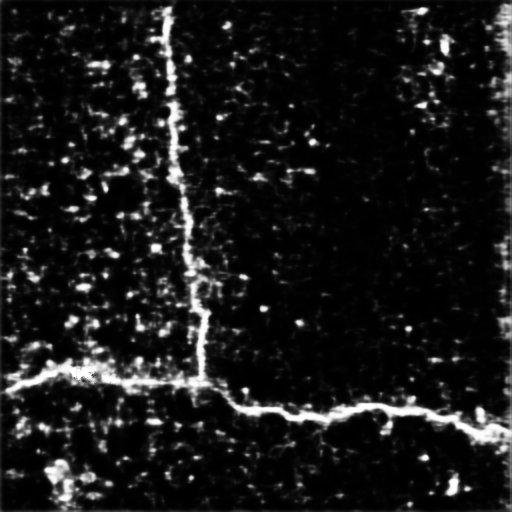} &
    \includegraphics[width=0.15\linewidth]{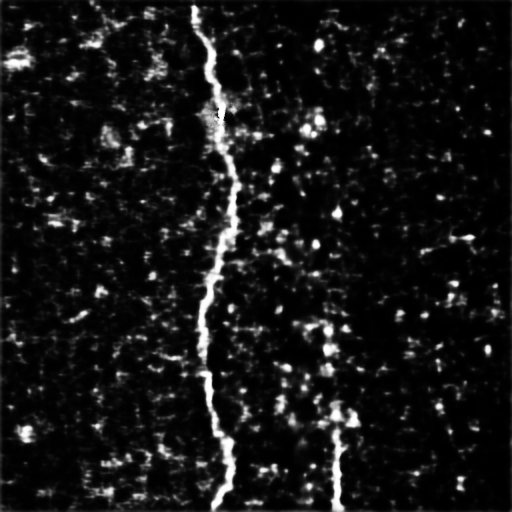} & 
    \includegraphics[width=0.15\linewidth]{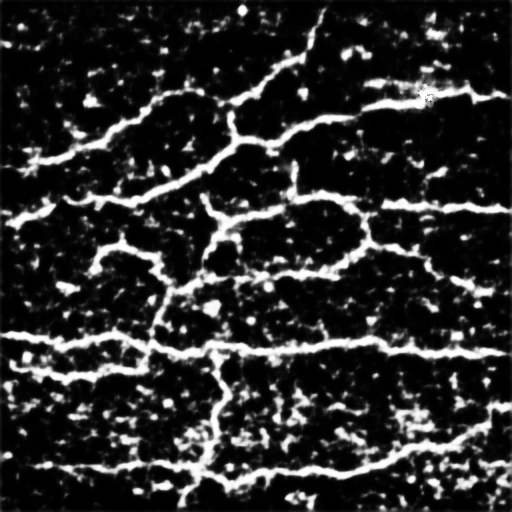} &
    \includegraphics[width=0.15\linewidth]{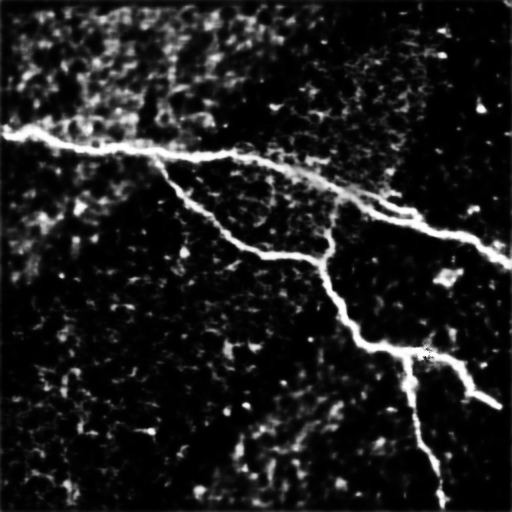} &
    \includegraphics[width=0.15\linewidth]{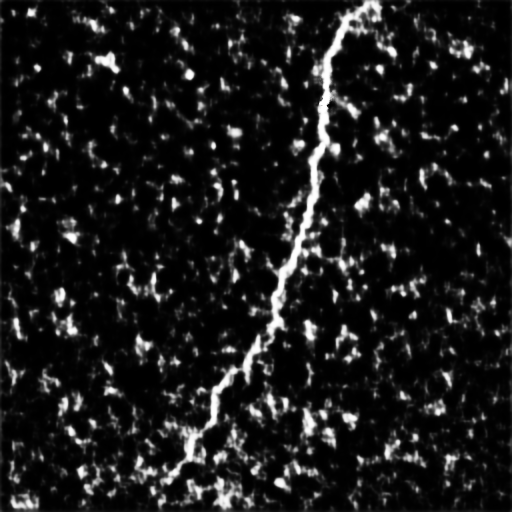} 
    \\

    \raisebox{1.5\normalbaselineskip}[0pt][0pt]{\rotatebox[origin=c]{0}{(e)}} &  
    \includegraphics[width=0.15\linewidth]{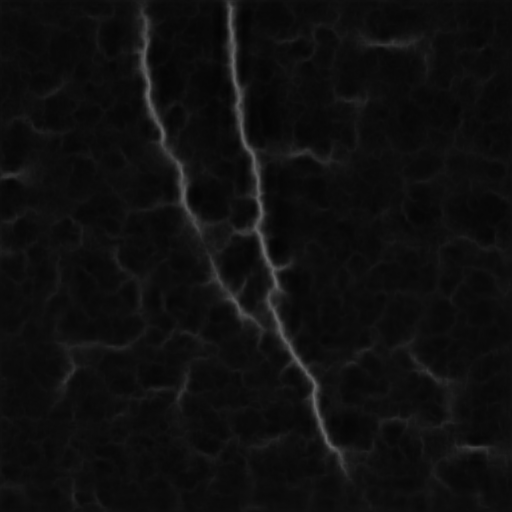} & 
    \includegraphics[width=0.15\linewidth]{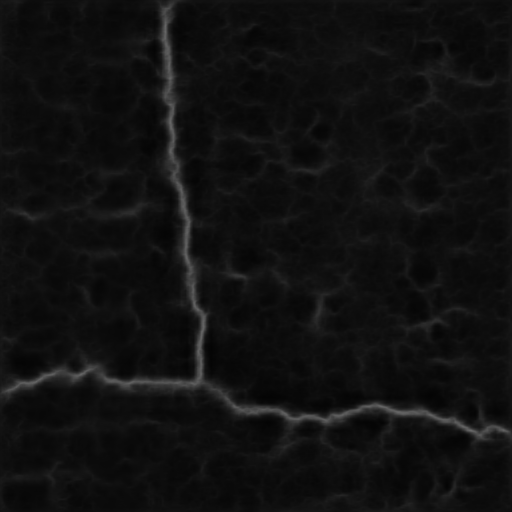} &
    \includegraphics[width=0.15\linewidth]{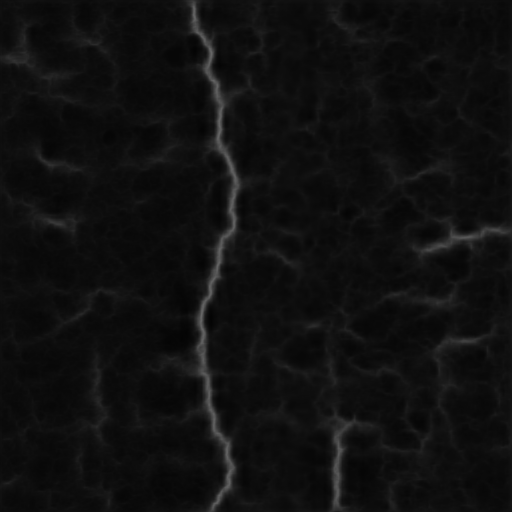} & 
    \includegraphics[width=0.15\linewidth]{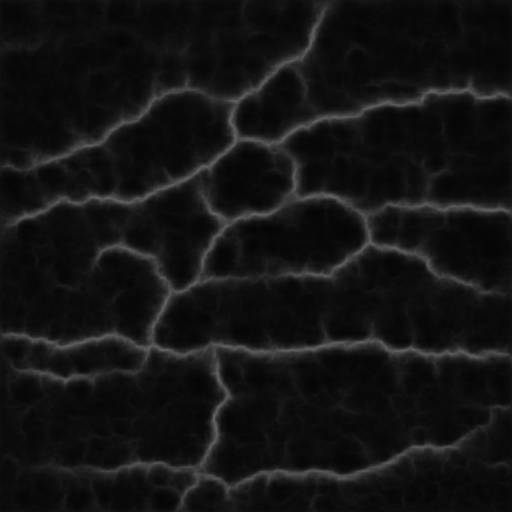}& 
    \includegraphics[width=0.15\linewidth]{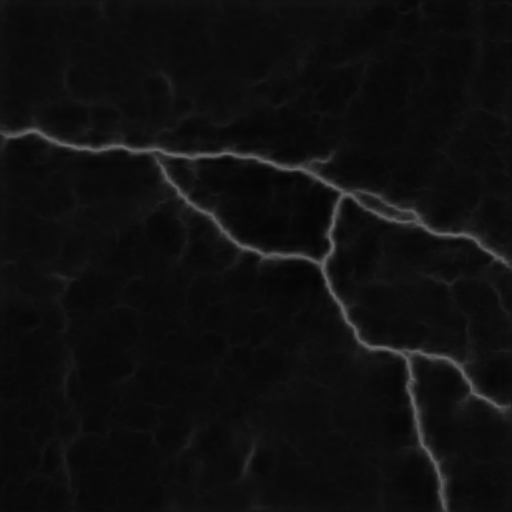} &
    \includegraphics[width=0.15\linewidth]{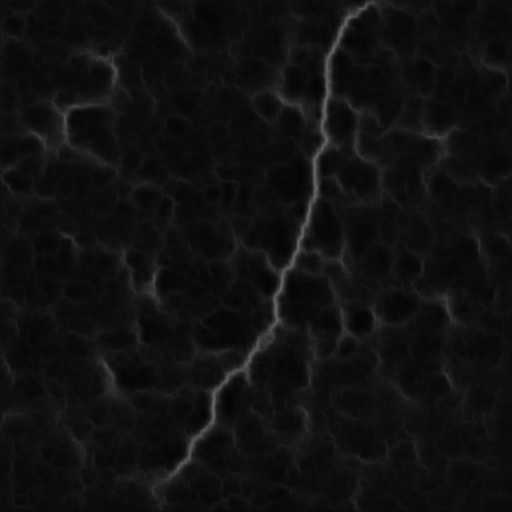}
    \\

    \raisebox{1.5\normalbaselineskip}[0pt][0pt]{\rotatebox[origin=c]{0}{(f))}} &  
    \includegraphics[width=0.15\linewidth]{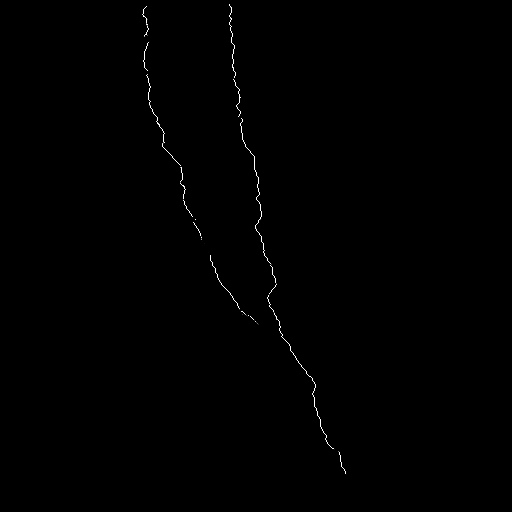} & 
    \includegraphics[width=0.15\linewidth]{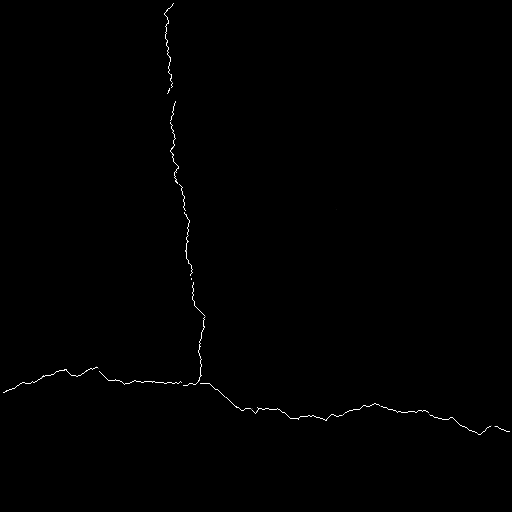} &
    \includegraphics[width=0.15\linewidth]{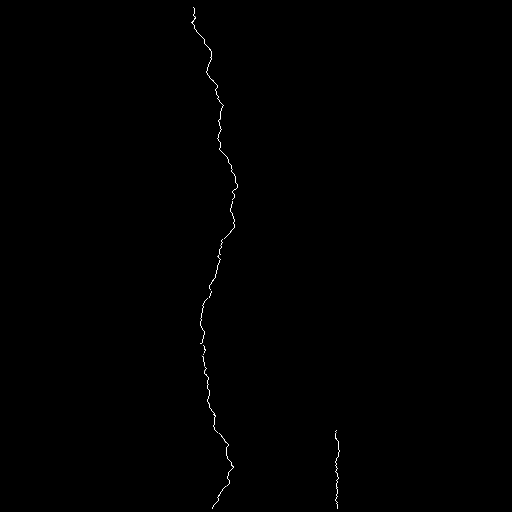} & 
    \includegraphics[width=0.15\linewidth]{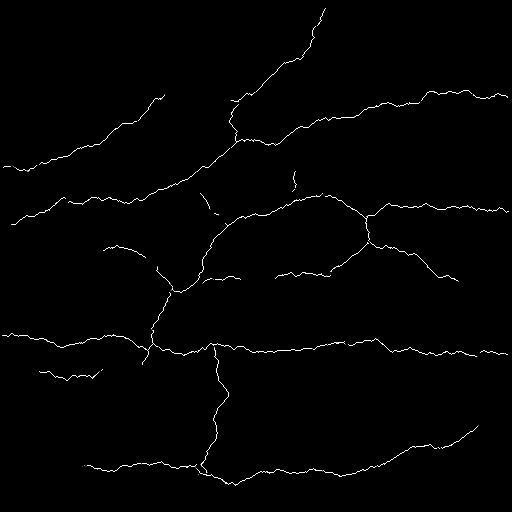} &
    \includegraphics[width=0.15\linewidth]{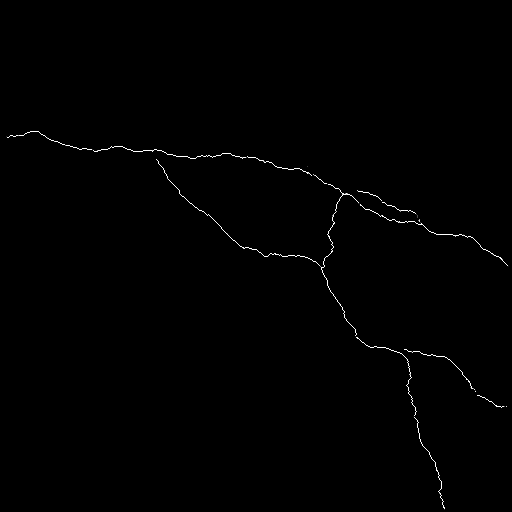} &
    \includegraphics[width=0.15\linewidth]{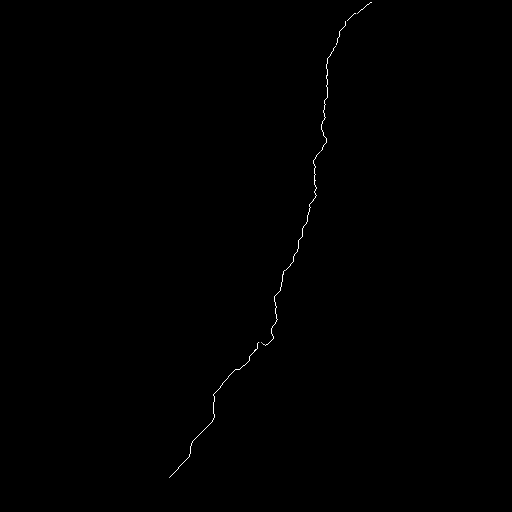} 
    \\

    \raisebox{1.5\normalbaselineskip}[0pt][0pt]{\rotatebox[origin=c]{0}{(g)}} &  

    \includegraphics[width=0.15\linewidth]{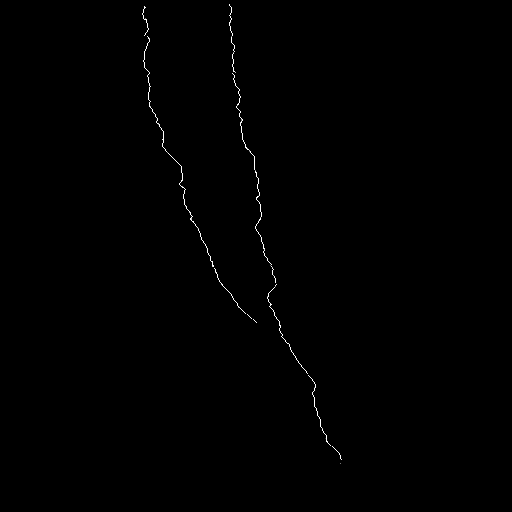} & 
    \includegraphics[width=0.15\linewidth]{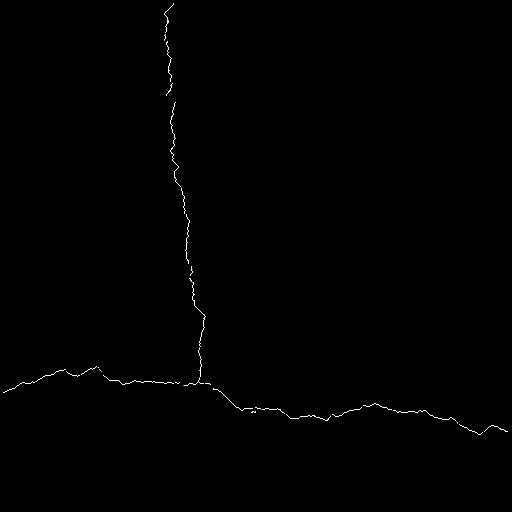} &
    \includegraphics[width=0.15\linewidth]{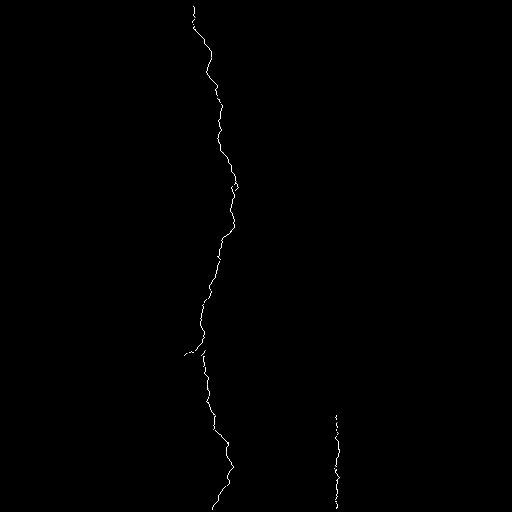} & 
    \includegraphics[width=0.15\linewidth]{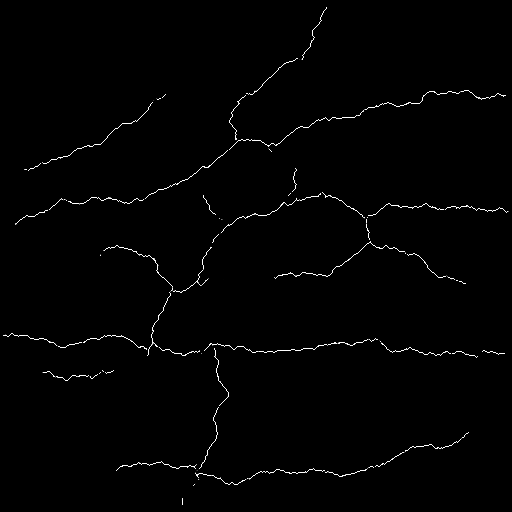} &
    \includegraphics[width=0.15\linewidth]{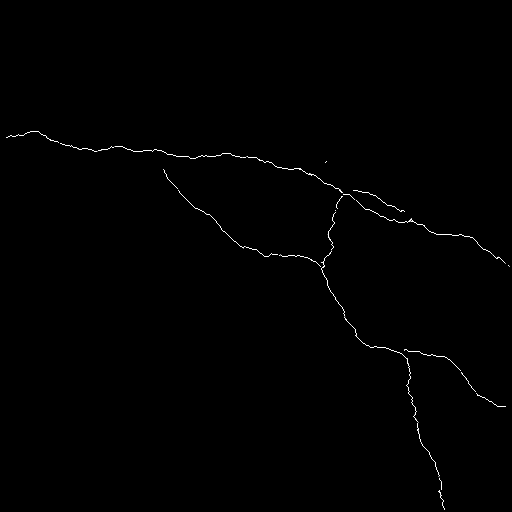} &
    \includegraphics[width=0.15\linewidth]{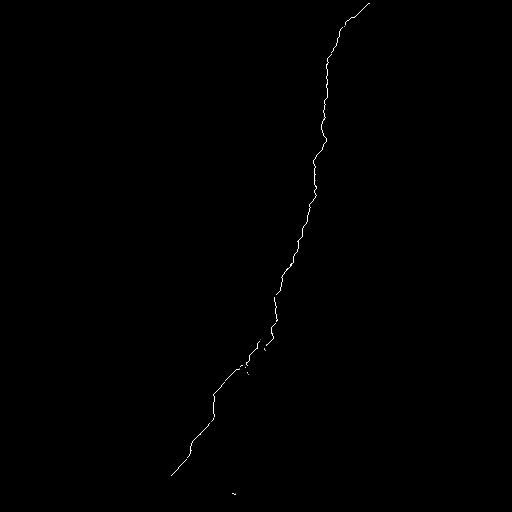}
    
    \\

    \raisebox{1.5\normalbaselineskip}[0pt][0pt]{\rotatebox[origin=c]{0}{(h))}} &  
    \includegraphics[width=0.15\linewidth]{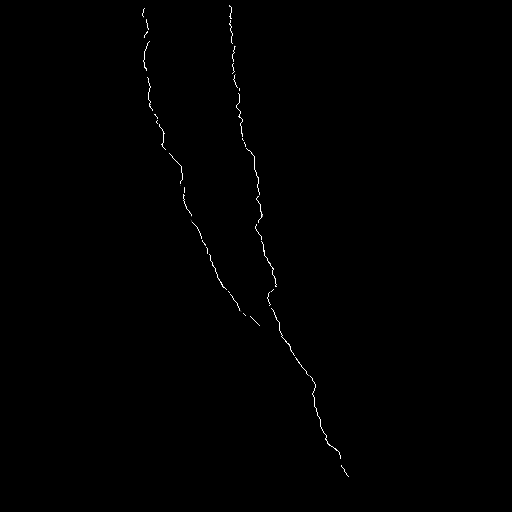} & 
    \includegraphics[width=0.15\linewidth]{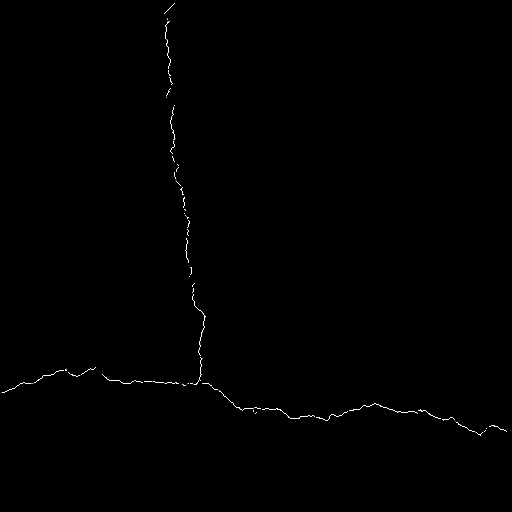} &
    \includegraphics[width=0.15\linewidth]{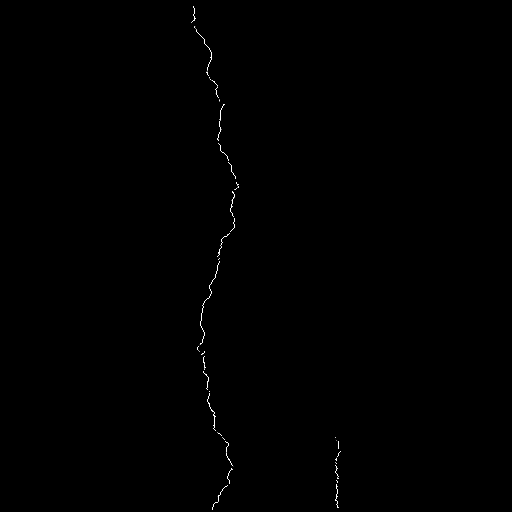} & 
    \includegraphics[width=0.15\linewidth]{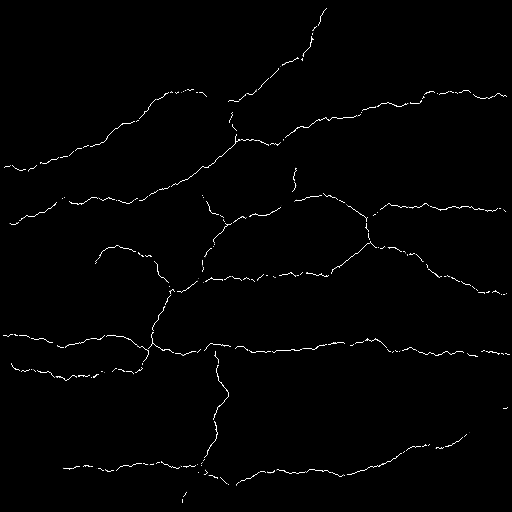} &
    \includegraphics[width=0.15\linewidth]{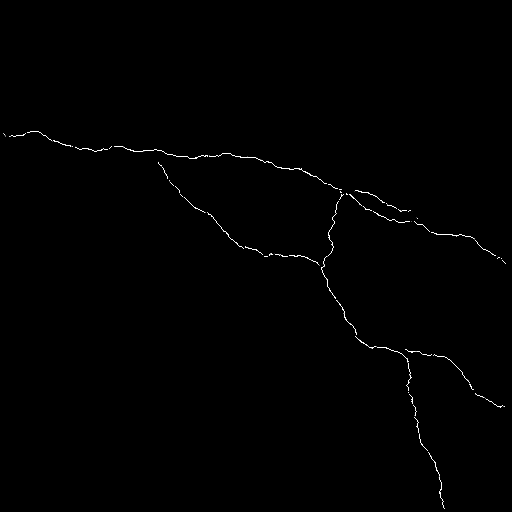} &
    \includegraphics[width=0.15\linewidth]{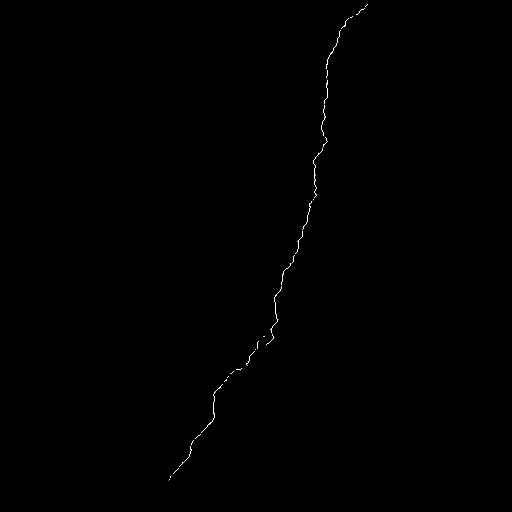} 
    \\
\end{tabular}
\caption{\footnotesize {\textbf{Visual results on CRKWH100.} (a) Input image. (b) Ground truth. (c) UNet. (d) CrackformerII. (e) DeepCrack. (f) cGAN\_CBAM (pixel). (g) cGAN\_CBAM\_Ig (pixel). (h)cGAN\_LSA (pixel)}.
}
\label{fig:visual_crkwh100}
\end{figure}

\begin{figure}
\centering
\footnotesize
\renewcommand{\tabcolsep}{1pt} 
\renewcommand{\arraystretch}{0.2} 
\begin{tabular}{ccccccc}
    \raisebox{1.5\normalbaselineskip}[0pt][0pt]{\rotatebox[origin=c]{0}{(a)}} &  
    \includegraphics[width=0.15\linewidth]{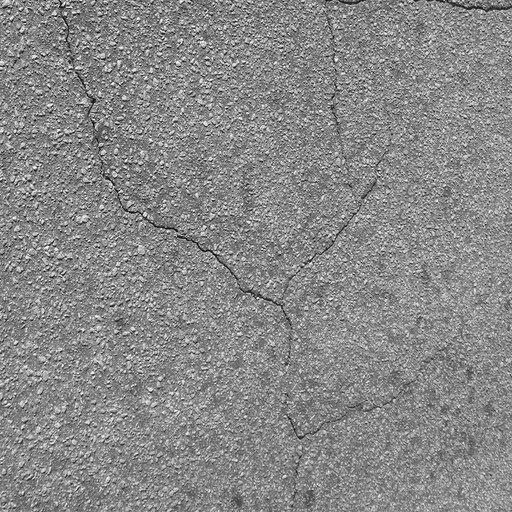} & 
    \includegraphics[width=0.15\linewidth]{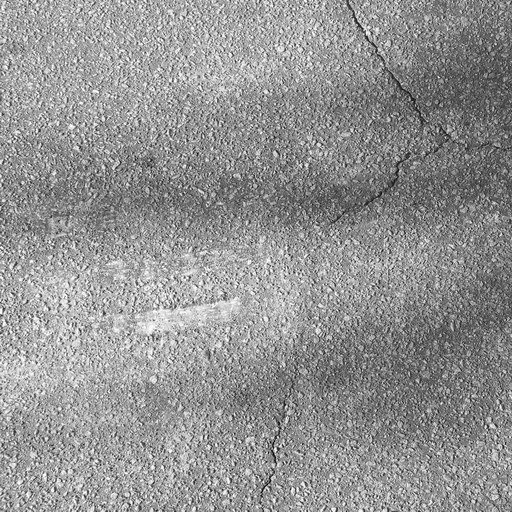} &
    \includegraphics[width=0.15\linewidth]{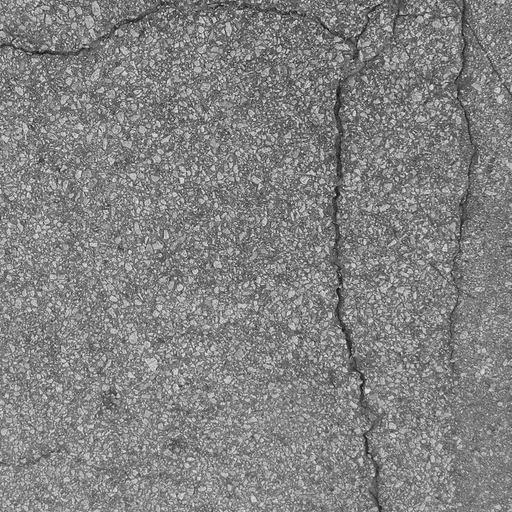} &
    \includegraphics[width=0.15\linewidth]{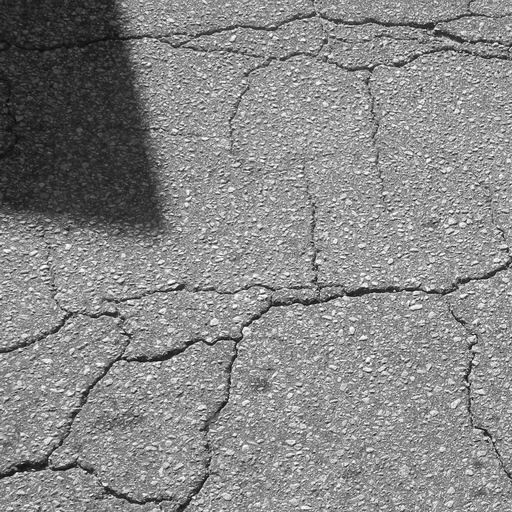} &
    \includegraphics[width=0.15\linewidth]{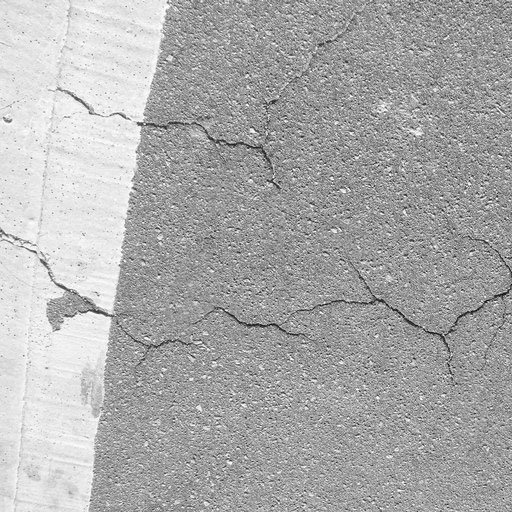} & 
    \includegraphics[width=0.15\linewidth]{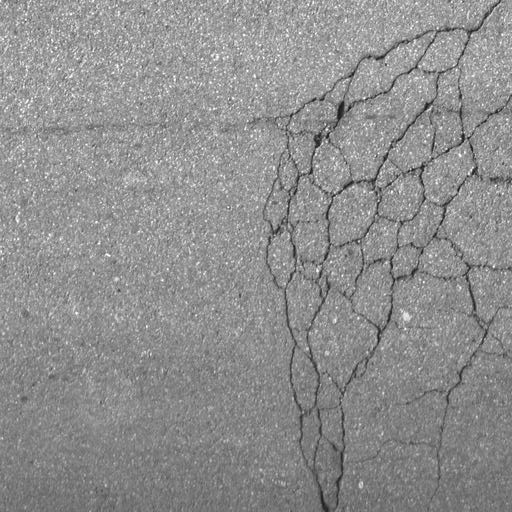} 
    \\

    \raisebox{1.5\normalbaselineskip}[0pt][0pt]{\rotatebox[origin=c]{0}{(b)}} &  
    \includegraphics[width=0.15\linewidth]{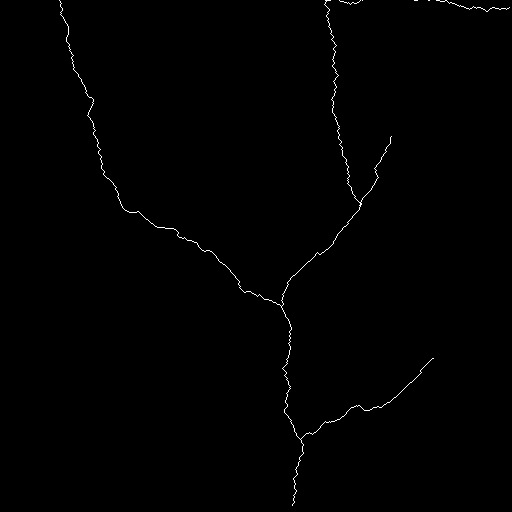} & 
    \includegraphics[width=0.15\linewidth]{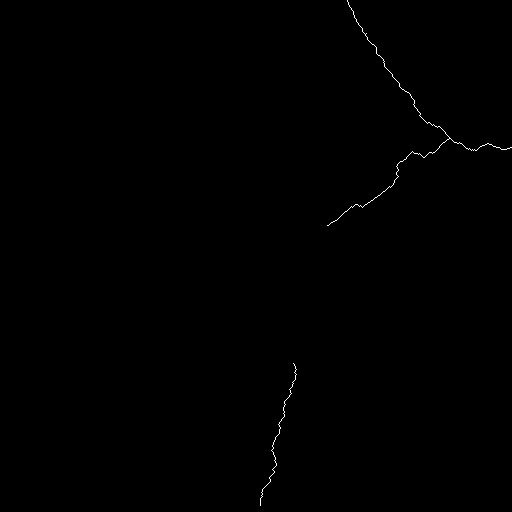} &
    \includegraphics[width=0.15\linewidth]{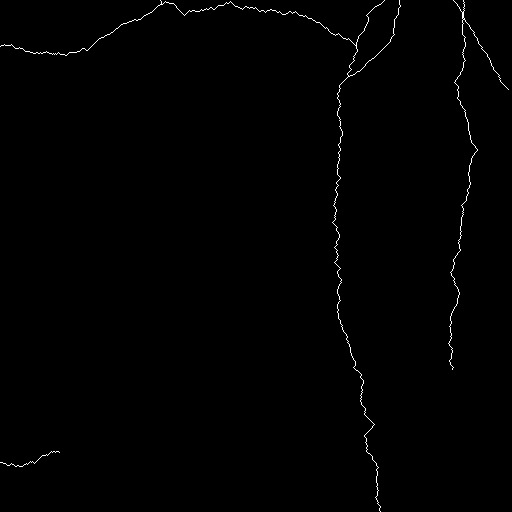} &
    \includegraphics[width=0.15\linewidth]{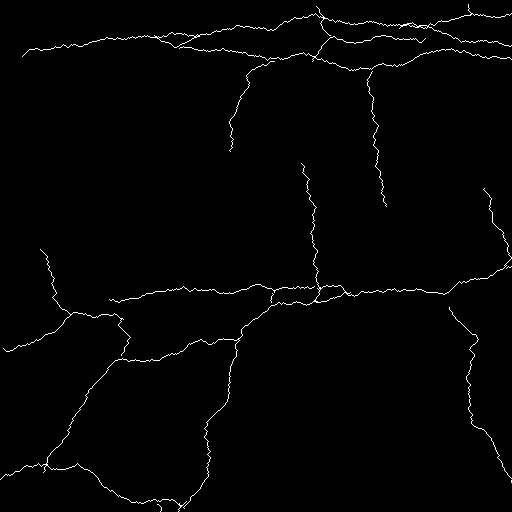} &
    \includegraphics[width=0.15\linewidth]{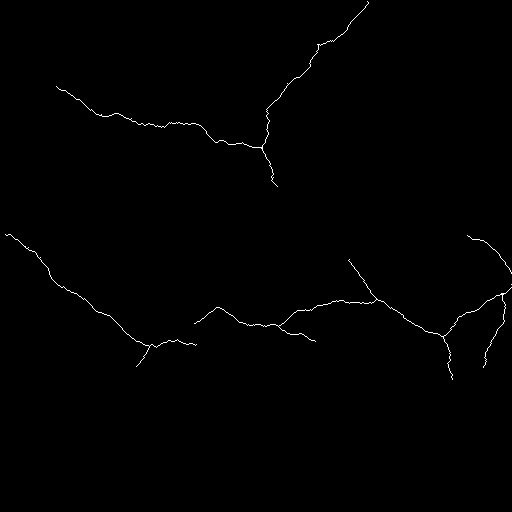} &
    \includegraphics[width=0.15\linewidth]{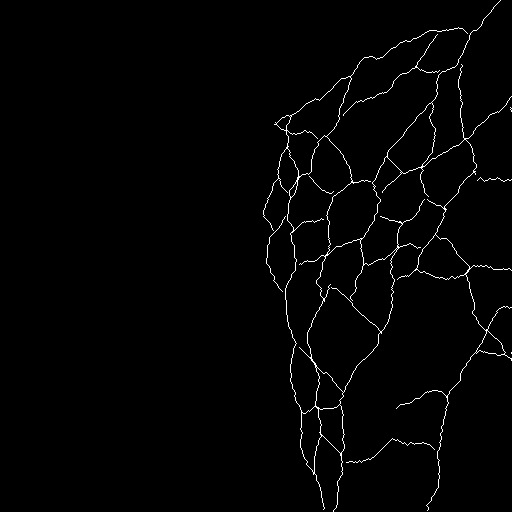} 
    \\
    \raisebox{1.5\normalbaselineskip}[0pt][0pt]{\rotatebox[origin=c]{0}{(c)}} &  
    \includegraphics[width=0.15\linewidth]{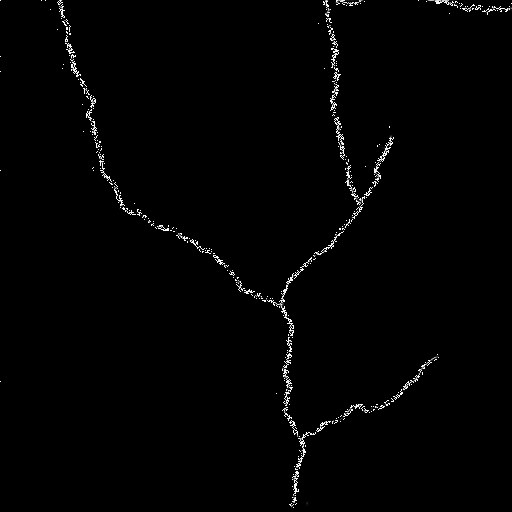} & 
    \includegraphics[width=0.15\linewidth]{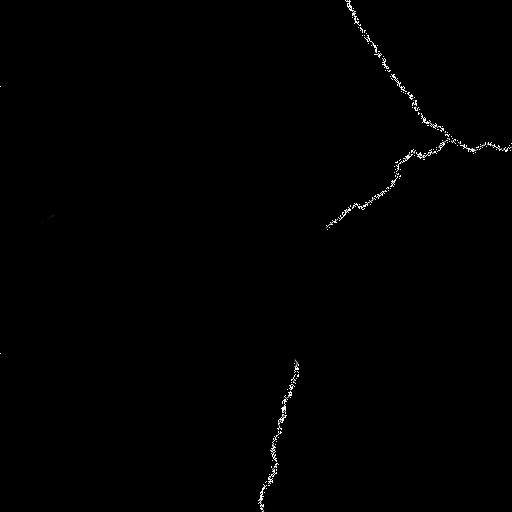} &
    \includegraphics[width=0.15\linewidth]{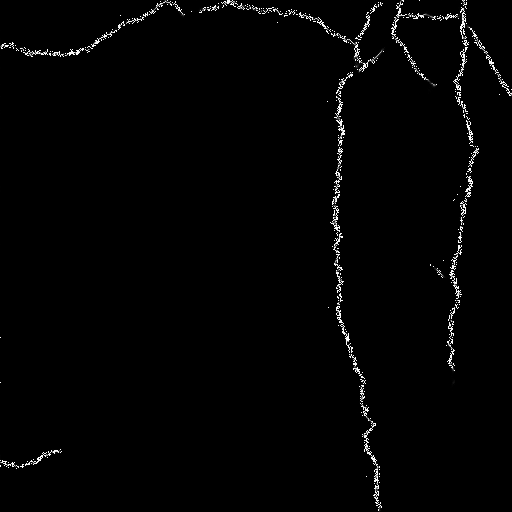} &
    \includegraphics[width=0.15\linewidth]{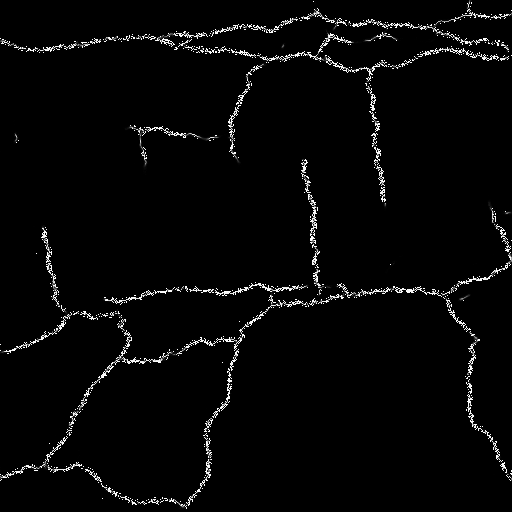} &
    \includegraphics[width=0.15\linewidth]{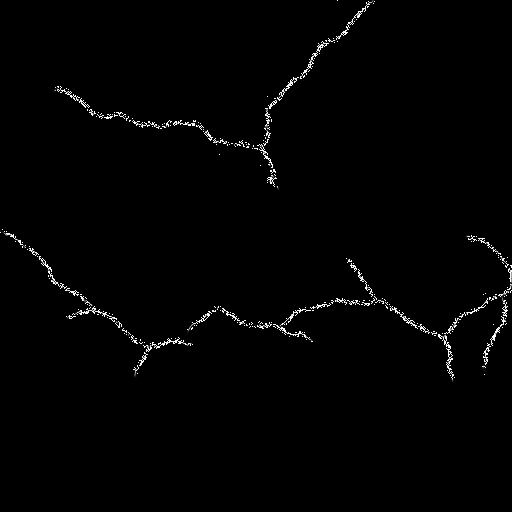}&
    \includegraphics[width=0.15\linewidth]{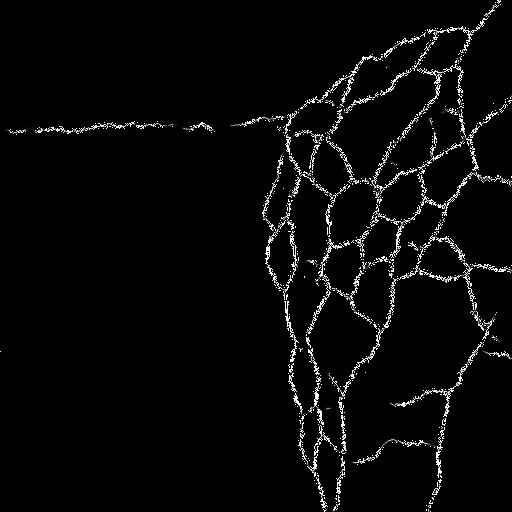}
    \\

    \raisebox{1.5\normalbaselineskip}[0pt][0pt]{\rotatebox[origin=c]{0}{(d)}} &  
    \includegraphics[width=0.15\linewidth]{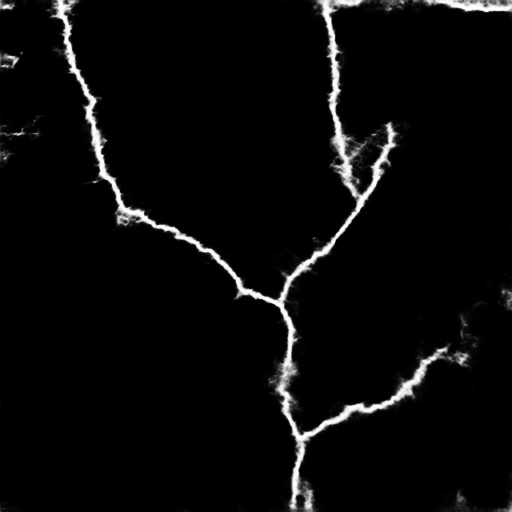} & 
    \includegraphics[width=0.15\linewidth]{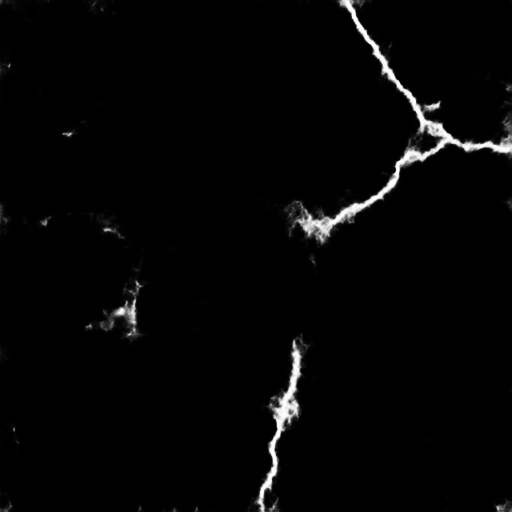} &
    \includegraphics[width=0.15\linewidth]{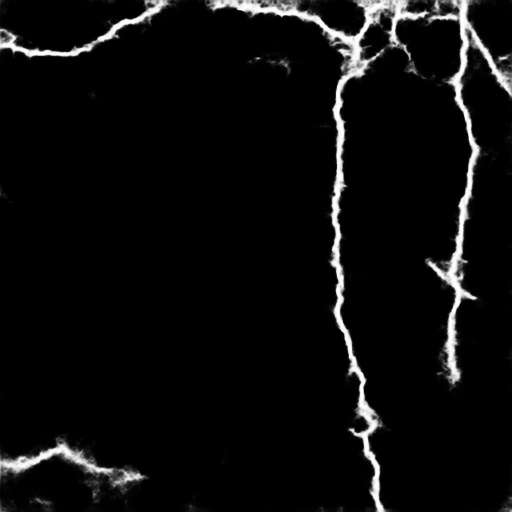} &
    \includegraphics[width=0.15\linewidth]{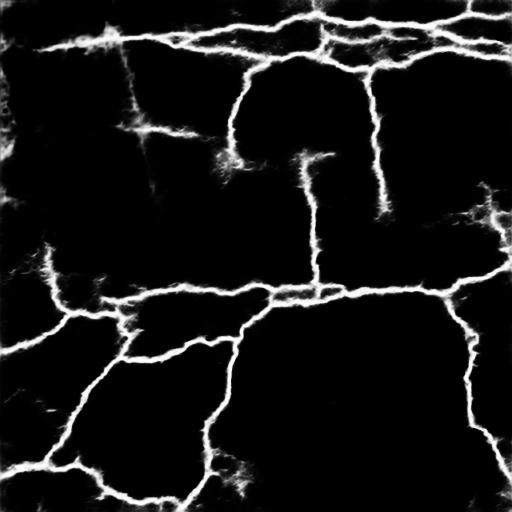} &
    \includegraphics[width=0.15\linewidth]{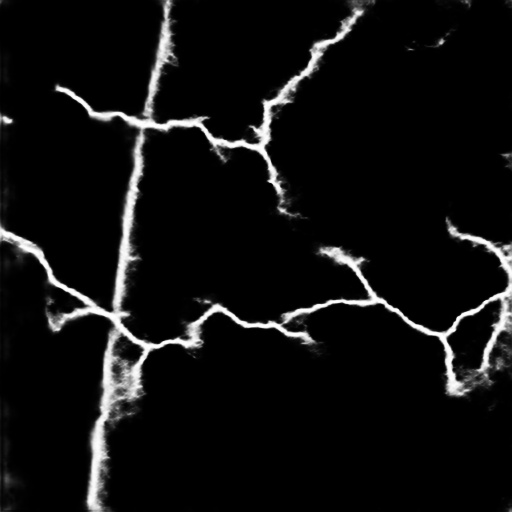} &
    \includegraphics[width=0.15\linewidth]{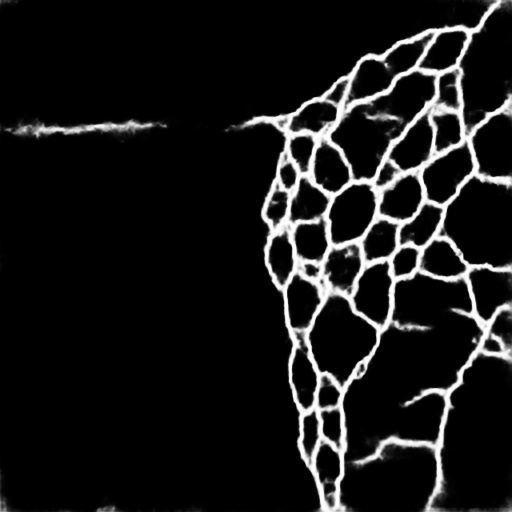} 
    \\

    \raisebox{1.5\normalbaselineskip}[0pt][0pt]{\rotatebox[origin=c]{0}{(e)}} &  
    \includegraphics[width=0.15\linewidth]{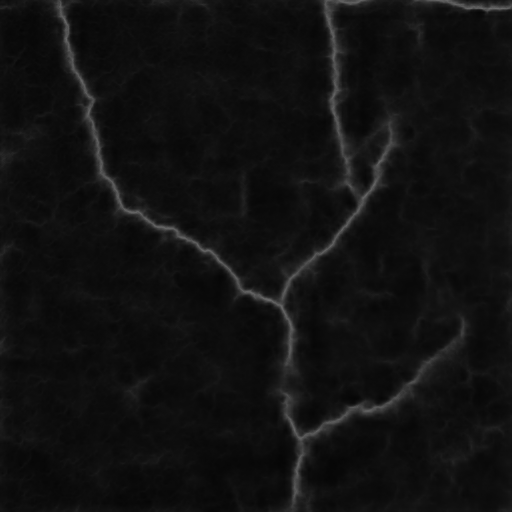} & 
    \includegraphics[width=0.15\linewidth]{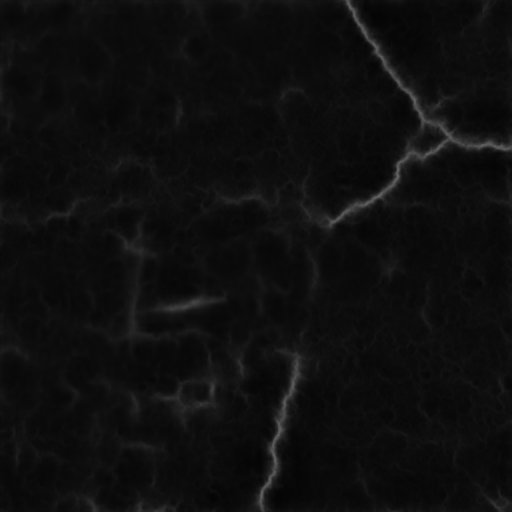} &
    \includegraphics[width=0.15\linewidth]{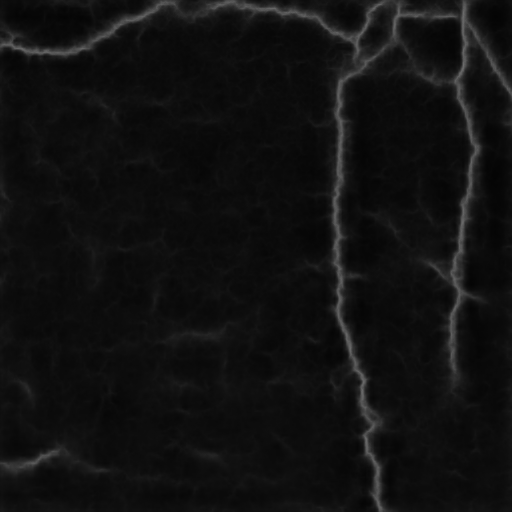} &
    \includegraphics[width=0.15\linewidth]{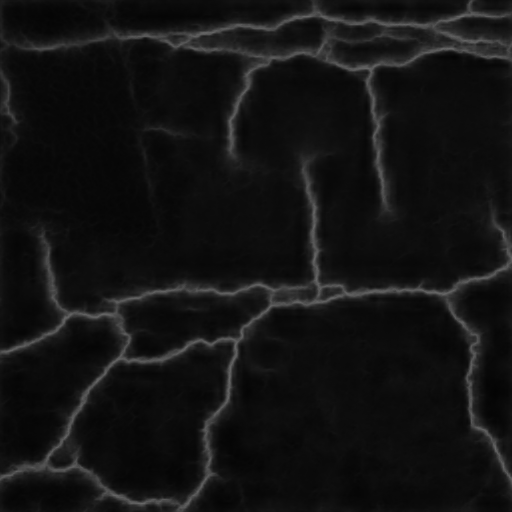} &
    \includegraphics[width=0.15\linewidth]{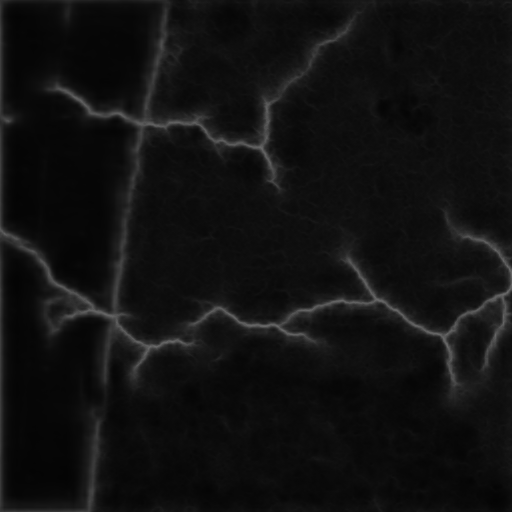}& 
    \includegraphics[width=0.15\linewidth]{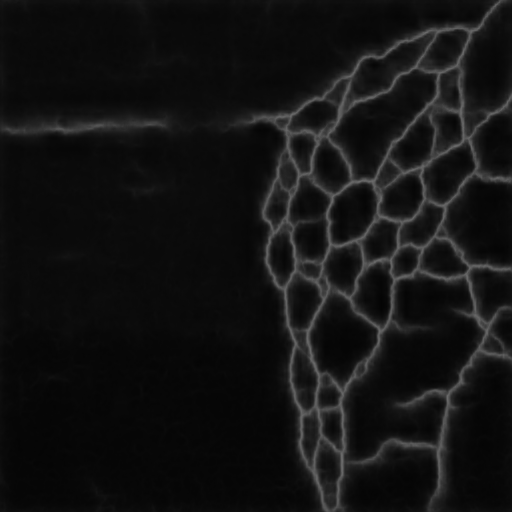}
    \\

    \raisebox{1.5\normalbaselineskip}[0pt][0pt]{\rotatebox[origin=c]{0}{(f)}} &  
    \includegraphics[width=0.15\linewidth]{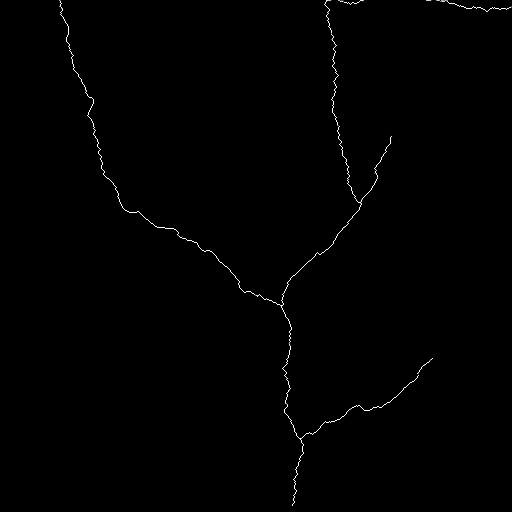} & 
    \includegraphics[width=0.15\linewidth]{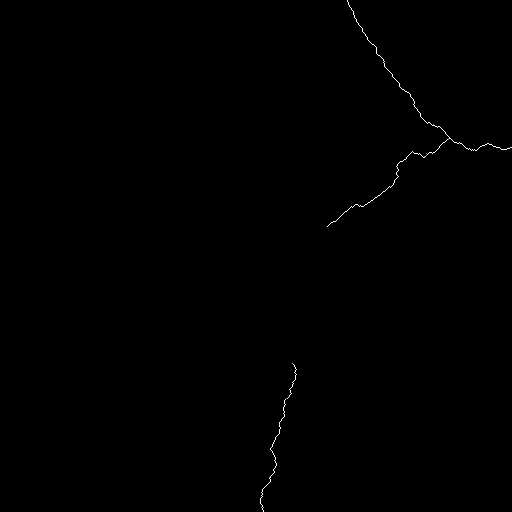} &
    \includegraphics[width=0.15\linewidth]{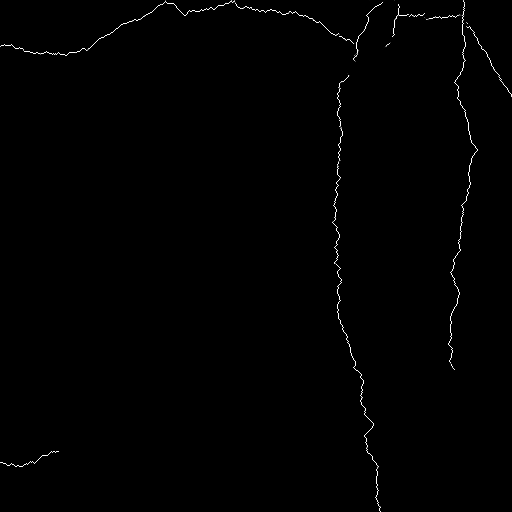} &
    \includegraphics[width=0.15\linewidth]{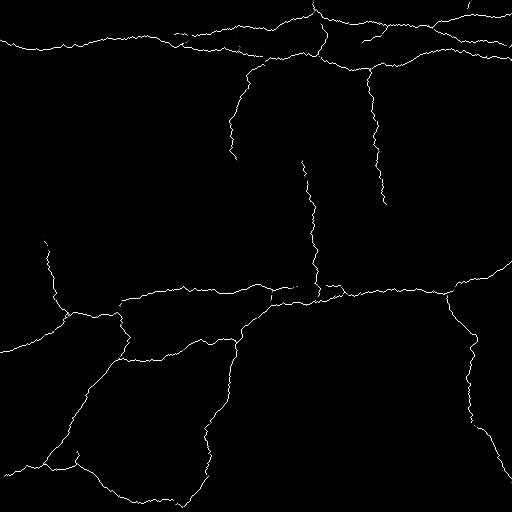} &
    \includegraphics[width=0.15\linewidth]{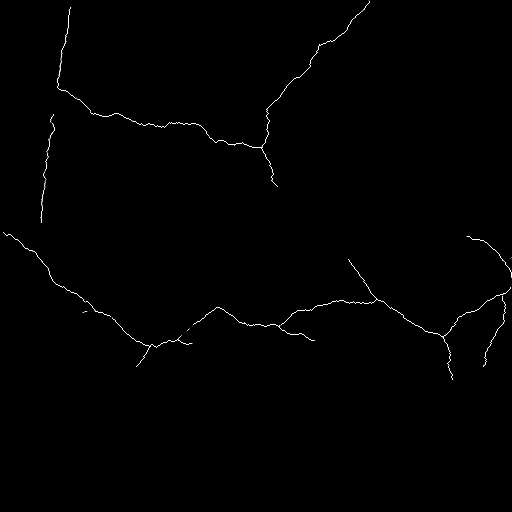} &
    \includegraphics[width=0.15\linewidth]{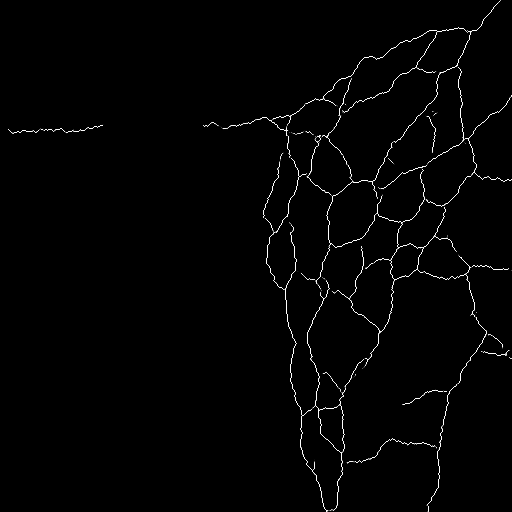} 
    \\

    \raisebox{1.5\normalbaselineskip}[0pt][0pt]{\rotatebox[origin=c]{0}{(g)}} &  
    \includegraphics[width=0.15\linewidth]{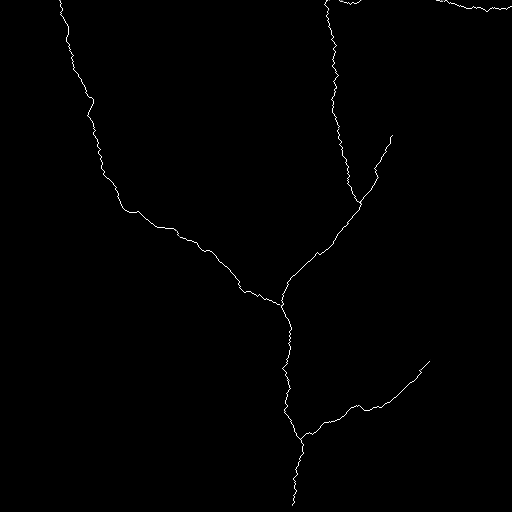} & 
    \includegraphics[width=0.15\linewidth]{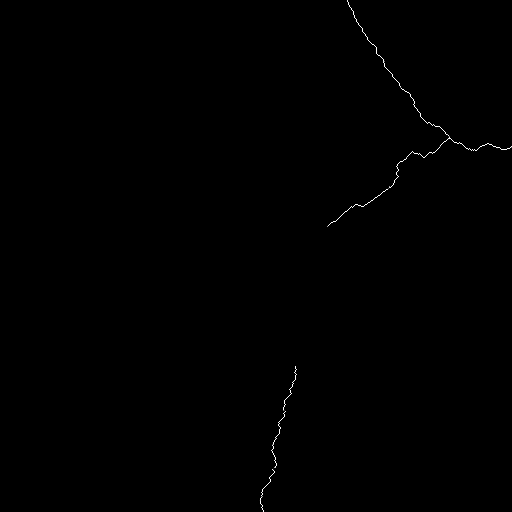} &
    \includegraphics[width=0.15\linewidth]{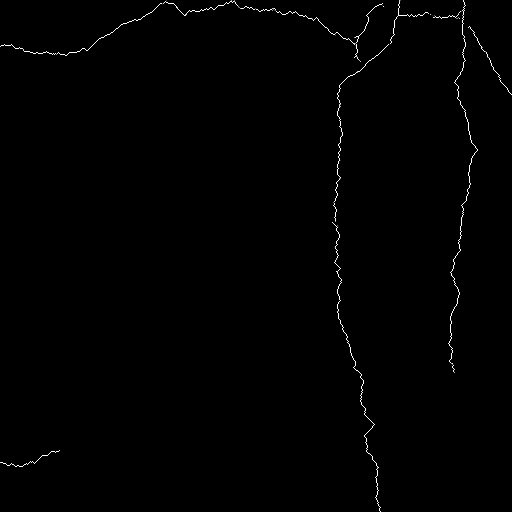} &
    \includegraphics[width=0.15\linewidth]{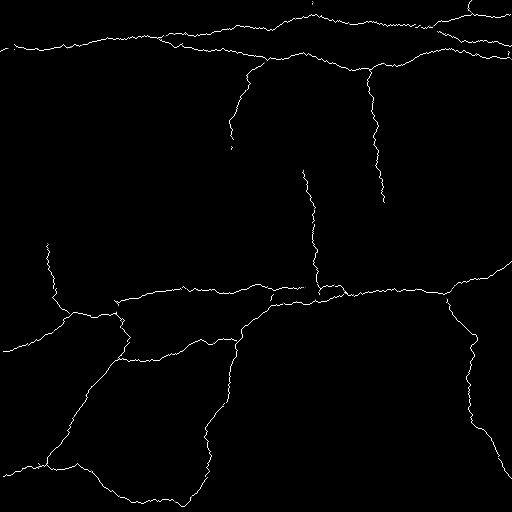} &
    \includegraphics[width=0.15\linewidth]{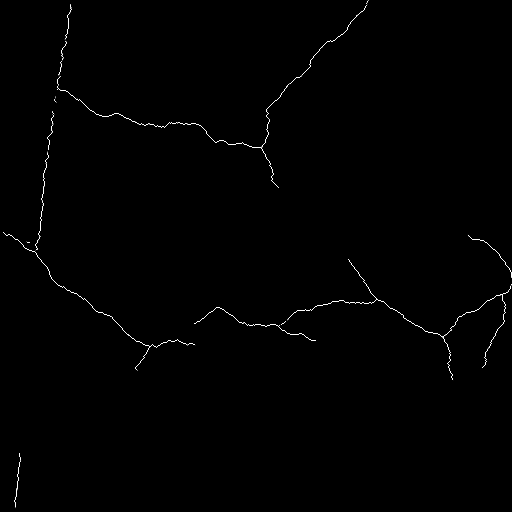} &
    \includegraphics[width=0.15\linewidth]{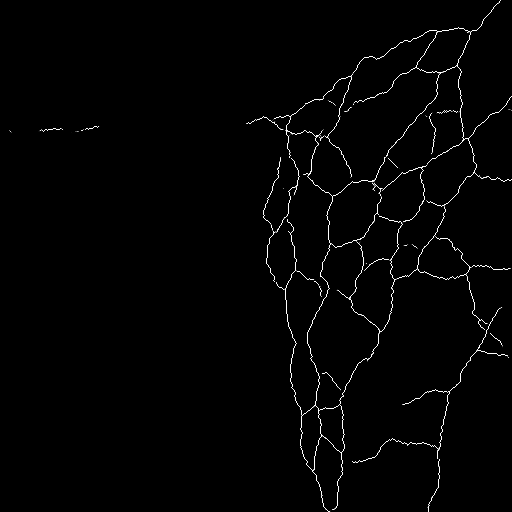} 
    \\

    \raisebox{1.5\normalbaselineskip}[0pt][0pt]{\rotatebox[origin=c]{0}{(h)}} &  
    \includegraphics[width=0.15\linewidth]{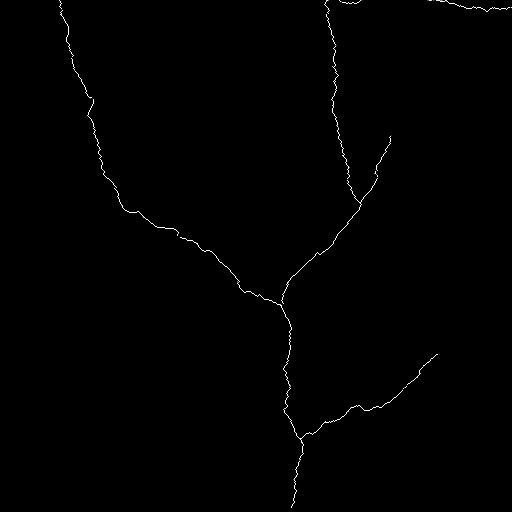} & 
    \includegraphics[width=0.15\linewidth]{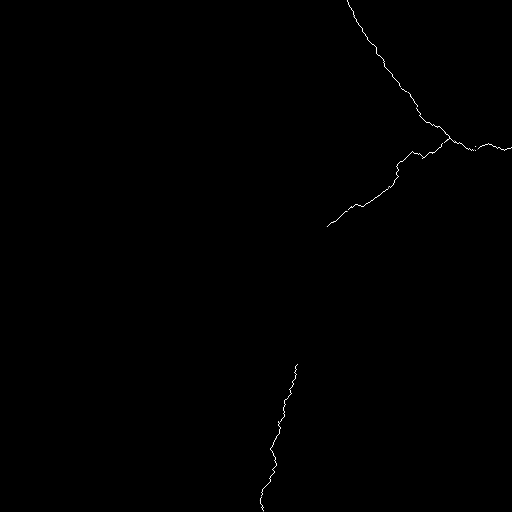} &
    \includegraphics[width=0.15\linewidth]{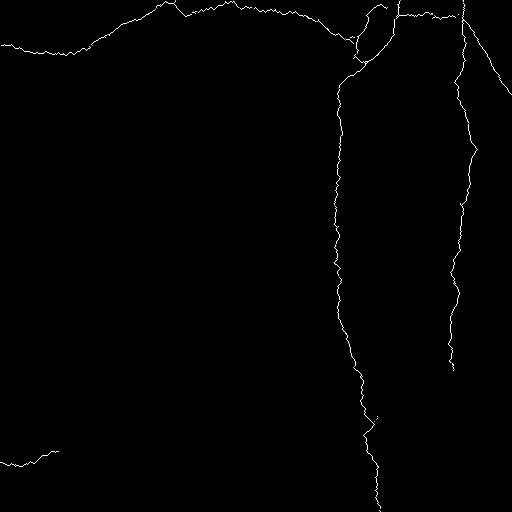} &
    \includegraphics[width=0.15\linewidth]{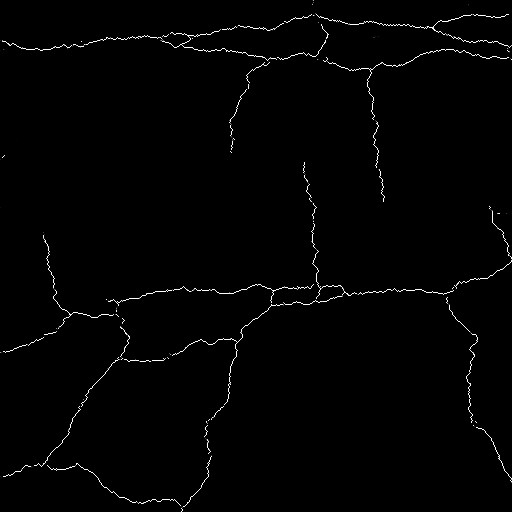} &
    \includegraphics[width=0.15\linewidth]{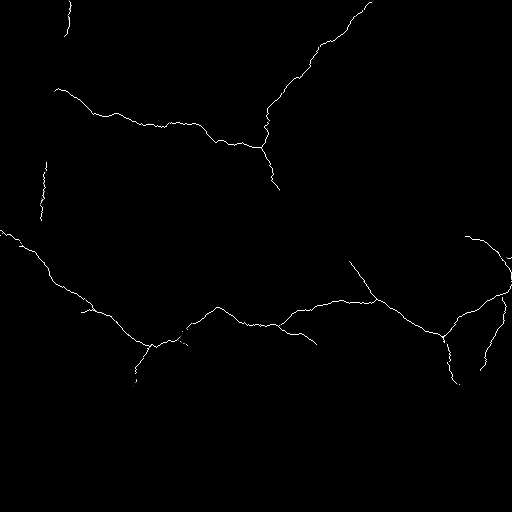} &
    \includegraphics[width=0.15\linewidth]{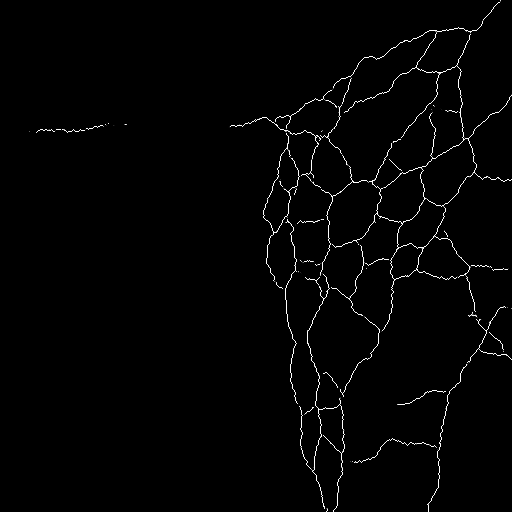} 
    \\
\end{tabular}
\caption{\footnotesize {\textbf{Visual results on CrackTree260.} (a) Input image. (b) Ground truth. (c) UNet. (d) CrackformerII. (e) DeepCrack. (f) cGAN\_CBAM (pixel). (g) cGAN\_CBAM\_Ig (pixel). (h)cGAN\_LSA (pixel).}
}
\label{fig:visual_cracktree260}
\end{figure}

\subsection{Ablation Study}
We compared our implementation with different structures based on the backbone to verify the effectiveness of each variant by conducting extensive experiments on datasets. 

\begin{itemize}

    \item \textit{1) Study of attention mechanisms}: To investigate the effect of various attention mechanisms, we trained three versions of our proposed framework: cGAN\_LSA, cGAN\_CBAM, and cGAN\_CBAM\_Ig. cGAN\_LSA indicates the generator in Fig. \ref{Fig:generator} embedding with the LSA module Fig. \ref{Fig:atten_lsa}. cGAN\_CBAM and cGAN\_CBAM\_Ig work with the common CBAM attention module and the CBAM for ignoring module in Fig.\ref{Fig:atten_cbam} separately. 
    
    The corresponding experimental results on the six datasets are shown in Table. II-VII. It can be seen that the proposed framework with attention mechanisms shows better and more stable performance than the competing methods on the six datasets. Especially, the CBAM module with the pixel-level discriminator has achieved the best ODS and OIS on DeepCrack-DB, CFD, CrackLS315, and CRKWH100. 

    \item \textit{2) Study of discriminator:} Fig. \ref{Fig:discriminators} illustrates two different discriminators: pixel-level and image-level. We first conduct the ablation experiments on the proposed framework with Fig. \ref{Fig:discriminators}(a) and Fig. \ref{Fig:discriminators}(b). Observing the results shown in Table. II-VII, the pixel-level discriminator can perform better in most cases than the image-level discriminator on the six datasets under the same conditions. For instance, cGAN\_LSA with pixel-level discriminator has achieved better ODS and OIS on CRACK500, DeepCrack-DB, CFD, CrackLS315, and CrackTree260 than with image-level discriminator.

    \item \textit{3) Study of loss functions}: We further conduct the ablation experiments involving loss functions on DeepCrack-DB and CrackLS315. We conduct the related experiments based on the cGAN\_CBAM\_Ig architecture with the pixel discriminator, and then we train one model with $\mathcal{L}_{cGANs}$, $\mathcal{L}_{KL}$, $\mathcal{L}_{CE}$, and $\mathcal{L}_{TL}$, one model with $\mathcal{L}_{cGANs}$, $\mathcal{L}_{KL}$, $\mathcal{L}_{CE}$, and $\mathcal{L}_{Side}$, and one model with $\mathcal{L}_{cGANs}$, $\mathcal{L}_{KL}$, $\mathcal{L}_{CE}$, as shown in Table. VIII and Table. IX. It can be observed that the Side loss and Tversky loss can bring absolute improvements.
\end{itemize}

\subsection{Results and Comparisons}
We compare the proposed framework with related semantic segmentation models on six publicly pavement datasets with five evaluation metrics in Table. II-VII and precision-recall curves in Fig. 9. We adopt UNet \cite{Ronneberger2015}, HED \cite{Xie2015}, FPHB \cite{Yang2019}, V-GAN with pixel-level discriminator \cite{Son2017}, V-GAN with image-level discriminator \cite{Son2017}, DeepCrack \cite{Liu2019}, and CrackformerII \cite{Liu2023CrackFormerSegmentation} as competing method. The results show that the proposed framework with the pixel-level discriminator achieves the highest score on ODS and OIS compared with other competing models on the six pavement datasets. In addition, part of the visual results of the six datasets are presented. We only present the visual results obtained from our proposed framework with the pixel-level discriminator.
\begin{itemize}
    \item 1) \textit{CRACK500:} The CRACK500 contains pavement images with diverse cracks from thin to wide, blur to sharp, etc. The quantitative results on CRACK500 are shown in Table. II. As shown in the table, the best ODS and Mean IOU are achieved by cGAN\_CBAM with the pixel-level discriminator as 0.7675 and 0.7988, respectively. The competing method DeepCrack \cite{Liu2019} has the best OIS and AP at 0.7511 and 0.8219, respectively. The second-rank performance of ODS, OIS, and Mean IOU are from the cGAN\_CBAM\_Ig with the pixel-level discriminator at 0.7638, 0.7463, and 0.7963. From the visual results in Fig. \ref{fig:visual_crack500}, it can be observed that the cGAN\_LSA can learn more subtle cracks compared to other attention mechanisms.

    \item 2) \textit{DeepCrack-DB:} The pavement images of this dataset contain diverse shapes of crack under different illumination, especially, since the pavement states are occluded by shadows, curbs, etc. As shown in Table. III, our proposed framework outperforms all competing methods. The cGAN\_CBAM with the pixel-level discriminator obtains the best ODS of 0.8926, Global Accuracy of 0.9924, and Mean IOU of 0.8991. The highest OIS is achieved by the cGAN\_CBAM\_Ig with the pixel-level discriminator at 0.8765. In Fig. \ref{fig:visual_deepcrack}, our proposed framework with CBAM ignoring attention mechanism has shown the ability to distinguish the occlusions and with the LSA attention mechanism can learn more subtle crack pixels.

    \item 3) \textit{CFD:} The pavement images of this dataset are quite noisy. As shown in Table. IV, the cGAN\_CBAM with the pixel-level discriminator achieves the highest ODS, OIS, Global Accuracy, and Mean IOU at 0.7968, 0.7986, 0.9935, and 0.8278. The cGAN\_CBAM\_Ig with the pixel-level discriminator obtains the second rank of ODS, OIS, Global Accuracy, and Mean IOU. The cGAN\_CBAM\_Ig obtains improvements of 22.2\% and 18.5\% in terms of ODS and OIS compared to the results from DeepCrack \cite{Liu2019}, respectively. Fig. \ref{fig:visual_cfd} demonstrates that our framework can suppress the background under poor exposure status and occlusions. 

    \item 4) \textit{CrackLS315:} This dataset is famous for its notorious states since the invisible cracks, noisy backgrounds, and under dark illumination. The quantitative results in Table. V from our proposed framework have gained a huge improvement compared to the results from the competing methods. The cGAN\_CBAM with pixel-level discriminator achieves the best performance under all evaluation metrics as ODS: 0.5418, OIS: 0.5006, AP: 0.3520, Global Accuracy: 0.9980, and Mean IOU: 0.6846. The ODS of cGAN\_CBAM is 121.0\%, 46.9\%, and 71.7\% higher than the ODS of UNet, DeepCrack, and CrackformerII. The second-rank ODS of cGAN\_CBAM\_Ig is 117.0\%, 44.8\%, and 68.6\% higher than the ODS of UNet, DeepCrack, and CrackformerII.From the visual results in Fig. \ref{fig:visual_crackls315}, it can be observed that our proposed framework can generate feature maps close to the ground truth even with the hard cases. 

    \item 5) \textit{CRKWH100:} This dataset is another one containing hard cases. As shown in Table. VI, the cGAN\_CBAM\_Ig with the image-level discriminator achieves the highest ODS at 0.7982 which is 67.7\% and 168.5\% higher than the ODS of DeepCrack and CrackformerII. The best AP: 0.7493, Global Accuracy: 0.9989, and Mean IOU: 0.8315 are achieved by the cGAN\_CBAM\_Ig with the image-level discriminator too. Even the worst learned ODS: 0.7465 from cGAN\_LSA is 56.9\% higher than the ODS of DeepCrack. The visual results in Fig. \ref{fig:visual_crkwh100} demonstrate that our proposed framework can generate feature maps closer to the ground truth on this dataset for complex and simple cases. In addition, the LSA attention mechanism can preserve more subtle parts for complex cases.   

     \item 6) \textit{CrackTree260:} The pavement images of this dataset contain complex structure cracks and various occlusions. From the quantitative results shown in Table. VII, our proposed framework outperforms the competing methods, especially when working with the pixel-level discriminator. The cGAN\_CBAM\_Ig with the pixel-level discriminator achieves the best ODS, OIS, Global Accuracy, and Mean IOU at 0.8912, 0.9123, 0.9990, and 0.9014, respectively. The visual results in Fig. \ref{fig:visual_cracktree260} demonstrate that our proposed framework can generate feature maps closer to the ground truth on this dataset too. 
\end{itemize}
To further make a comparison, the precision-and-recall curves are shown in Fig. \ref{fig:pre_rec}. Besides the CRACK500 and DeepCrack-DB, our proposed framework outperforms the competing methods on the other four benchmark datasets in a robust manner, especially when working with the pixel-level discriminator.

\section{Conclusion}
In this article, We proposed a cGANs-based framework to address anomalous pavement surface conditions detection tasks, which is considered as binary semantic segmentation on imbalanced pavement datasets. To achieve steady and robust performance on diverse benchmark datasets, the proposed framework is trained with a two-stage strategy to capture the subtle but informative multiscale feature maps by embedding attention mechanisms. Moreover, we utilize and investigate various loss functions to address imbalance problems for better performance. We have verified the effectiveness of our proposed framework through comprehensive experiments on six benchmark datasets compared to seven competing methods with five evaluation metrics, precision-and-recall curves, and visual results. The evaluations show that our proposed framework achieved SOTA performance on the datasets in a robust manner without an obvious computational complexity growth.


%


\appendix

\ifCLASSOPTIONcaptionsoff
  \newpage
\fi



%
\bibliographystyle{IEEEtran}
\bibliography{references}

\begin{thebibliography}{10}
\providecommand{\url}[1]{#1}
\csname url@samestyle\endcsname
\providecommand{\newblock}{\relax}
\providecommand{\bibinfo}[2]{#2}
\providecommand{\BIBentrySTDinterwordspacing}{\spaceskip=0pt\relax}
\providecommand{\BIBentryALTinterwordstretchfactor}{4}
\providecommand{\BIBentryALTinterwordspacing}{\spaceskip=\fontdimen2\font plus
\BIBentryALTinterwordstretchfactor\fontdimen3\font minus
  \fontdimen4\font\relax}
\providecommand{\BIBforeignlanguage}[2]{{%
\expandafter\ifx\csname l@#1\endcsname\relax
\typeout{** WARNING: IEEEtran.bst: No hyphenation pattern has been}%
\typeout{** loaded for the language `#1'. Using the pattern for}%
\typeout{** the default language instead.}%
\else
\language=\csname l@#1\endcsname
\fi
#2}}
\providecommand{\BIBdecl}{\relax}
\BIBdecl

\bibitem{Yang2019}
F.~Yang, L.~Zhang, S.~Yu, D.~Prokhorov, X.~Mei, and H.~Ling, ``{Feature Pyramid
  and Hierarchical Boosting Network for Pavement Crack Detection},'' \emph{IEEE
  Transactions on Intelligent Transportation Systems}, pp. 1--11, 2019.

\bibitem{Zhang2021CrackGAN:Learning}
K.~Zhang, Y.~Zhang, and H.~D. Cheng, ``{CrackGAN: Pavement Crack Detection
  Using Partially Accurate Ground Truths Based on Generative Adversarial
  Learning},'' \emph{IEEE Transactions on Intelligent Transportation Systems},
  vol.~22, no.~2, pp. 1306--1319, 2021.

\bibitem{Sampath2021}
V.~Sampath, I.~Maurtua, J.~Jos{\'{e}}, A.~Mart{\'{i}}n, and A.~Gutierrez, ``{A
  survey on generative adversarial networks for imbalance problems in computer
  vision tasks},'' \emph{Journal of Big Data}, vol.~8, no.~27, 2021.

\bibitem{Tabernik2020}
D.~Tabernik, S.~{\v{S}}ela, J.~Skvar{\v{c}}, and D.~Sko{\v{c}}aj,
  ``{Segmentation-based deep-learning approach for surface-defect detection},''
  \emph{Journal of Intelligent Manufacturing}, vol.~31, no.~3, pp. 759--776,
  2020.

\bibitem{Liu2019}
Y.~Liu, J.~Yao, X.~Lu, R.~Xie, and L.~Li, ``{DeepCrack : A deep hierarchical
  feature learning architecture for crack segmentation},''
  \emph{Neurocomputing}, vol. 338, pp. 139--153, 2019.

\bibitem{Xia2020}
Y.~Xia, Y.~Zhang, F.~Liu, W.~Shen, and A.~L. Yuille, ``{Synthesize Then
  Compare: Detecting Failures and Anomalies for Semantic Segmentation},''
  \emph{Lecture Notes in Computer Science (including subseries Lecture Notes in
  Artificial Intelligence and Lecture Notes in Bioinformatics)}, vol. 12346
  LNCS, pp. 145--161, 2020.

\bibitem{Rezaei2019a}
M.~Rezaei, H.~Yang, K.~Harmuth, and C.~Meinel, ``{Conditional generative
  adversarial refinement networks for unbalanced medical image semantic
  segmentation},'' \emph{Proceedings - 2019 IEEE Winter Conference on
  Applications of Computer Vision, WACV 2019}, pp. 1836--1845, 2019.

\bibitem{Chawla2002}
N.~V. Chawla, K.~W. Bowyer, L.~O. Hall, and W.~P. Kegelmeyer, ``{SMOTE:
  Synthetic Minority Over-sampling Technique},'' \emph{Journal of Artificial
  Intelligence Research}, vol.~16, no. Sept. 28, pp. 321--357, 2002.

\bibitem{Simard2003}
P.~Y. Simard, D.~Steinkraus, and J.~C. Platt, ``{Best practices for
  convolutional neural networks applied to visual document analysis},''
  \emph{Proceedings of the International Conference on Document Analysis and
  Recognition, ICDAR}, vol. 2003-Janua, pp. 958--963, 2003.

\bibitem{Rezaei2017a}
M.~Rezaei, H.~Yang, and C.~Meinel, ``{Deep Neural Network with l2-Norm Unit for
  Brain Lesions Detection},'' \emph{Lecture Notes in Computer Science
  (including subseries Lecture Notes in Artificial Intelligence and Lecture
  Notes in Bioinformatics)}, vol. 10637 LNCS, pp. 798--807, 2017.

\bibitem{Lin2020}
T.~Y. Lin, P.~Goyal, R.~Girshick, K.~He, and P.~Dollar, ``{Focal Loss for Dense
  Object Detection},'' \emph{IEEE Transactions on Pattern Analysis and Machine
  Intelligence}, vol.~42, no.~2, pp. 318--327, 2020.

\bibitem{Lee2005}
C.~H. Lee, R.~Greiner, and M.~Schmidt, ``{Support vector random fields for
  spatial classification},'' \emph{Lecture Notes in Computer Science (including
  subseries Lecture Notes in Artificial Intelligence and Lecture Notes in
  Bioinformatics)}, vol. 3721 LNAI, pp. 121--132, 2005.

\bibitem{Krahenbuhl2011}
P.~Kr{\"{a}}henb{\"{u}}hl and V.~Koltun, ``{Efficient inference in fully
  connected crfs with Gaussian edge potentials},'' \emph{Advances in Neural
  Information Processing Systems 24: 25th Annual Conference on Neural
  Information Processing Systems 2011, NIPS 2011}, pp. 1--9, 2011.

\bibitem{Caesar2015}
H.~Caesar, J.~Uijlings, and V.~Ferrari, ``{Joint Calibration for Semantic
  Segmentation},'' in \emph{BMVC (British Machine Vision Conference)}, 2015,
  pp. 1--29.

\bibitem{Yang2014}
J.~Yang, B.~Price, S.~Cohen, and M.~H. Yang, ``{Context driven scene parsing
  with attention to rare classes},'' \emph{Proceedings of the IEEE Computer
  Society Conference on Computer Vision and Pattern Recognition}, pp.
  3294--3301, 2014.

\bibitem{Shi2016}
Y.~Shi, L.~Cui, Z.~Qi, F.~Meng, and Z.~Chen, ``{Automatic road crack detection
  using random structured forests},'' \emph{IEEE Transactions on Intelligent
  Transportation Systems}, vol.~17, no.~12, pp. 3434--3445, 2016.

\bibitem{Lim2014}
R.~S. Lim, H.~M. La, and W.~Sheng, ``{A robotic crack inspection and mapping
  system for bridge deck maintenance},'' \emph{IEEE Transactions on Automation
  Science and Engineering}, vol.~11, no.~2, pp. 367--378, 2014.

\bibitem{Nakazawa2019}
T.~Nakazawa and D.~V. Kulkarni, ``{Anomaly detection and segmentation for wafer
  defect patterns using deep Convolutional Encoder-Decoder Neural Network
  Architectures in Semiconductor Manufacturing},'' \emph{IEEE Transactions on
  Semiconductor Manufacturing}, vol.~32, no.~2, pp. 250--256, 2019.

\bibitem{Son2017}
J.~Son, S.~J. Park, and K.-H. Jung, ``{Retinal Vessel Segmentation in
  Fundoscopic Images with Generative Adversarial Networks},''
  \emph{arXiv:1706.09318}, 2017.

\bibitem{Zou2019DeepCrack:Detection}
Q.~Zou, Z.~Zhang, Q.~Li, X.~Qi, Q.~Wang, and S.~Wang, ``{DeepCrack: Learning
  hierarchical convolutional features for crack detection},'' \emph{IEEE
  Transactions on Image Processing}, vol.~28, no.~3, pp. 1498--1512, 2019.

\bibitem{Mohan2018}
A.~Mohan and S.~Poobal, ``{Crack detection using image processing: A critical
  review and analysis},'' \emph{Alexandria Engineering Journal}, vol.~57,
  no.~2, pp. 787--798, 2018.

\bibitem{Nithya2015a}
P.~S.~S. Nithya, P.~Sathya, and S.~Pradeepa, ``{A nondestructive sensing robot
  for crack detection and deck maintenance},'' \emph{International Journal For
  Research in Applied Science and Engineering Technology}, vol.~3, no.~Iv, pp.
  663--673, 2015.

\bibitem{Meksen2010}
T.~M. Meksen, B.~Boudraa, R.~Drai, and M.~Boudraa, ``{Automatic crack detection
  and characterization during ultrasonic inspection},'' \emph{Journal of
  Nondestructive Evaluation}, vol.~29, no.~3, pp. 169--174, 2010.

\bibitem{Cubero-Fernandez2017a}
A.~Cubero-Fernandez, F.~J. Rodriguez-Lozano, R.~Villatoro, J.~Olivares, and
  J.~M. Palomares, ``{Efficient pavement crack detection and classification},''
  \emph{Eurasip Journal on Image and Video Processing}, vol. 2017, no.~1, 2017.

\bibitem{Mirza2014}
M.~Mirza and S.~Osindero, ``{Conditional Generative Adversarial Nets},''
  \emph{arXiv preprint arXiv:1411.1784}, pp. 1--7, 2014.

\bibitem{Xie2015}
S.~Xie and Z.~Tu, ``{Holistically-nested edge detection},'' \emph{Proceedings
  of the IEEE International Conference on Computer Vision}, vol. 2015 Inter,
  pp. 1395--1403, 2015.

\bibitem{Simonyan2015VeryRecognition}
K.~Simonyan and A.~Zisserman, ``{Very deep convolutional networks for
  large-scale image recognition},'' in \emph{3rd International Conference on
  Learning Representations, ICLR 2015 - Conference Track Proceedings}, 2015,
  pp. 1--14.

\bibitem{Liu2021CrackFormer:Detection}
H.~Liu, X.~Miao, C.~Mertz, C.~Xu, and H.~Kong, ``{CrackFormer: Transformer
  Network for Fine-Grained Crack Detection},'' \emph{Proceedings of the IEEE
  International Conference on Computer Vision}, pp. 3763--3772, 2021.

\bibitem{Liu2023CrackFormerSegmentation}
H.~Liu, J.~Yang, X.~Miao, C.~Mertz, and H.~Kong, ``{CrackFormer Network for
  Pavement Crack Segmentation},'' \emph{IEEE Transactions on Intelligent
  Transportation Systems}, vol.~24, no.~9, pp. 9240--9252, 2023.

\bibitem{Badrinarayanan2017SegNet:Segmentation}
V.~Badrinarayanan, A.~Kendall, and R.~Cipolla, ``{SegNet: A Deep Convolutional
  Encoder-Decoder Architecture for Image Segmentation},'' \emph{IEEE
  Transactions on Pattern Analysis and Machine Intelligence}, vol.~39, no.~12,
  pp. 2481--2495, 2017.

\bibitem{Kamran2021}
S.~A. Kamran, K.~F. Hossain, A.~Tavakkoli, S.~L. Zuckerbrod, K.~M. Sanders, and
  S.~A. Baker, ``{RV-GAN: Segmenting Retinal Vascular Structure in Fundus
  Photographs Using a Novel Multi-scale Generative Adversarial Network},''
  \emph{Lecture Notes in Computer Science (including subseries Lecture Notes in
  Artificial Intelligence and Lecture Notes in Bioinformatics)}, vol. 12908
  LNCS, pp. 34--44, 2021.

\bibitem{Bahdanau2015NeuralTranslate}
D.~Bahdanau, K.~Cho, and Y.~Bengio, ``{Neural Machine Translation by Jointly
  Learning to Align and Translate},'' in \emph{ICLR}, 2015.

\bibitem{Xu2015ShowAttention}
K.~Xu, J.~L. Ba, R.~Kiros, K.~Cho, A.~Courville, R.~Salakhutdinov, R.~S. Zemel,
  and Y.~Bengio, ``{Show, Attend and Tell: Neural Image Caption Generation with
  Visual Attention},'' in \emph{Proceedings of the 32nd International
  Conference on Machine Learning}, 2015, pp. 2048--2057.

\bibitem{Oktay2018b}
O.~Oktay, J.~Schlemper, L.~L. Folgoc, M.~Lee, M.~Heinrich, K.~Misawa, K.~Mori,
  S.~McDonagh, N.~Y. Hammerla, B.~Kainz, B.~Glocker, and D.~Rueckert,
  ``{Attention U-Net: Learning Where to Look for the Pancreas},''
  \emph{arXiv:1804.03999}, 2018.

\bibitem{Schlemper2019}
J.~Schlemper, O.~Oktay, M.~Schaap, M.~Heinrich, B.~Kainz, B.~Glocker, and
  D.~Rueckert, ``{Attention gated networks: Learning to leverage salient
  regions in medical images},'' \emph{Medical Image Analysis}, vol.~53, pp.
  197--207, 2019.

\bibitem{Hu2018Squeeze-and-ExcitationNetworks}
J.~Hu, L.~Shen, and G.~Sun, ``{Squeeze-and-Excitation Networks},''
  \emph{Proceedings of the IEEE Conference on Computer Vision and Pattern
  Recognition (CVPR)}, pp. 7132--7141, 2018.

\bibitem{Laakom2021LearningCNNs}
F.~Laakom, K.~Chumachenko, J.~Raitoharju, A.~Iosifidis, and M.~Gabbouj,
  ``{Learning to ignore: rethinking attention in CNNs},'' in \emph{The British
  Machine Vision Conference (BMVC)}, 2021, pp. 1--13.

\bibitem{Wang2022UCTransNet:Transformer}
H.~Wang, P.~Cao, J.~Wang, and O.~R. Zaiane, ``{UCTransNet: Rethinking the Skip
  Connections in U-Net from a Channel-Wise Perspective with Transformer},'' in
  \emph{Proceedings of the 36th AAAI Conference on Artificial Intelligence,
  AAAI 2022}, vol.~36, 2022, pp. 2441--2449.

\bibitem{Brauwers2023ALearning}
G.~Brauwers and F.~Frasincar, ``{A General Survey on Attention Mechanisms in
  Deep Learning},'' \emph{IEEE Transactions on Knowledge and Data Engineering},
  vol.~35, no.~4, pp. 3279--3298, 2023.

\bibitem{Woo2018CBAMModule}
S.~Woo, J.~Park, J.-y. Lee, and I.~S. Kweon, ``{CBAM : Convolutional Block
  Attention Module},'' in \emph{European Conference on Computer Vision}, 2018.

\bibitem{Dosovitskiy2021ANSCALE}
A.~Dosovitskiy, L.~Beyer, A.~Kolesnikov, D.~Weissenborn, X.~Zhai,
  T.~Unterthiner, M.~Dehghani, M.~Minderer, G.~Heigold, S.~Gelly, J.~Uszkoreit,
  and N.~Houlsby, ``{AN IMAGE IS WORTH 16X16 WORDS: TRANSFORMERS FOR IMAGE
  RECOGNITION AT SCALE},'' in \emph{ICLR 2021 - 9th International Conference on
  Learning Representations}, 2021.

\bibitem{Wang2022MixedSegmentation}
H.~Wang, S.~Xie, L.~Lin, Y.~Iwamoto, X.~H. Han, Y.~W. Chen, and R.~Tong,
  ``{Mixed Transformer U-Net for Medical Image Segmentation},'' \emph{ICASSP,
  IEEE International Conference on Acoustics, Speech and Signal Processing -
  Proceedings}, vol. 2022-May, pp. 2390--2394, 2022.

\bibitem{Lin2017FeatureDetection}
T.~Y. Lin, P.~Doll{\'{a}}r, R.~Girshick, K.~He, B.~Hariharan, and S.~Belongie,
  ``{Feature pyramid networks for object detection},'' \emph{Proceedings - 30th
  IEEE Conference on Computer Vision and Pattern Recognition, CVPR 2017}, vol.
  2017-Janua, pp. 936--944, 2017.

\bibitem{Abraham2019ASegmentation}
N.~Abraham and N.~M. Khan, ``{A novel focal tversky loss function with improved
  attention u-net for lesion segmentation},'' in \emph{Proceedings -
  International Symposium on Biomedical Imaging}, vol. 2019-April, 2019, pp.
  683--687.

\bibitem{Zhang2020ARecognition}
L.~Zhang, C.~Zhang, S.~Quan, H.~Xiao, G.~Kuang, L.~Liu, and L.~Liu, ``{A class
  imbalance loss for imbalanced object recognition},'' \emph{IEEE Journal of
  Selected Topics in Applied Earth Observations and Remote Sensing}, vol.~13,
  pp. 2778--2792, 2020.

\bibitem{Cui2019Class-balancedSamples}
Y.~Cui, M.~Jia, T.~Y. Lin, Y.~Song, and S.~Belongie, ``{Class-balanced loss
  based on effective number of samples},'' in \emph{Proceedings of the IEEE
  Computer Society Conference on Computer Vision and Pattern Recognition}, vol.
  2019-June, 2019, pp. 9260--9269.

\bibitem{Salehi2017TverskyNetworks}
S.~S.~M. Salehi, D.~Erdogmus, and A.~Gholipour, ``{Tversky loss function for
  image segmentation using 3D fully convolutional deep networks},'' in
  \emph{Lecture Notes in Computer Science (including subseries Lecture Notes in
  Artificial Intelligence and Lecture Notes in Bioinformatics)}, vol. 10541
  LNCS, 2017, pp. 379--387.

\bibitem{Yeung2022UnifiedSegmentation}
M.~Yeung, E.~Sala, C.~B. Sch{\"{o}}nlieb, and L.~Rundo, ``{Unified Focal loss:
  Generalising Dice and cross entropy-based losses to handle class imbalanced
  medical image segmentation},'' \emph{Computerized Medical Imaging and
  Graphics}, vol.~95, no. November 2021, 2022.

\bibitem{Cover1991ElementsTheory}
T.~M. Cover and J.~Thomas, \emph{{Elements of Information Theory}}.\hskip 1em
  plus 0.5em minus 0.4em\relax John Wiley {\&} Sons, Inc., 1991, vol.~1.

\bibitem{Goodfellow2013}
I.~Goodfellow, J.~Pouget-Abadie, M.~Mirza, B.~Xu, D.~Warde-Farley, S.~Ozair,
  A.~Courville, and Y.~Bengio, ``{Generative Adversarial Nets},'' in
  \emph{NIPS}, 2013, p. iii.

\bibitem{Ronneberger2015}
O.~Ronneberger, P.~Fischer, and T.~Brox, ``{U-net: Convolutional networks for
  biomedical image segmentation},'' in \emph{International Conference on
  Medical Image Computing and Computer-Assisted Intervention}, vol. 9351, 2015,
  pp. 234--241.

\bibitem{Zhu2017UnpairedNetworks}
J.~Y. Zhu, T.~Park, P.~Isola, and A.~A. Efros, ``{Unpaired Image-to-Image
  Translation Using Cycle-Consistent Adversarial Networks},'' in
  \emph{Proceedings of the IEEE International Conference on Computer Vision},
  vol. 2017-Octob, 2017, pp. 2242--2251.

\bibitem{Johnson2016PerceptualSuper-resolution}
J.~Johnson, A.~Alahi, and L.~Fei-Fei, ``{Perceptual losses for real-time style
  transfer and super-resolution},'' in \emph{Lecture Notes in Computer Science
  (including subseries Lecture Notes in Artificial Intelligence and Lecture
  Notes in Bioinformatics)}, vol. 9906 LNCS, 2016, pp. 694--711.

\bibitem{Nguyen2017}
V.~Nguyen, T.~F.~Y. Vicente, M.~Zhao, M.~Hoai, D.~Samaras, and S.~Brook,
  ``{Shadow Detection with Conditional Generative Adversarial Networks},'' in
  \emph{IEEE International Conference on Computer Vision Shadow}, 2017.

\bibitem{Liu2019a}
L.~Liu, M.~Muelly, J.~Deng, T.~Pfister, and L.~J. Li, ``{Generative modeling
  for small-data object detection},'' \emph{Proceedings of the IEEE
  International Conference on Computer Vision}, vol. 2019-Octob, pp.
  6072--6080, 2019.

\bibitem{Sun2022DMA-Net:Segmentation}
X.~Sun, Y.~Xie, L.~Jiang, Y.~Cao, and B.~Liu, ``{DMA-Net: DeepLab With
  Multi-Scale Attention for Pavement Crack Segmentation},'' \emph{IEEE
  Transactions on Intelligent Transportation Systems}, vol.~23, no.~10, pp.
  18\,392--18\,403, 2022.

\bibitem{Kingma2015Adam:Optimization}
D.~P. Kingma and J.~L. Ba, ``{Adam: A method for stochastic optimization},'' in
  \emph{3rd International Conference on Learning Representations, ICLR 2015 -
  Conference Track Proceedings}, 2015, pp. 1--15.

\bibitem{Otsu1979a}
N.~Otsu, ``{A Threshold Selection Method from Gray-Level Histogram},''
  \emph{IEEE Transactions on Systems, Man, and Cybernetics}, vol.~C, no.~1, pp.
  62--66, 1979.

\bibitem{Xiao2023PavementTransformers}
S.~Xiao, K.~Shang, K.~Lin, Q.~Wu, H.~Gu, and Z.~Zhang, ``{Pavement crack
  detection with hybrid-window attentive vision transformers},''
  \emph{International Journal of Applied Earth Observation and Geoinformation},
  vol. 116, no. December 2022, p. 103172, 2023.

\end{thebibliography}

%

\begin{IEEEbiographynophoto}
{Lei Xu} received the B.S.E.E degree from East China Normal University, ShangHai, China, in 2006, the M.S.E.E. degree from the University of Tampere, Finland, in 2017. From 2006 to 2013, she was an Engineer in ShangHai, where she was involved with on-train communication systems design. Her current research interests include Artificial Intelligence, data science, and Machine Learning. She is currently a PhD candidate at Tampere University.\end{IEEEbiographynophoto}

\begin{IEEEbiographynophoto}{Moncef Gabbouj}
received his BS degree in 1985 from Oklahoma State University, and his MS and PhD degrees from Purdue University, in 1986 and 1989, respectively, all in electrical engineering. Dr. Gabbouj is a Professor of Signal Processing at the Department of Computing Sciences, Tampere University, Tampere, Finland. He was Academy of Finland Professor during 2011-2015. His research interests include Big Data analytics, multimedia content-based analysis, indexing and retrieval, artificial intelligence, machine learning, pattern recognition, nonlinear signal and image processing and analysis, voice conversion, and video processing and coding. Dr. Gabbouj is a Fellow of the IEEE and member of the Academia Europaea and the Finnish Academy of Science and Letters. He is the past Chairman of the IEEE CAS TC on DSP and committee member of the IEEE Fourier Award for Signal Processing. He served as associate editor and guest editor of many IEEE, and international journals and Distinguished Lecturer for the IEEE CASS. Dr. Gabbouj is the Finland Site Director of the NSF IUCRC funded Center for Visual and Decision Informatics (CVDI) and leads the Artificial Intelligence Research Task Force of the Ministry of Economic Affairs and Employment funded Research Alliance on Autonomous Systems (RAAS).\end{IEEEbiographynophoto}




\end{document}